\documentclass[letterpaper]{article} 
\usepackage[]{aaai2027}  
\usepackage[hyphens]{url}  
\usepackage{graphicx} 
\usepackage{natbib}  
\usepackage{caption} 
\usepackage{algorithm}
\usepackage{algorithmic}
\usepackage{makecell}
\usepackage{newfloat}
\usepackage{listings}
\DeclareCaptionStyle{ruled}{labelfont=normalfont,labelsep=colon,strut=off} 
\floatstyle{ruled}
\newfloat{listing}{tb}{lst}{}
\floatname{listing}{Listing}

\usepackage{booktabs}

\nocopyright

\title{LayoutLite: Token-Level Implicit Layout Analysis for Efficient Document OCR}
\author{
    Xudong Liu\textsuperscript{\rm 1}\equalcontrib,
    Bicheng Wan\textsuperscript{\rm 1,\rm 2}\equalcontrib,
    Yulin Jin\textsuperscript{\rm 1}
}
\affiliations{
    \textsuperscript{\rm 1}Yuanli Technology (Beijing) Co., Ltd.\\
    \textsuperscript{\rm 2}Beijing Normal University\\
}

\usepackage{subcaption, amssymb, amsmath, array, graphicx}

\begin{document}

\maketitle

\begin{abstract}
End-to-end OCR systems based on vision-language models have achieved strong performance in complex document OCR, but their efficiency is severely limited by the large number of visual tokens produced from high-resolution document images. Many of these tokens correspond to blank margins or visually redundant regions, yet directly applying generic visual token compression methods may remove OCR-critical fine-grained details. In this paper, we propose LayoutLite, a lightweight plug-and-play module for efficient document OCR. Instead of relying on explicit document layout detection, LayoutLite performs implicit layout analysis at the token level between the vision encoder and the language decoder. It aggregates multi-layer visual representations from the vision encoder, and predicts an importance score for each visual token with a lightweight scoring network. Low-information tokens are then removed before entering the language decoder while preserving the original spatial positional information of retained tokens. To train LayoutLite without human annotations, we cast token selection as a reinforcement learning problem and optimize it with a group-relative policy optimization objective driven by OCR output consistency, together with an auxiliary layout supervision signal to stabilize training. Experiments on OmniDocBench v1.7 demonstrate that LayoutLite can substantially reduce visual token length and inference cost with negligible degradation in recognition quality. We further evaluate LayoutLite on two representative OCR-specialized VLMs, FireRed-OCR and Logics-Parsing-V2. Under up to 50\% token compression, LayoutLite preserves almost the same OmniDocBench v1.7 score on both models while reducing prefill latency, FLOPs, and KV cache memory by over 40\%, with only a small additional inference overhead. These results show that token-level implicit layout analysis is an effective and practical approach for accelerating VLM-based OCR systems.

\end{abstract}

\begin{links}
    \link{Code}{https://github.com/dpxudong/LayoutLite/}
\end{links}
\section{Introduction}

\begin{figure}[]
    \includegraphics[width=0.50\textwidth]{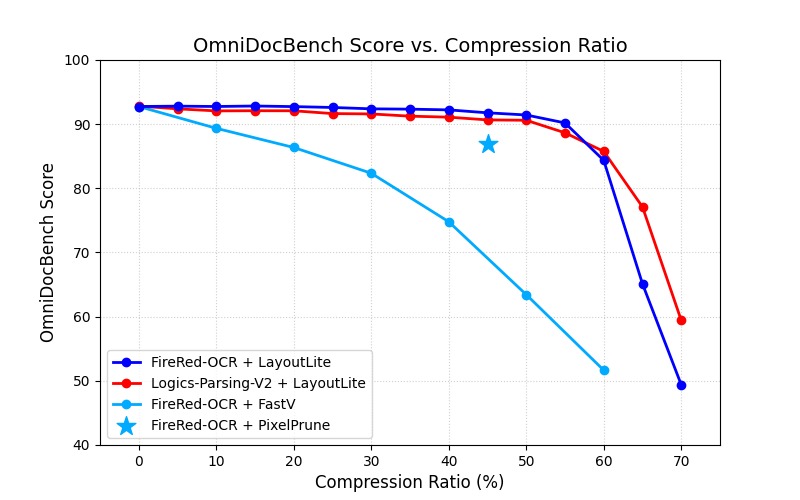}
    \caption{Performance comparison under different visual token compression ratios on OmniDocBench. LayoutLite maintains highly stable OCR performance on different models when the compression ratio is below 50\%, outperforming existing methods.}
    \label{fig:score}
\end{figure}
\begin{figure}[]
    \includegraphics[width=0.46\textwidth]{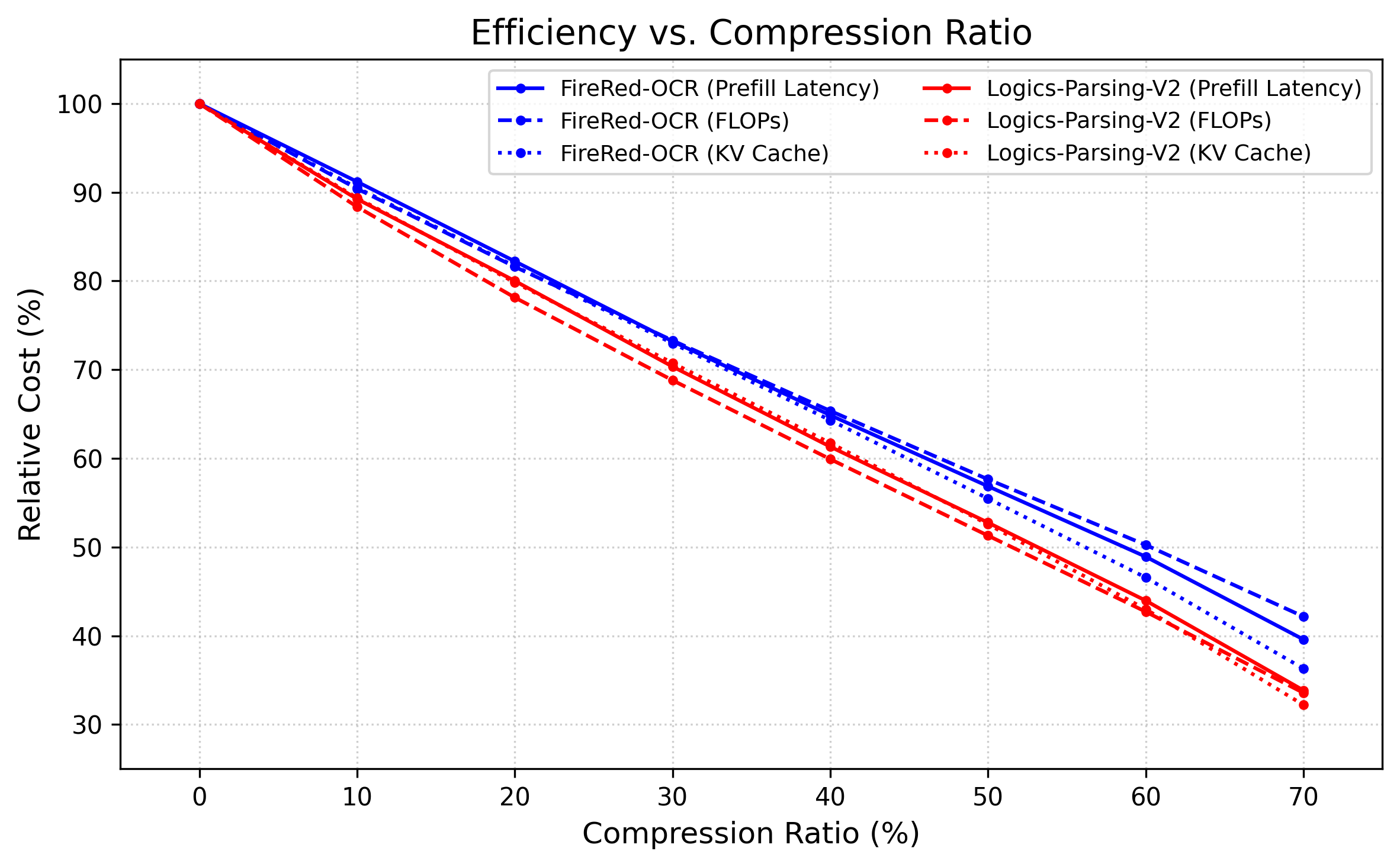}
    \caption{Efficiency comparison under different visual token compression ratios. Relative prefill latency, FLOPs, and KV cache (including the overhead introduced by LayoutLite). All three metrics decrease almost linearly as the compression ratio increases, reducing to approximately 50–60\% of the original cost at 50\% compression.}
    \label{fig:efficiency}
\end{figure}

\begin{figure*}[t]
\centering
\begin{subfigure}[t]{0.24\textwidth}
    \centering
    \setlength{\fboxsep}{0pt}
    \fbox{\includegraphics[width=\linewidth]{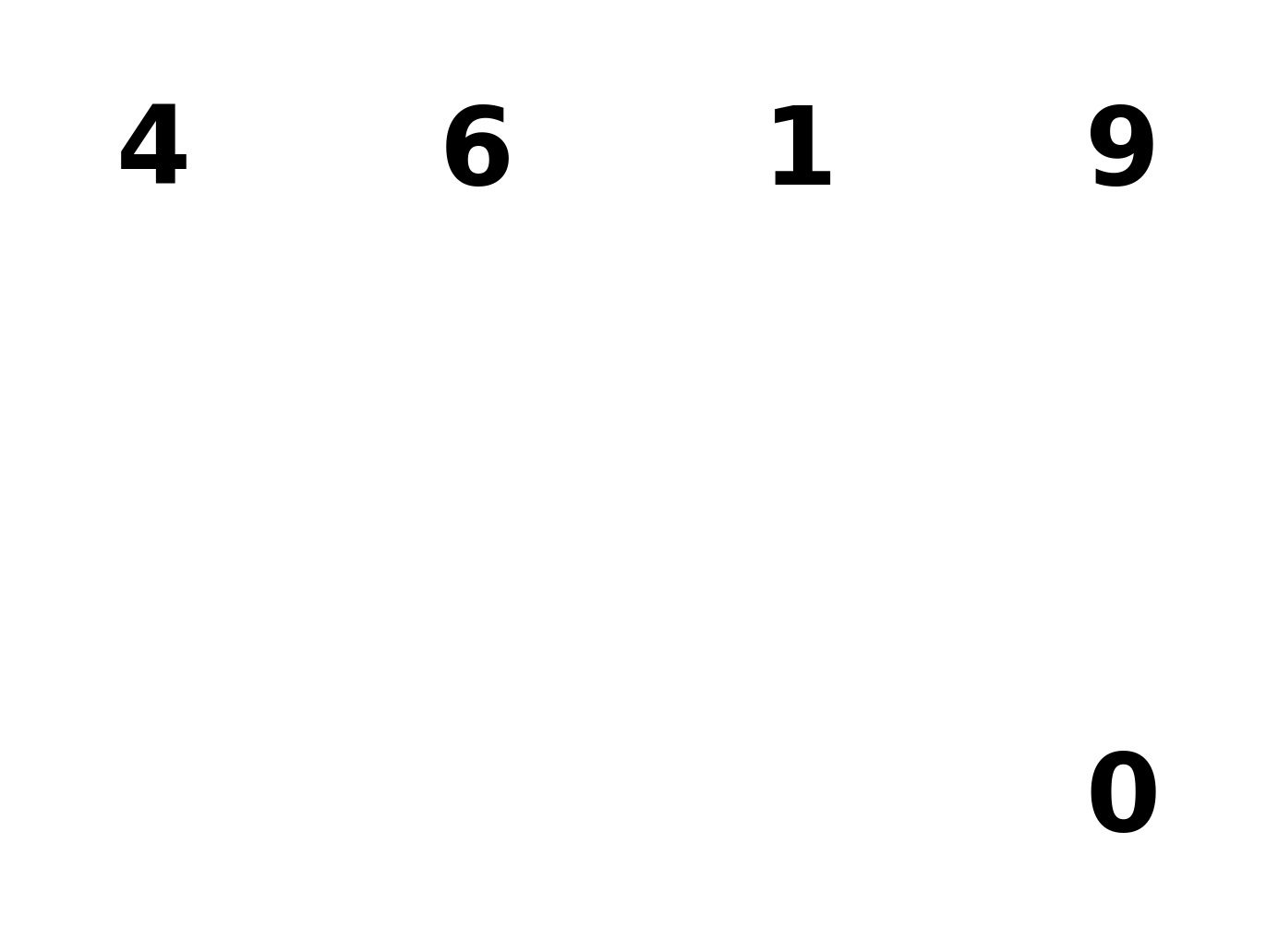}}\\[2pt]
    {\small\textbf{Model Output:} 4 6 1 9 0}
\end{subfigure}
\hfill
\begin{subfigure}[t]{0.24\textwidth}
    \centering
    \setlength{\fboxsep}{0pt}
    \fbox{\includegraphics[width=\linewidth]{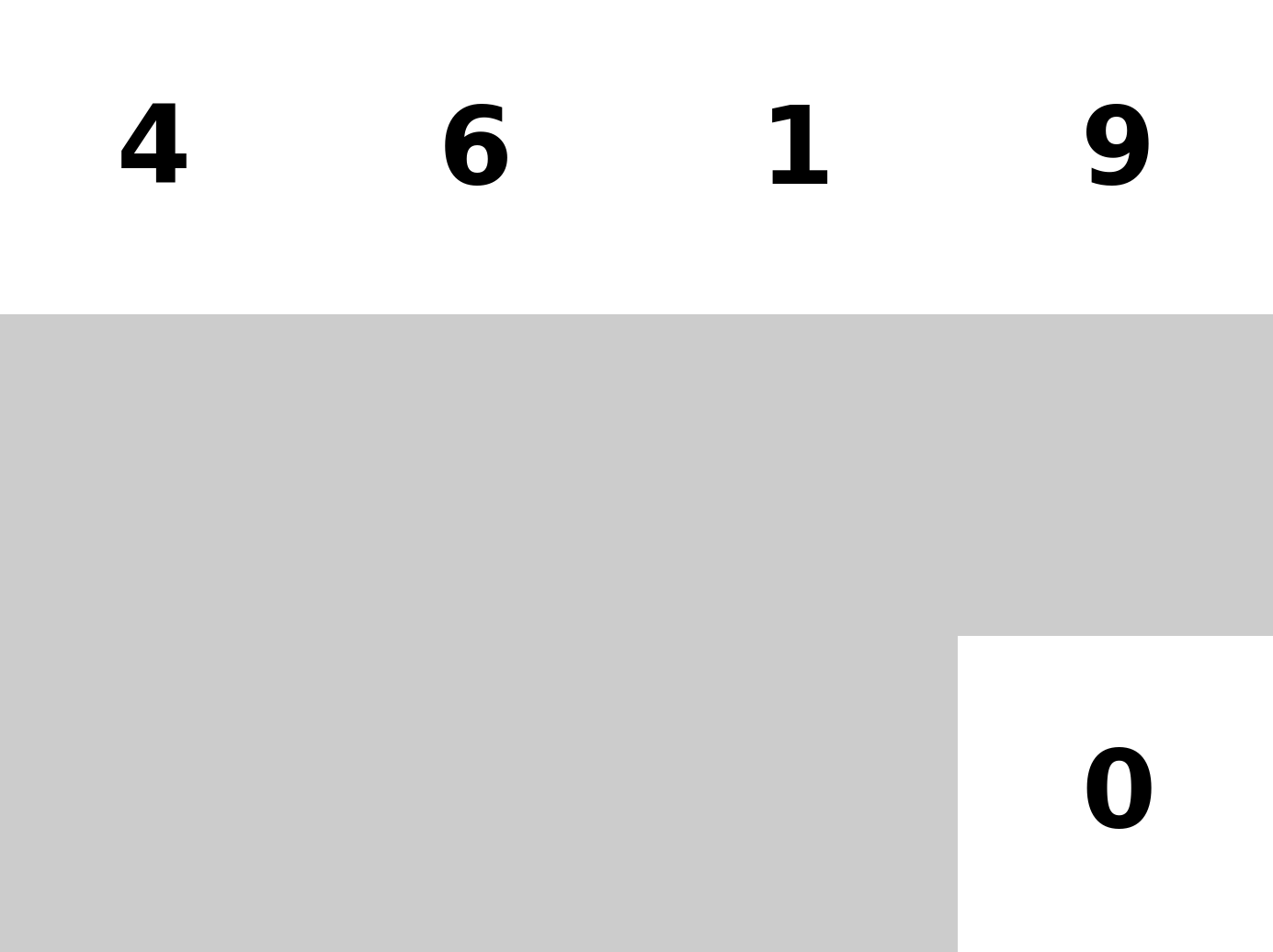}}\\[2pt]
    {\small\textbf{Model Output:} 4 6 1 9 0}
\end{subfigure}
\hfill
\begin{subfigure}[t]{0.24\textwidth}
    \centering
    \setlength{\fboxsep}{0pt}
    \fbox{\includegraphics[width=\linewidth]{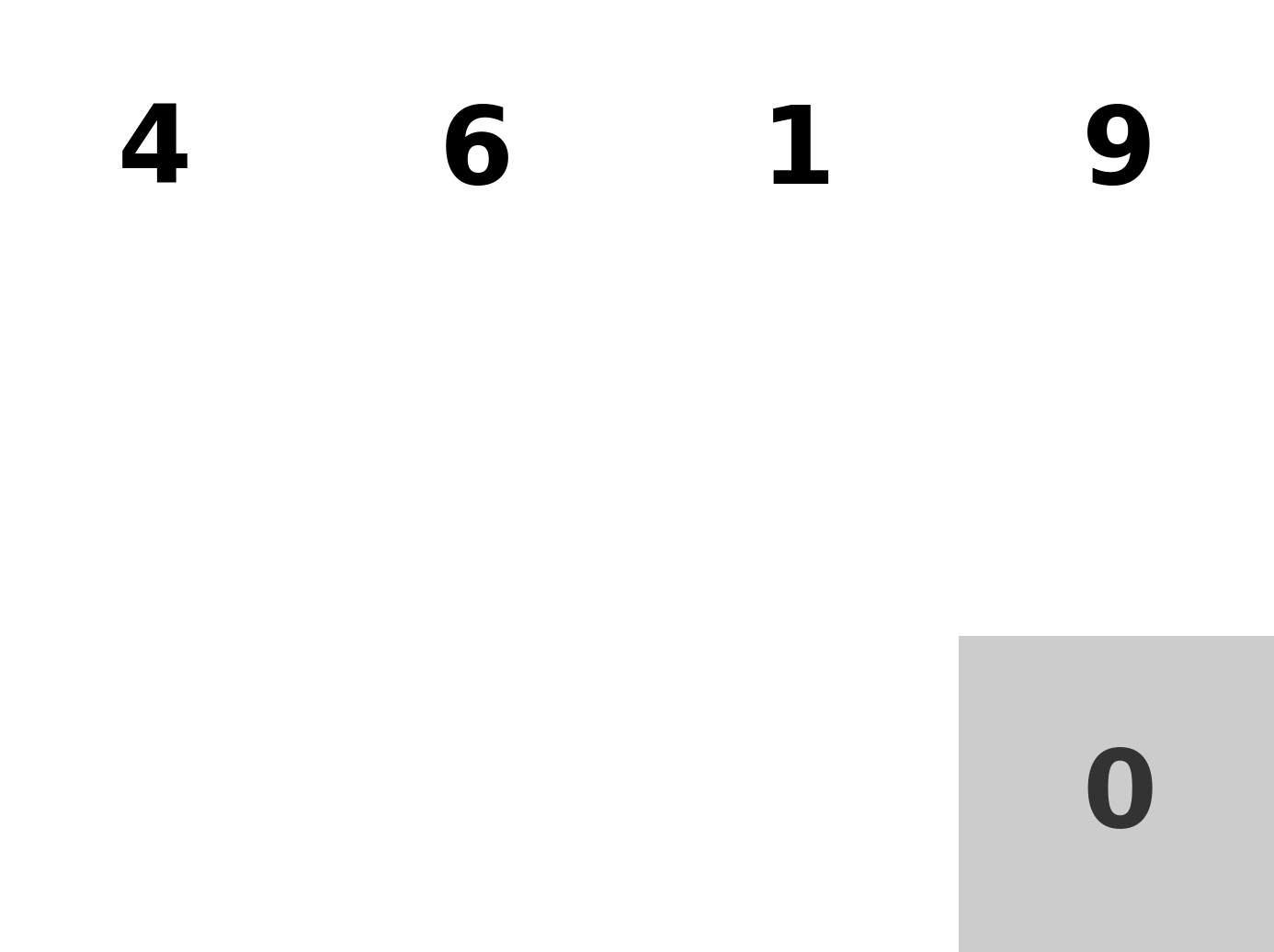}}\\[2pt]
    {\small\textbf{Model Output:} 4 6 1 9}
\end{subfigure}
\hfill
\begin{subfigure}[t]{0.24\textwidth}
    \centering
    \setlength{\fboxsep}{0pt}
    \fbox{\includegraphics[width=\linewidth]{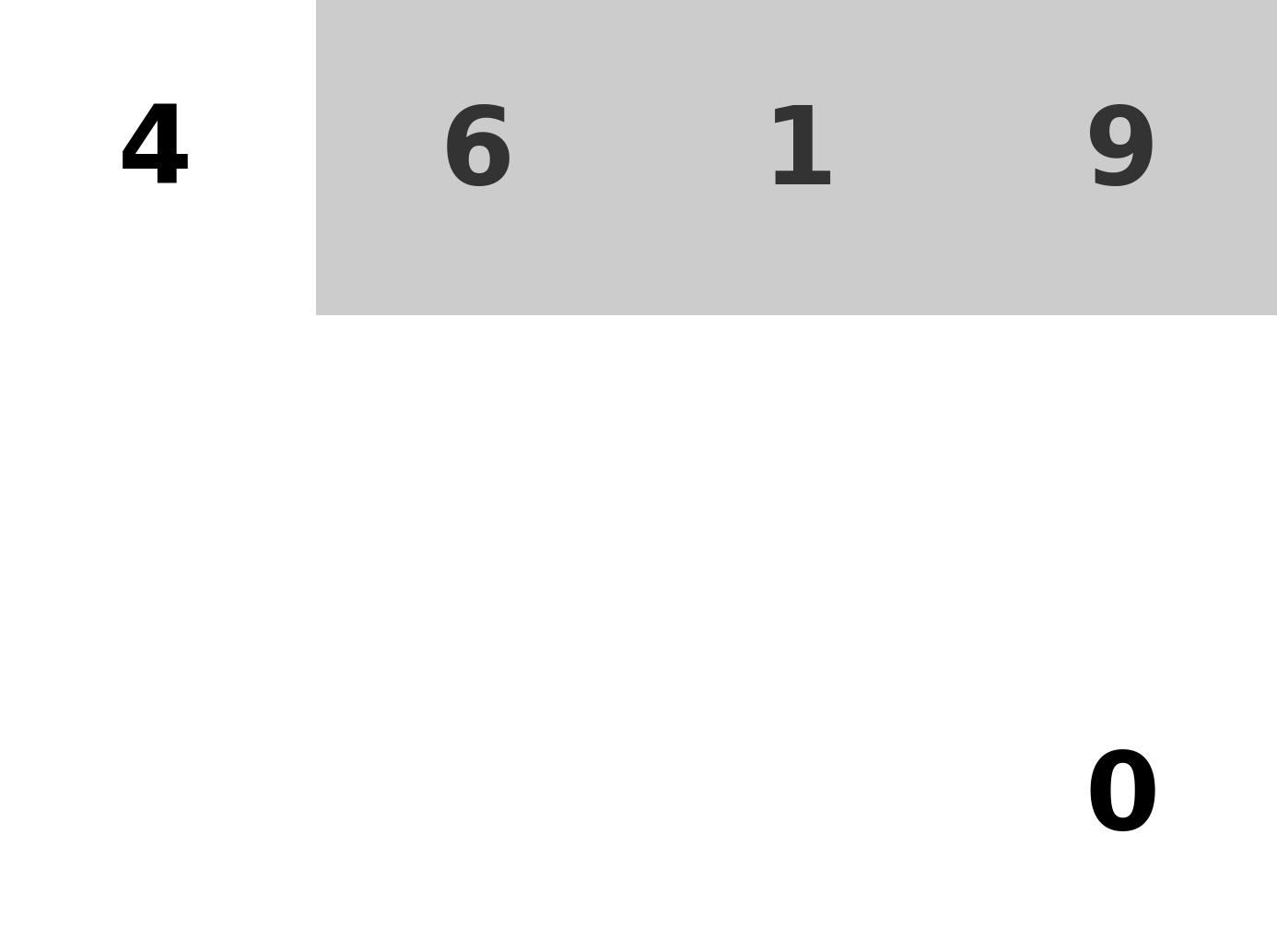}}\\[2pt]
    {\small\textbf{Model Output:} 4 0}
\end{subfigure}
\caption{Comparison of the impact of removing different visual tokens on the output of FireRed-OCR. Gray regions indicate deleted tokens. Removing tokens in blank areas has no effect, whereas removing tokens over numeric regions causes the model to omit the corresponding digits.}
\label{fig:model_output_comparison}
\end{figure*}

Recent advances in document OCR have been largely driven by Vision-Language Models (VLMs), which unify visual perception and language generation within a single framework. Benefiting from their strong multimodal reasoning capabilities, modern OCR systems built upon VLMs are able to handle complex document elements such as mathematical formulas and tables. Existing document OCR systems mainly follow three paradigms. The first adopts dynamic-resolution visual encoders, such as HunyuanOCR \cite{hunyuanvisionteam2025hunyuanocrtechnicalreport}, FireRed-OCR \cite{fireredocr}, and Logics-Parsing-V2 \cite{an2026logicsparsingomnitechnicalreport}, which allocate visual tokens according to the input resolution but still encode blank or redundant regions, since token allocation is driven by image size rather than information content. The second employs a two-stage pipeline, such as PaddleOCR-VL \cite{cui2025paddleocrvlboostingmultilingualdocument} and MinerU \cite{wang2024mineru}, which first performs layout analysis and then recognizes cropped regions; while effective in practice, such methods rely heavily on accurate layout detection and suffer from error propagation under complex layouts or distortions. The third uses fixed-resolution encoders, such as GOT-OCR2.0 \cite{wei2024general}, DeepSeek-OCR-V2 \cite{wei2026deepseek}, and Unlimited-OCR \cite{yin2026unlimitedocrworks}, which resize each image to a predefined resolution and inevitably encode numerous blank regions into visual tokens. Despite their different designs, none of these paradigms explicitly identifies and removes low-information visual tokens before language decoding, leaving redundant visual tokens as a common source of computational overhead in document OCR systems.

In this work, we propose \textbf{LayoutLite}, a lightweight plug-and-play module for efficient document OCR that unifies the strengths of layout analysis and visual token compression. Unlike conventional two-stage pipelines that perform \emph{explicit} layout analysis by detecting region boundaries, LayoutLite performs \emph{implicit layout analysis at the token level}: rather than predicting any explicit layout region, it directly scores the informativeness of each individual visual token between the vision encoder and the language decoder, and discards low-information tokens. Specifically, the proposed module aggregates representations from multiple layers of the vision encoder and jointly evaluates the information content of each visual token. Tokens that carry little useful information are removed before entering the language decoder, thereby reducing the visual context length and accelerating the decoding process.

The proposed method is motivated by a simple observation. As shown in Fig.~\ref{fig:model_output_comparison}, when the visual tokens corresponding to certain digits are removed from an image before being fed into the text decoder, the decoder simply skips the removed digits and outputs only the remaining ones. Such phenomenon is observed across different OCR models. This demonstrates that visual tokens in OCR models mainly encode localized information from their corresponding image regions. Such locality suggests that implicit layout analysis can identify and discard redundant visual tokens, thereby reducing the token count without sacrificing OCR accuracy.

A key advantage of LayoutLite is that it preserves OCR-sensitive visual details while eliminating redundant visual regions. Because it operates directly on visual tokens rather than relying on detected layout regions, LayoutLite remains robust under unconventional layouts, rotated documents, and layout detection failures. The module requires no fine-tuning of the original model and can be trained using only a small amount of unlabeled document data.

Our contributions can be summarized as follows:
\begin{itemize}
    \item We identify the limitations of existing OCR acceleration approaches, including explicit layout-based pipelines and generic visual token compression methods.
    \item We propose LayoutLite, a lightweight and plug-and-play module that performs implicit layout analysis at the token level between visual encoding and text decoding.
    \item Experimental results demonstrate that LayoutLite effectively reduces visual context length and inference cost with negligible performance degradation, and can even slightly improve recognition accuracy under moderate compression ratios.
\end{itemize}

\section{Related Work}

\subsubsection{VLM-based OCR}
Recent advances in vision-language models (VLMs) have driven the development of end-to-end document OCR systems. Representative models, including FireRed-OCR~\cite{fireredocr}, Logics-Parsing-V2~\cite{an2026logicsparsingomnitechnicalreport}, DeepSeek-OCR-V2~\cite{wei2025deepseek}, QianfanOCR~\cite{dong2026qianfanocrunifiedendtoendmodel}, Unlimited-OCR \cite{yin2026unlimitedocrworks} and HunyuanOCR~\cite{hunyuanvisionteam2025hunyuanocrtechnicalreport}, directly translate document images into structured text within a unified autoregressive framework. Compared with traditional multi-stage pipelines such as PaddleOCR-VL~\cite{cui2025paddleocrvlboostingmultilingualdocument} and MinerU~\cite{wang2024mineru}, these models avoid error propagation across independently optimized stages and demonstrate stronger capabilities in understanding complex document structures.

However, OCR methods entirely based on VLMs suffer from massive visual tokens generated by high-resolution document images. Moreover, unlike natural scenes, document pages are typically occupied by uniform white spaces, margins, and blank filler regions, introducing severe redundancy. The explosion of visual tokens not only significantly escalates the model's inference overhead, but also excessively bloats the context length. Consequently, the downstream LLM becomes highly susceptible to context-overload issues, leading to generation errors such as repetitive outputs, skipped information, or structural faults.
\subsubsection{Vision Token Compression in VLMs}

\begin{figure*}[t]
    \centering
    \includegraphics[width=\textwidth]{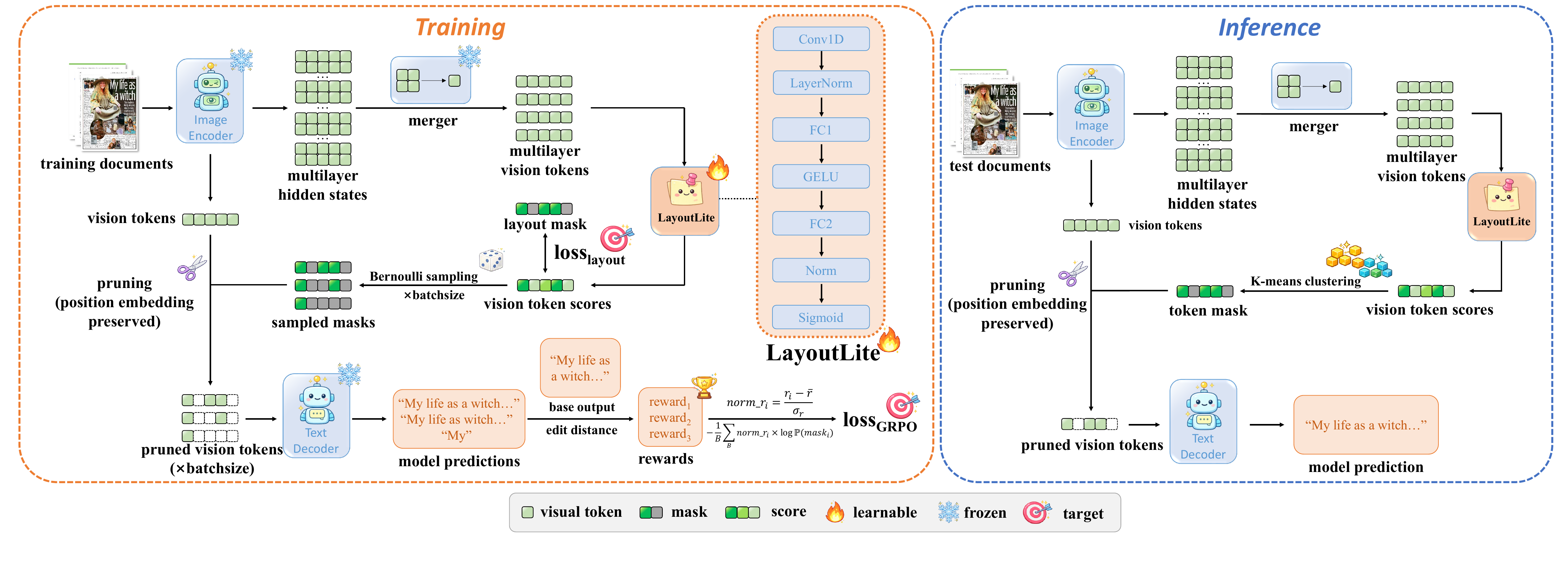}
    \caption{Overall framework of LayoutLite. Left: training with GRPO and layout supervision. Right: inference with cluster-based token pruning.}
    \label{fig:framework}
\end{figure*}

Various token pruning methods have been proposed to improve inference efficiency. Among these, representative training-free approaches that require no fine-tuning of the original models include FastV \cite{chen2024image}, a training-free, plug-and-play inference acceleration method. Based on the observation that visual tokens receive diminishingly low attention scores in deep layers, FastV dynamically prunes redundant visual tokens in subsequent layers according to an attention-score ranking, thereby significantly reducing FLOPs without sacrificing multimodal performance. PixelPrune \cite{wang2026pixelprunepixelleveladaptivevisual} also proposes a training-free visual token compression strategy that operates directly in the pixel space before visual encoding. PixelPrune identifies redundant image patches through predictive coding and removes patches with low information content. Other training-free methods include DivPrune \cite{alvar2025divprune} and FitPrune \cite{ye2025fit}. Remaining methods typically require fine-tuning on the LLM or more modules, including Q-Former \cite{li2023blip}, Vision Selector \cite{zhu2025visionselectorendtoendlearnablevisual} and VisionZip \cite{yang2025visionzip}. Although these methods perform well on general benchmarks, their direct application to OCR tasks often leads to suboptimal performance, or otherwise requires a substantial amount of additional fine-tuning.

\section{Method}

\subsection{Overall Framework}

LayoutLite is a lightweight plug-and-play module inserted between the vision encoder and the language model of a VLM, as illustrated in Fig.~\ref{fig:framework}. Given an input image, the vision encoder produces hierarchical visual representations across its layers. Let $\mathbf{H}_i \in \mathbb{R}^{N \times D \times M^2}$ denote the hidden states from the $i$-th layer, where $N$ is the number of visual tokens, $D$ is the hidden dimension, and $M$ is the Patch Merger factor. After the final Patch Merger, the encoder outputs the original visual tokens $\mathbf{V}_n \in \mathbb{R}^{N \times D}$, which serve as the standard visual input to the language model.

Instead of relying solely on the final-layer representations, LayoutLite leverages intermediate features from multiple layers. Specifically, we select a set of hidden states $\{\mathbf{H}_{i_1}, \ldots, \mathbf{H}_{i_d}\}$ and apply the same Patch Merger to obtain multi-level visual tokens:
$$
\mathbf{V} = \{\text{Merger}(\mathbf{H}_{i_k}) \mid k=1, \ldots, d\}, \quad \mathbf{V} \in \mathbb{R}^{d \times N \times D},
$$
which provide complementary information from different semantic depths. These tokens are fed into the LayoutLite module, which predicts a token importance score $\mathbf{S} \in [0,1]^{N}$ indicating the informativeness of each visual token.

Based on these scores, we adopt a cluster-based thresholding strategy to separate informative from low-information tokens under a target compression ratio. For the $i$-th image, we perform K-means clustering ($K=2$) on the predicted scores to obtain two cluster centers $l_i$ and $r_i$, and define the pruning threshold as
\[
\text{threshold}_i=l_i+\alpha(r_i-l_i),
\]
where $\alpha$ is a global parameter shared across all images and determined via binary search to satisfy the desired overall compression ratio. This yields a binary token mask $\mathbf{M} \in \{0,1\}^{N}$ per image, removing tokens in the low-score cluster. Compared with global thresholding, this strategy exploits each image's own score distribution, providing more robust and adjustable compression.

Finally, the mask is applied to $\mathbf{V}_n$ to obtain the compressed sequence $\mathbf{V}_{out}$, which is fed into the language model for autoregressive decoding. During MRoPE encoding, each retained token preserves its original spatial coordinates in $\mathbf{V}_n$, so the document's spatial structure is maintained after pruning.

\subsection{Structure of LayoutLite}

LayoutLite takes the multilayer visual tokens $\mathbf{V}\in\mathbb{R}^{d\times N\times D}$ as input and predicts an importance score for each token. The design is motivated by the observation that informative and non-informative tokens exhibit different feature evolution patterns across the layers of the vision encoder. If the final token $\mathbf{V}_{n}[i]$ carries meaningful content, its representation gradually evolves as information is aggregated through self-attention; for redundant regions such as blank backgrounds, the token receives little information from its neighbors and changes only slightly across layers. The feature trajectory $\{\mathbf{V}_{1}[i], \ldots, \mathbf{V}_{d}[i]\}$ thus provides useful cues for estimating the informativeness of token $i$.

Based on this, we model the layer-wise feature evolution with a lightweight 1D convolution. For each token, its representations across layers are treated as a short sequence along the depth dimension, and a Conv1D layer with kernel size $d$ captures the variation patterns, analogous to modeling temporal changes across frames. The output is passed through two fully connected layers, then normalized and mapped by a Sigmoid to a scalar score $\mathbf{S}\in[0,1]^N$ per token. The module is lightweight and interpretable: under Qwen3-VL with $D=1024$ and $d=4$, it has only about 19M parameters, roughly 1\% of the original model size.

We train LayoutLite within a reinforcement learning framework, as illustrated in Fig.~\ref{fig:framework}, keeping the underlying VLM frozen. We formulate token selection as a policy learning problem because it is inherently discrete and non-differentiable, has no ground-truth annotation of redundant tokens, and can only be evaluated \emph{after} an entire mask is applied and the frozen VLM produces its output. These properties make supervised fine-tuning inapplicable, as the only reliable signal—downstream OCR quality—is available solely at the sequence level. Rather than fine-tuning the language model to tolerate a fixed pruning pattern, we use a \emph{model-agnostic} method that identifies genuinely uninformative tokens; since it is optimized purely through the frozen model's own OCR behavior, it transfers across VLM architectures without retraining the VLM and recovers document layout structure as an emergent behavior.

During training, only LayoutLite's parameters are updated. For each image, the vision encoder is run once to extract the visual tokens and multi-layer hidden states, which are cached and reused across sampled masks to generate multiple completions whose OCR rewards optimize the policy. This caching makes training highly sample-efficient: for Qwen3-VL-2B, LayoutLite converges with only a few hundred unlabeled document images and about 20 GB of GPU memory on a single NVIDIA A100, within a few hours.

\subsubsection{Reinforcement Learning Method}

Given this formulation, we adopt Group Relative Policy Optimization (GRPO) \cite{deepseek-math} as the optimization algorithm. Its core idea—estimating advantages by comparing a \emph{group} of outputs sampled under the same input—aligns naturally with visual token pruning: for each image we sample a group of different masks, apply them to the same visual features, and compare their effects on the OCR output, pushing the policy toward the best-performing masks in the group. This is exactly what token pruning needs, since the informativeness of a token can only be judged by how removing it, together with many others, affects the final output. In contrast, PPO \cite{schulman2017proximal} optimizes each trajectory individually and cannot exploit such comparisons across masks, while DPO \cite{rafailov2023direct} relies on manually constructed preference pairs. GRPO instead optimizes directly from a scalar reward, which fits our task well: pruning quality is measured by the consistency between the pruned and original outputs via the Levenshtein ratio, so a reward is readily computable and no preference pairs are needed.

Specifically, for each input we sample a group of \(B\) pruning masks. Let \(mask_i\) denote the \(i\)-th mask with reward \(R_i\) and \(\mathbb{P}(mask_i)\) its probability under the policy. The rewards are normalized within the group as $\hat{R_i}=(R_i-\bar{R})/\sigma_R$, and the objective is
\[
\mathcal{L}_{GRPO}=
-\frac{1}{B}\sum_{i=1}^{B}\hat{R}_i\log \mathbb{P}(mask_i).
\]
This is a simplified form of GRPO that keeps its core principle; unlike the original GRPO for LLMs, we omit the additional KL penalty.

The reward measures both OCR quality and pruning stability. We use the Levenshtein ratio between the pruned and original outputs as the primary signal,
\[
R_{ocr} = \text{LevRatio}(Y_{pruned}, Y_{full}),
\]
where $Y_{pruned}$ and $Y_{full}$ are the pruned and original outputs. This score in $[0,1]$ reflects how well pruning preserves the original OCR capability, so maximizing it encourages LayoutLite to discard only redundant tokens. However, optimizing $R_{ocr}$ alone yields a degenerate solution that retains all tokens, achieving perfect similarity without compression. We therefore add a soft regularization term to constrain the discard ratio around a target value $a$, giving the final reward
\[
R = \text{LevRatio}(Y_{pruned}, Y_{full}) - \lambda|r-a|^m,
\]
where $r$ is the actual discard ratio, $a\in(0,1)$ the desired compression ratio, and $m>1$ controls the smoothness of the penalty. This keeps the compression budget roughly constant so the policy improves OCR preservation rather than trivially adjusting the discard ratio. Since $a$, $m$, and $\lambda$ act only as constraints against collapse and do not affect which tokens are deemed important, their exact values have little impact on final performance.

\begin{table*}[htbp]
    \caption{Main results on OmniDocBench. All reported values are averaged across the 1651 images in OmniDocBench.}
    \centering
    \begin{tabular}{cccccc}
    \hline
    & compression ratio & & & & \\
    Method          & (\%) & Edit distance & Formula CDM & Table TEDS & OmniDocBench Score \\ \hline
    \\
    FireRed-OCR & 0 & 0.044 & 94.315 & 88.343 & 92.753 \\
    \\
    & 5 & \textbf{0.043} & 94.469 & 88.258 & 92.809 \\
    & 10 & 0.044 & 94.317 & \textbf{88.344} & 92.754 \\
    & 15 & 0.045 & \textbf{94.734} & 88.257 & \textbf{92.830} \\
    & 20 & 0.045 & 94.539 & 88.176 & 92.738 \\
    & 25 & 0.045 & 94.550 & 87.759 & 92.603 \\
    & 30 & 0.045 & 93.941 & 87.724 & 92.388 \\
    FireRed-OCR & 35 & 0.046 & 93.764 & 87.859 & 92.341 \\
    + LayoutLite & 40 & 0.049 & 93.955 & 87.625 & 92.227 \\
    & 45 & 0.055 & 93.564 & 87.246 & 91.770 \\
    & 50 & 0.056 & 93.052 & 86.844 & 91.432 \\
    & 55 & 0.057 & 91.923 & 84.357 & 90.193 \\
    & 60 & 0.063 & 88.093 & 71.271 & 84.355 \\
    & 65 & 0.093 & 70.802 & 33.824 & 65.109 \\
    & 70 & 0.188 & 48.372 & 18.585 & 49.386 \\

    \hline
    \end{tabular}
\end{table*}

Given the reward $R$, LayoutLite is optimized with the GRPO objective. We randomly initialize its parameters and set the bias of the final layer $FC2$ so that all tokens are initially assigned score $a$. To encourage exploration, we generate masks by Bernoulli sampling: given the predicted scores $S$, each token is retained independently with probability
\[
\mathbb{P}(\text{mask}[i]=\mathrm{True})=S[i],
\]
so higher-scored tokens are more likely to be kept. By observing how different sampled masks affect OCR performance under the GRPO reward, LayoutLite gradually learns an implicit layout analysis strategy that preserves OCR-critical regions while discarding redundant ones.

\subsubsection{Layout Supervision Method}

Although LayoutLite can be trained solely through reinforcement learning, we further introduce an auxiliary layout supervision signal to accelerate convergence and improve performance. The key idea is to leverage existing document layout analysis models to provide coarse region-level guidance for token importance estimation.

We assume that regions enclosed by document layout bounding boxes are more likely to contain informative content than regions outside these boxes. This assumption is likewise justified by the experiment illustrated in Fig.~\ref{fig:model_output_comparison}. Therefore, tokens corresponding to text blocks, tables, formulas, and figures should generally receive higher importance scores than tokens corresponding to margins, backgrounds, or other non-content regions.

\begin{figure}[t]
    \centering
    \includegraphics[width=\linewidth]{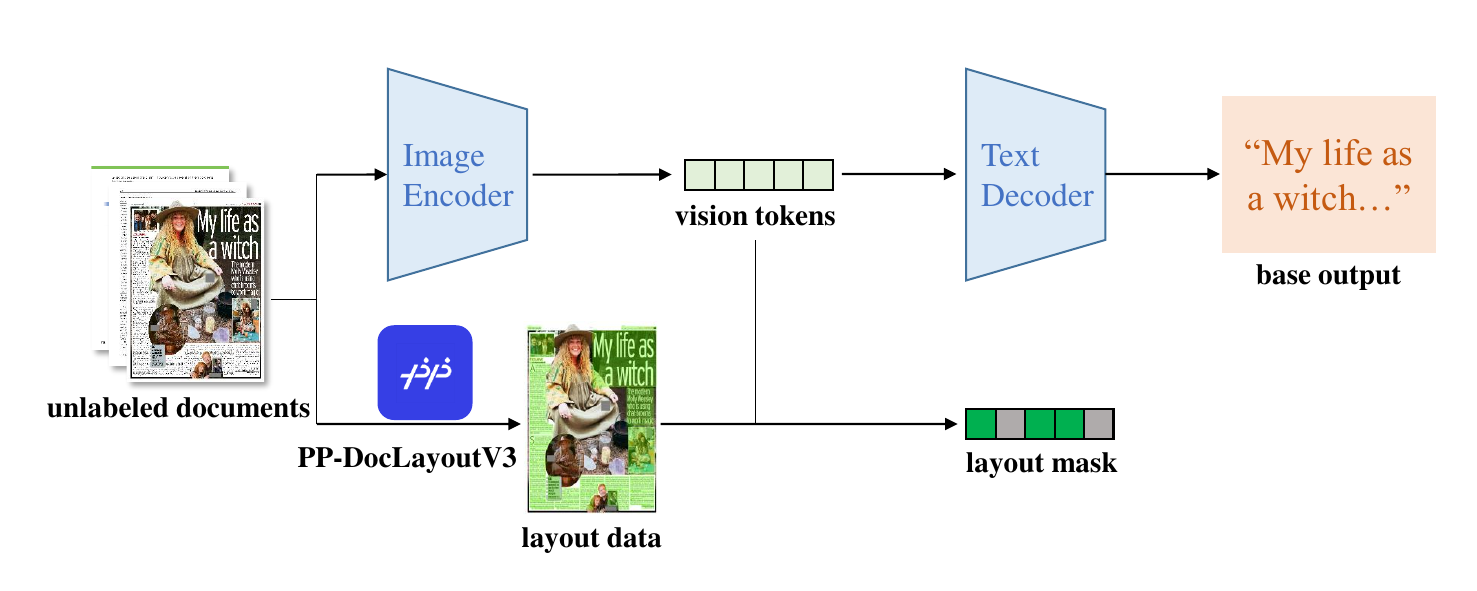}
    \caption{Basic output and layout mask generation procedure}
    \label{fig:dataproduce}
\end{figure}

To obtain layout supervision, we employ an off-the-shelf document layout analysis model PP-DocLayoutV3 \cite{cui2026rtdoclayoutrealtimeendtoenddocument} to generate bounding boxes for each training image, as shown in Fig.~\ref{fig:dataproduce}. The detected layout regions are then projected onto the visual token grid, producing a binary layout mask

\[
\mathbf{M}_{layout} \in \{0,1\}^{N},
\]

where a value of 1 indicates that the corresponding visual token completely or partially falls inside a detected layout region.

Given the token importance scores \(\mathbf{S}\), we encourage tokens within layout regions to receive higher scores than those outside the regions. Specifically, we define the layout supervision loss as

\[
\mathcal{L}_{layout}=
\frac{1}{|\Omega_{out}|}
\sum_{j\in\Omega_{out}} S_j
-
\frac{1}{|\Omega_{in}|}
\sum_{j\in\Omega_{in}} S_j,
\]

where \(\Omega_{in}\) and \(\Omega_{out}\) denote the sets of tokens inside and outside the layout regions, respectively. 

Minimizing \(loss_{layout}\) increases the score gap between content and non-content regions, encouraging LayoutLite to learn document-aware token selection patterns. Since the supervision is derived from automatically generated layout annotations rather than human labeling, the additional training cost is negligible. Furthermore, the layout loss serves only as an auxiliary optimization objective and does not constrain the model to exactly reproduce the output of the layout detector. Instead, it provides a useful initialization signal that complements the reinforcement learning objective and guides the model toward more effective implicit layout analysis.

The total loss function is defined as follow:

\[
\mathcal{L}_{total}=\mathcal{L}_{GRPO}+\mathcal{L}_{layout}
\]

\begin{table*}[htbp]
\caption{Efficiency analysis on OmniDocBench. We report the average Prefill Latency, FLOPs and KV Cache, along with latency and FLOPs brought by LayoutLite, measured over inference on all 1,651 images in OmniDocBench. All reported values are averaged across the entire benchmark.}
\centering
\begin{tabular}{ccccccc}
\hline
& compression ratio & Prefill Lat. & LayoutLite Lat. & Prefill FLOPs & LayoutLite FLOPs & KV Cache \\
Method & (\%) & (ms) & (ms) & (TFLOPs) & (TFLOPs) & (MB) \\
\hline
                    \\
FireRed-OCR         & 0 &  123.738 & -     & 21.260 & -     & 512.132 \\
                    \\
                    & 10 & 110.720 &       & 19.159 &       & 463.484 \\
                    & 20 & 99.617  &       & 17.287 &       & 418.267 \\
FireRed-OCR         & 30 & 88.446  &       & 15.520 &       & 373.735 \\
+ LayoutLite        & 40 & 78.141  & 2.118 & 13.833 & 0.061 & 329.311 \\
                    & 50 & 68.229  &       & 12.192 &       & 284.040 \\
                    & 60 & 58.400  &       & 10.619 &       & 238.456 \\
                    & 70 & 46.849  &       & 8.902  &       & 185.835 \\          

\hline
\end{tabular}
\end{table*}

\section{Experiments}

\subsection{Experimental setup}

\subsubsection{Model architectures}

We primarily evaluate LayoutLite on FireRed-OCR, a representative end-to-end OCR model obtained by fine-tuning Qwen3-VL-2B without additional components, which makes it a clean and unified VLM for evaluating visual token pruning. Its vision encoder has 24 layers with hidden dimension 1024, from which we uniformly extract hidden states at the 6th, 12th, 18th, and 24th (output) layers as the multi-layer visual features. Unless otherwise specified, LayoutLite uses a unified architecture with intermediate dimension 2048, GELU activation, and a depthwise 1D convolution of kernel size 4.

\subsubsection{Training setups}

We train on the Document Parsing subset of OCRBench \cite{Liu_2024}, which contains 600 document images covering diverse layouts and real-world scenarios. For each image, we obtain a base output by running the original OCR model and generate layout annotations with PP-DocLayoutV3 as auxiliary supervision. We adopt a fixed 50\% compression ratio and a GRPO batch size of 5, and train for a single epoch on one NVIDIA A100 GPU, requiring about 25\% of its memory and a few hours.

\subsubsection{Evaluation setups}

We evaluate LayoutLite on OmniDocBench v1.7 \cite{ouyang2024omnidocbenchbenchmarkingdiversepdf}, a mainstream benchmark for document parsing in real-world scenarios, under different visual token compression ratios. Besides the overall score, we report fine-grained metrics: Edit Distance, TEDS \cite{zhong2020imagebasedtablerecognitiondata} for table recognition, and CDM \cite{wang2025imagetexttransformingformula} for formula recognition, giving a comprehensive view of how token pruning affects different document elements. We also evaluated the performance of FastV and PixelPrune for comparison. Their results and details can be found in the appendix.

All scores are reproduced by us under a unified protocol rather than taken from the original papers. To ensure deterministic and reproducible results, we adopt greedy decoding for FireRed-OCR instead of the temperature-based sampling used in their official configurations, which removes sampling randomness and guarantees a fair, consistent comparison across compression ratios. Our reproduced scores may therefore differ slightly from the officially reported numbers.

\subsection{Main Results}
\begin{figure}[t]
    \centering
    \setlength{\tabcolsep}{3pt}
    \begin{tabular}{
        >{\centering\arraybackslash}m{0.47\linewidth}
        >{\centering\arraybackslash}m{0.47\linewidth}
        }
        Heatmap & Mask \\
        \includegraphics[width=0.49\linewidth]{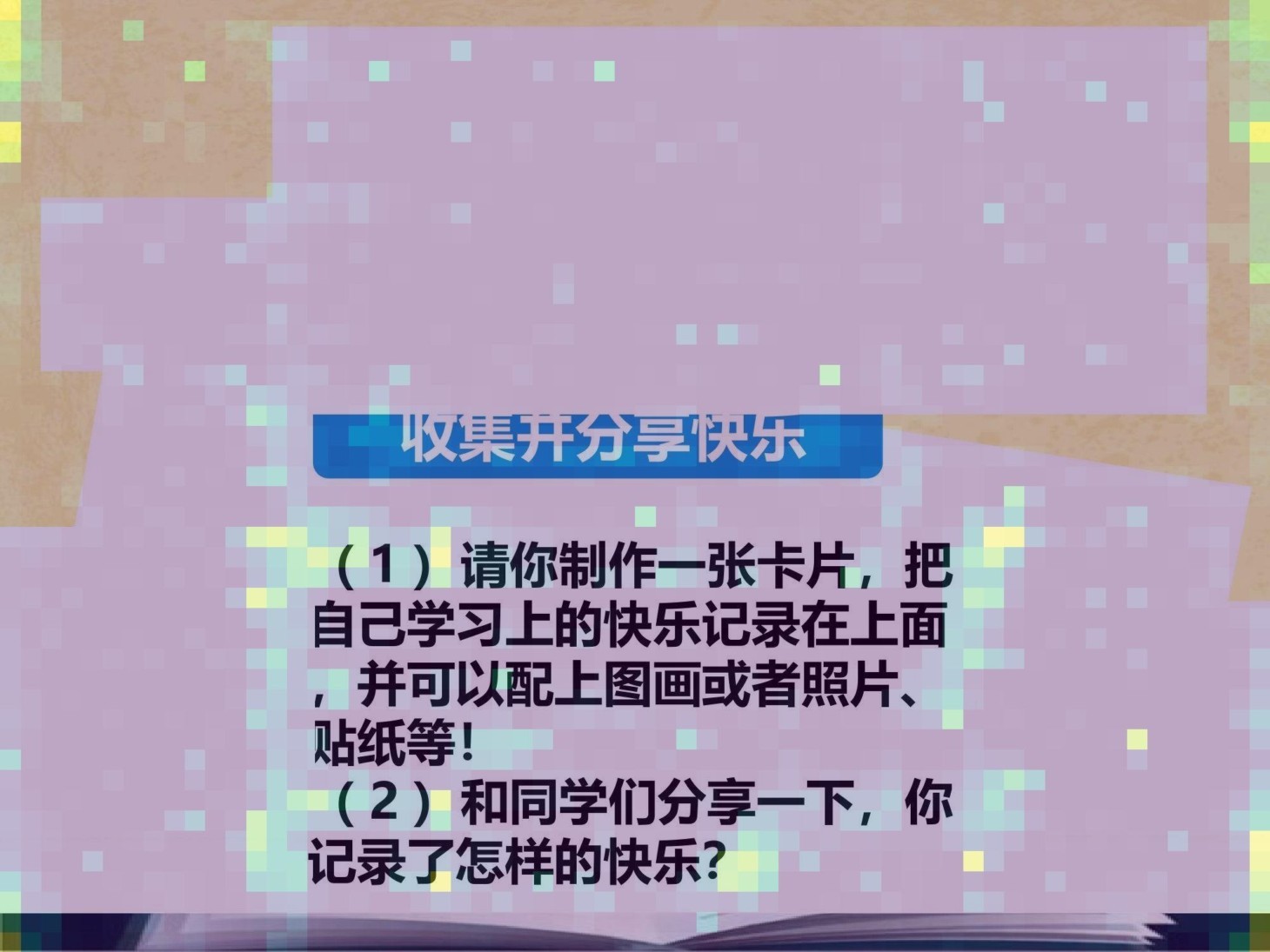}
        \includegraphics[width=0.49\linewidth]{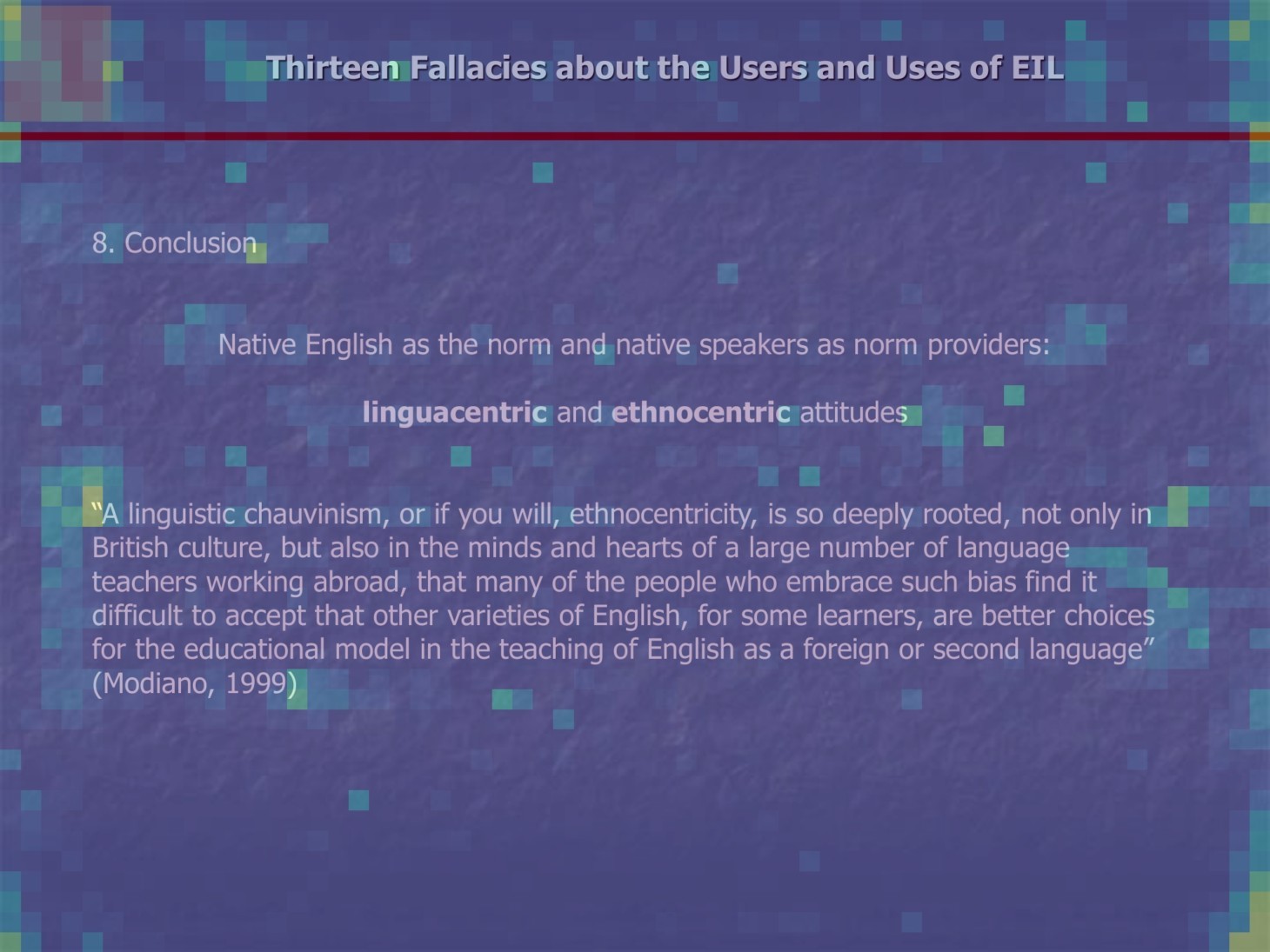}
        &
        \includegraphics[width=0.49\linewidth]{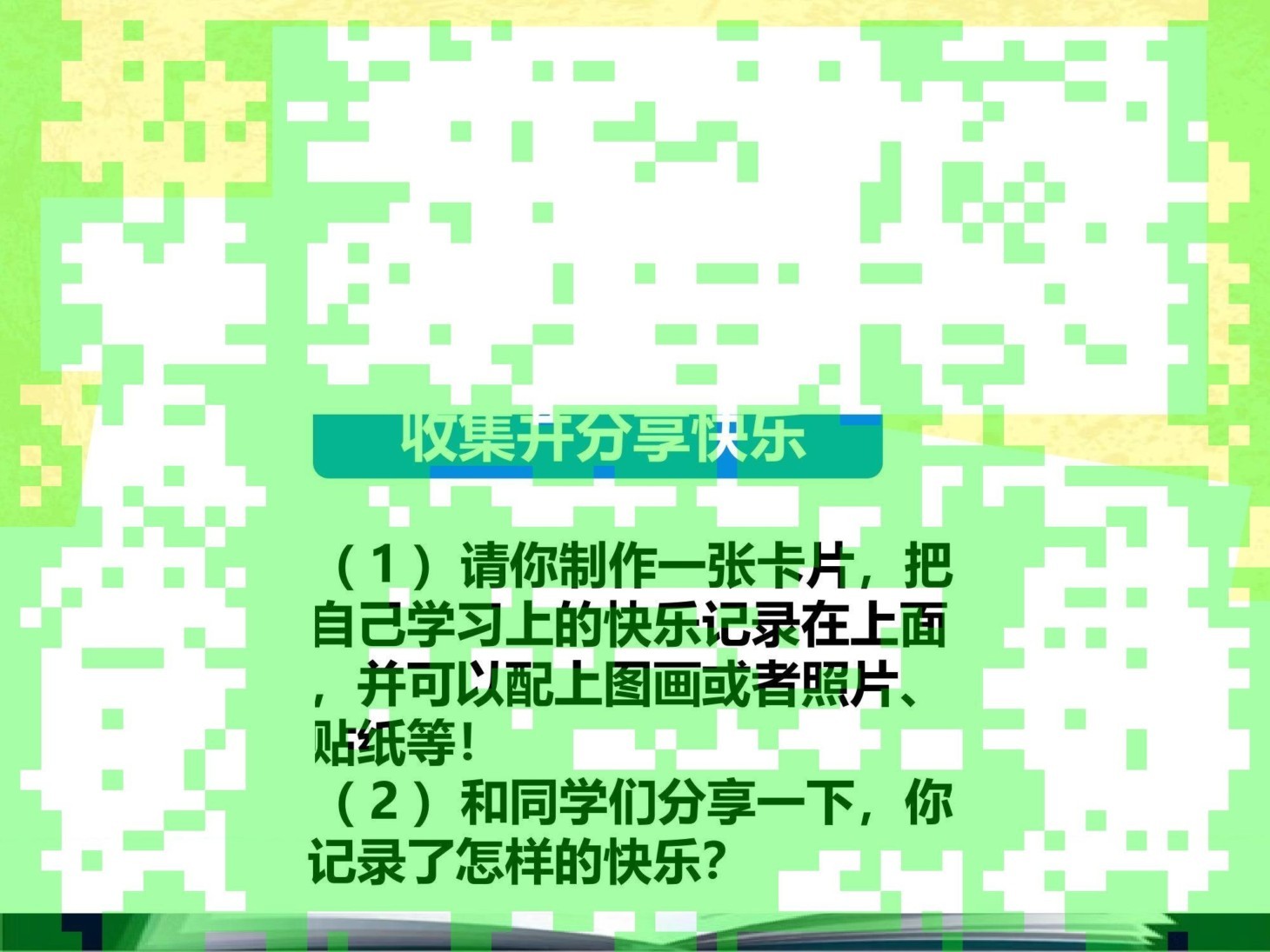}
        \includegraphics[width=0.49\linewidth]{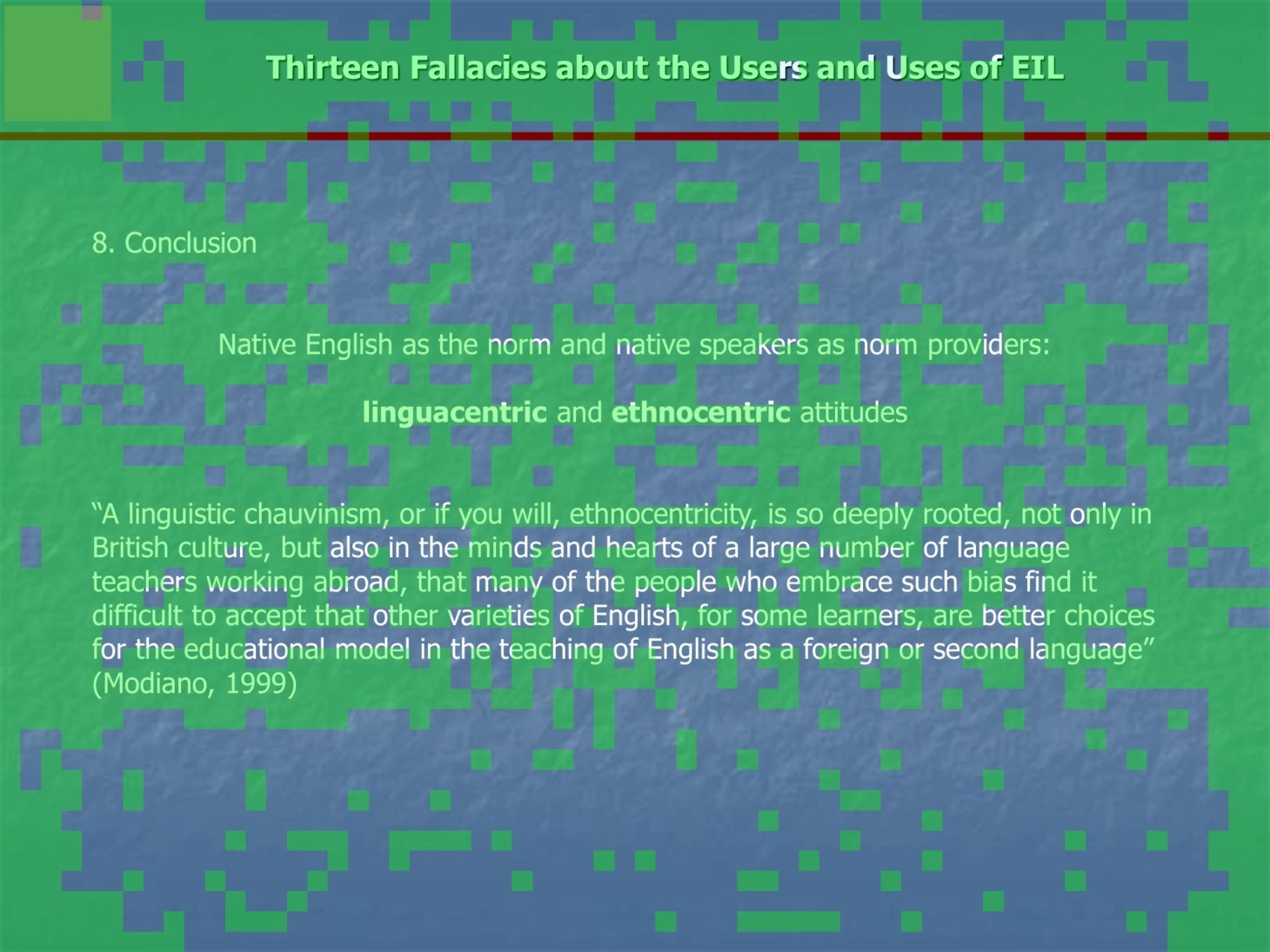}
        \\
        \multicolumn{2}{c}{\small FastV} \\
        -- &
        \includegraphics[width=0.49\linewidth]{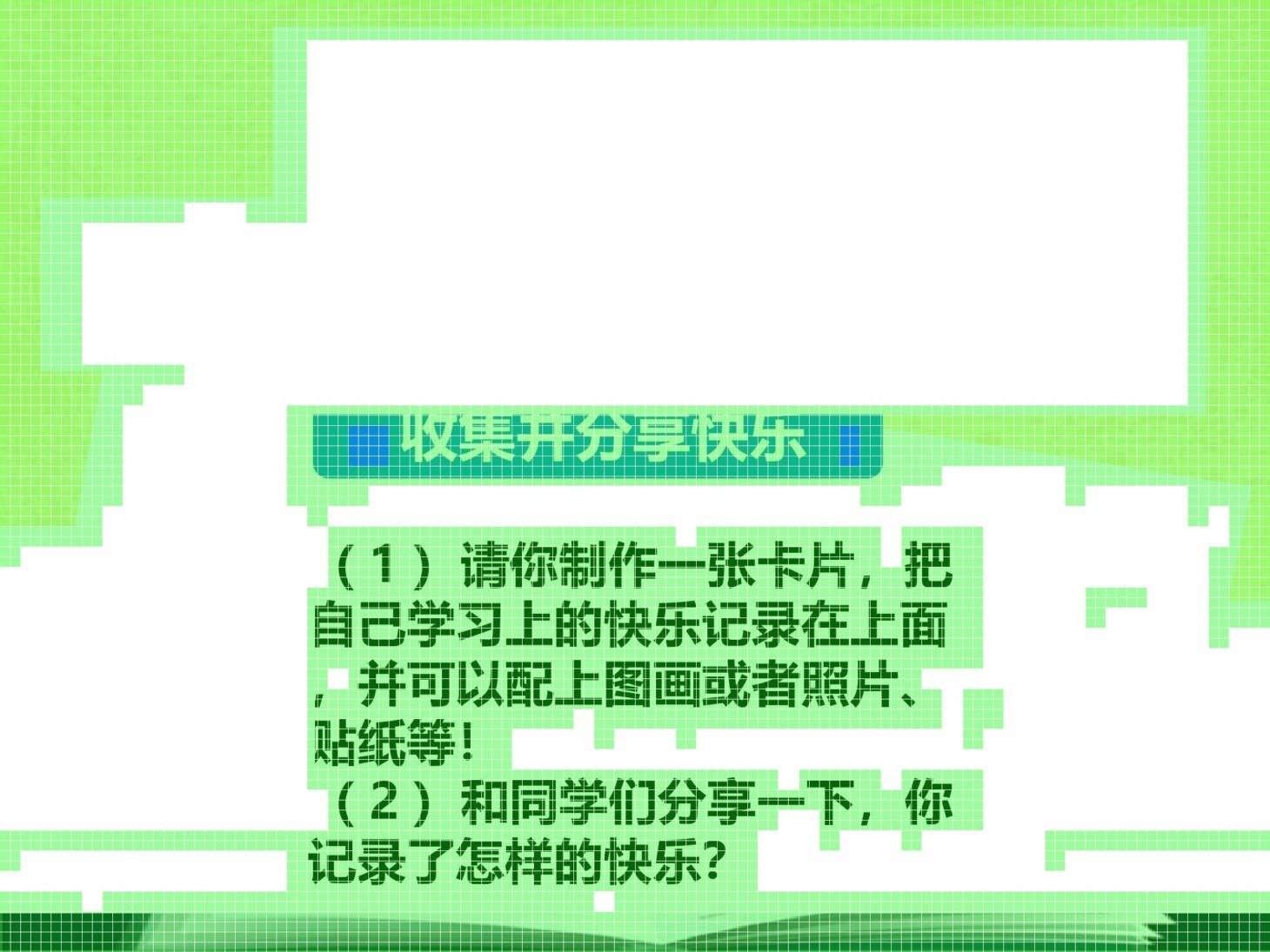}
        \includegraphics[width=0.49\linewidth]{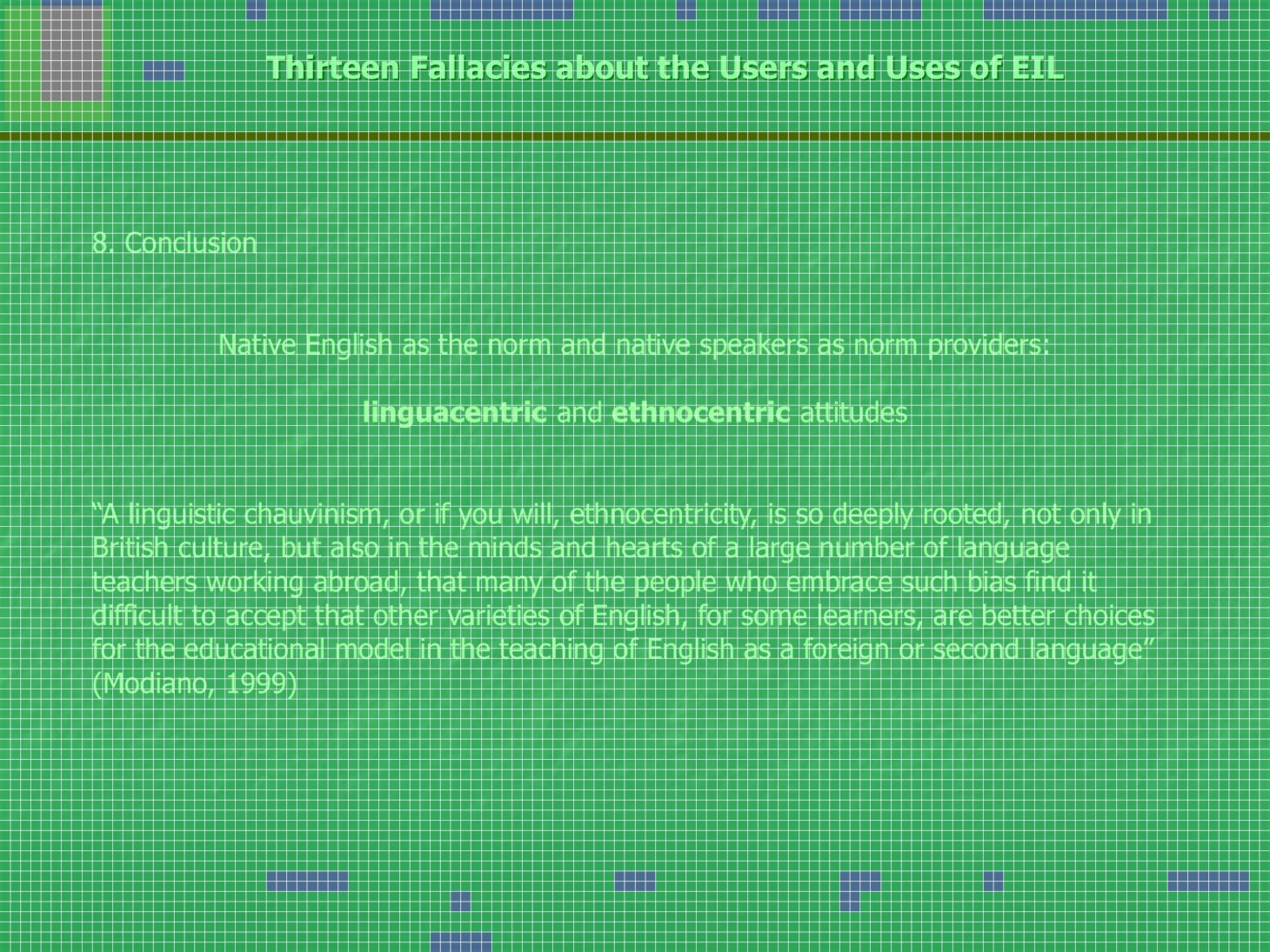}
        \\
        \multicolumn{2}{c}{\small PixelPrune} \\
        \includegraphics[width=0.49\linewidth]{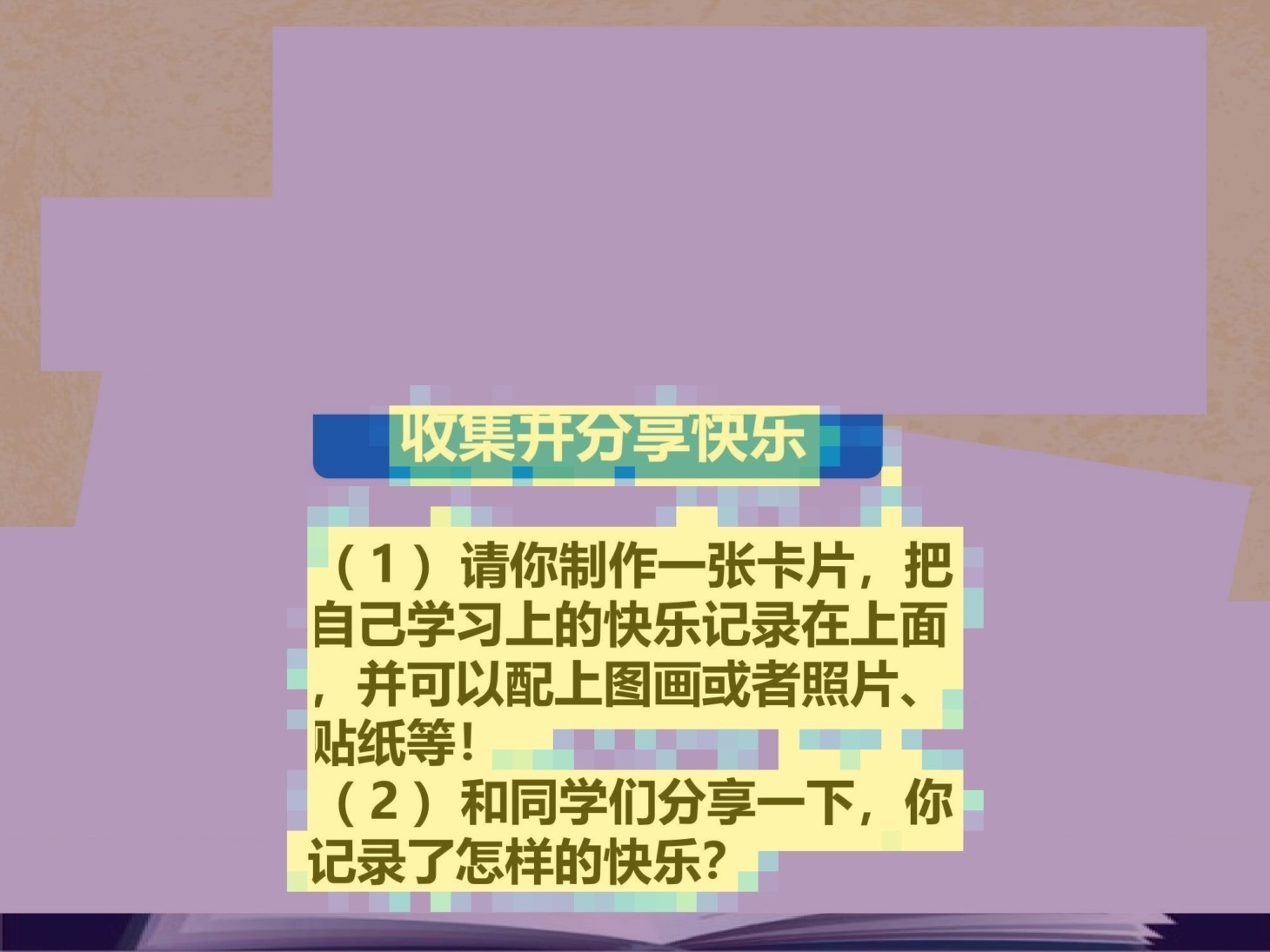}
        \includegraphics[width=0.49\linewidth]{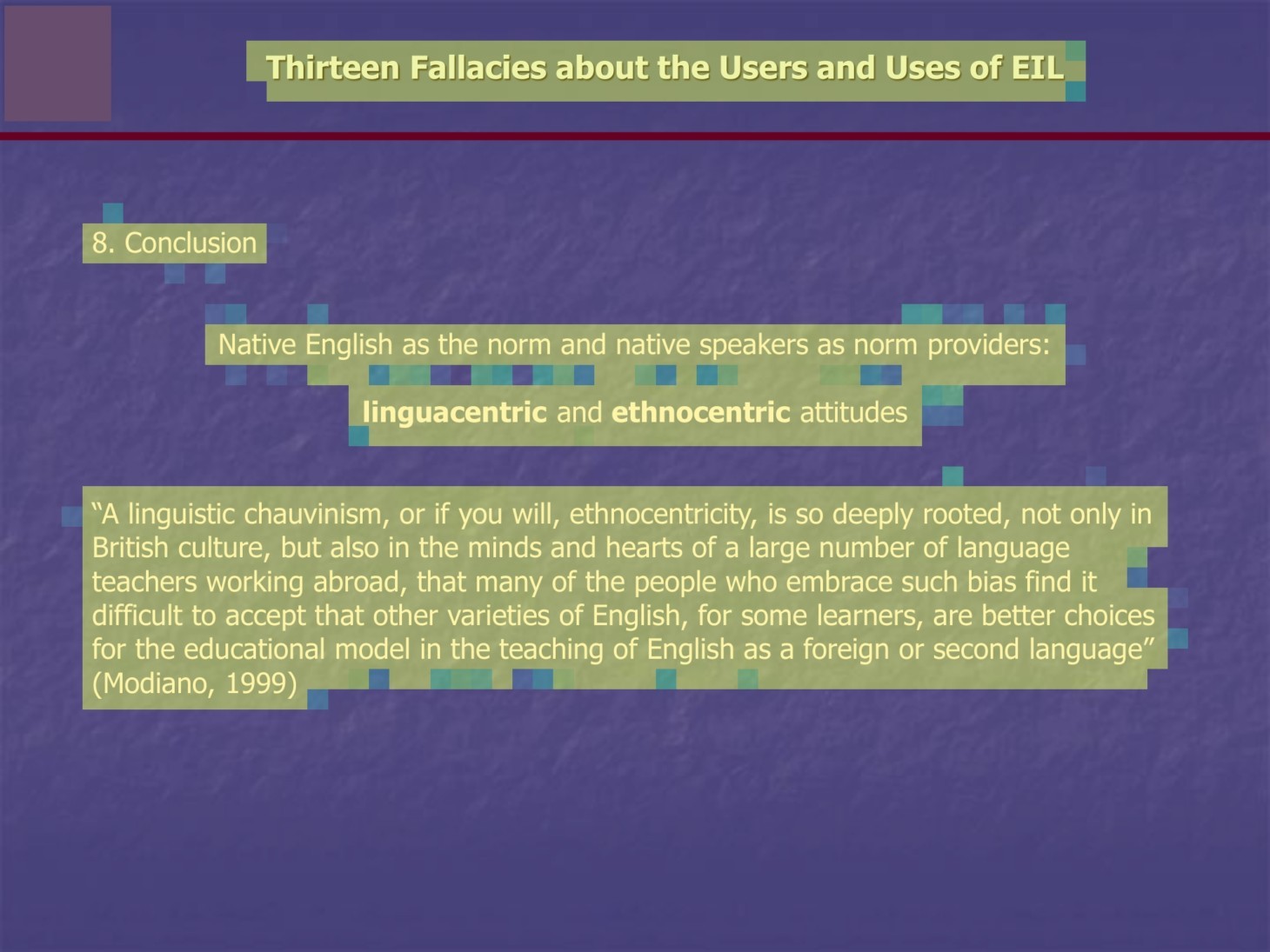}
        &
        \includegraphics[width=0.49\linewidth]{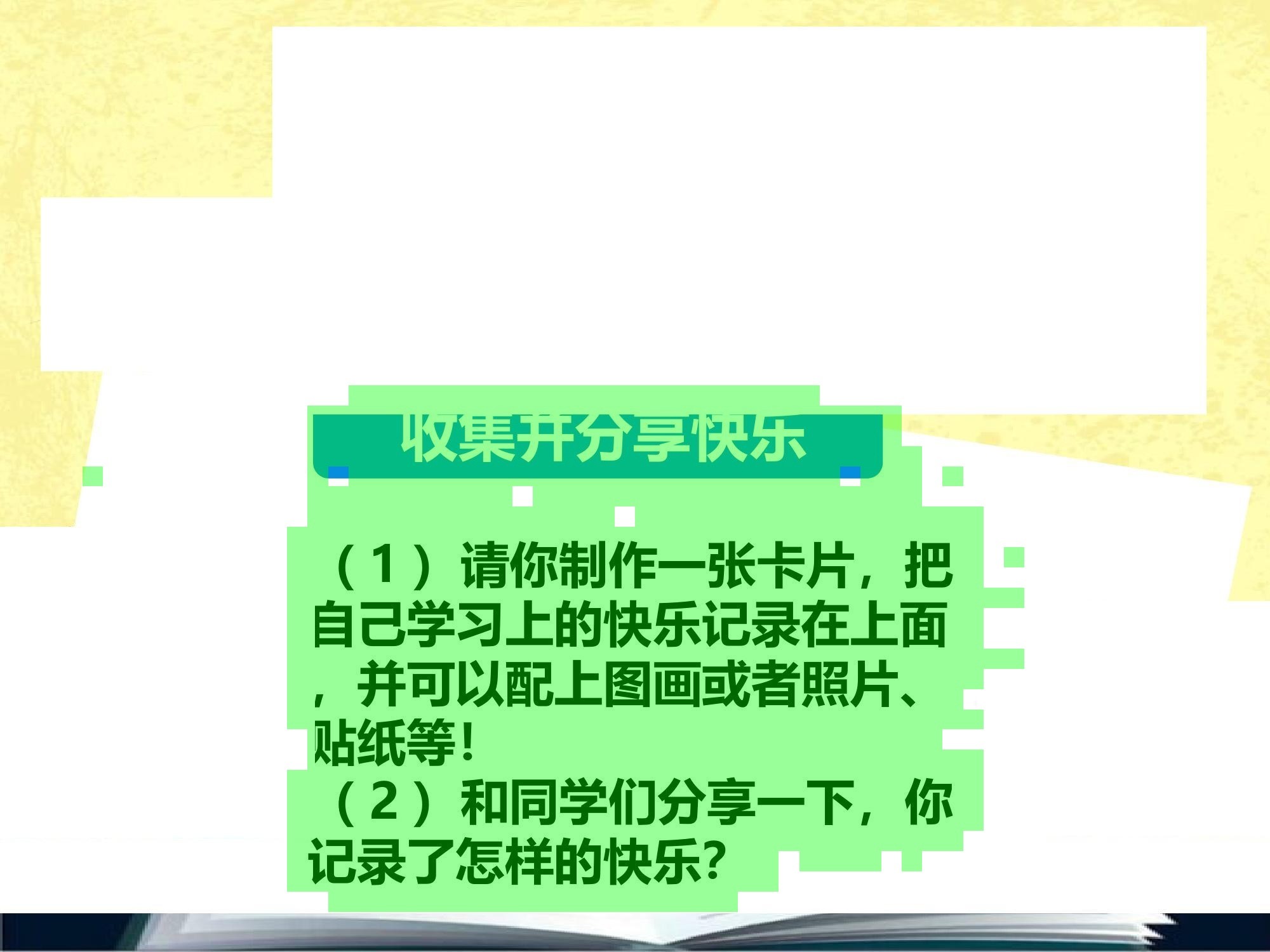}
        \includegraphics[width=0.49\linewidth]{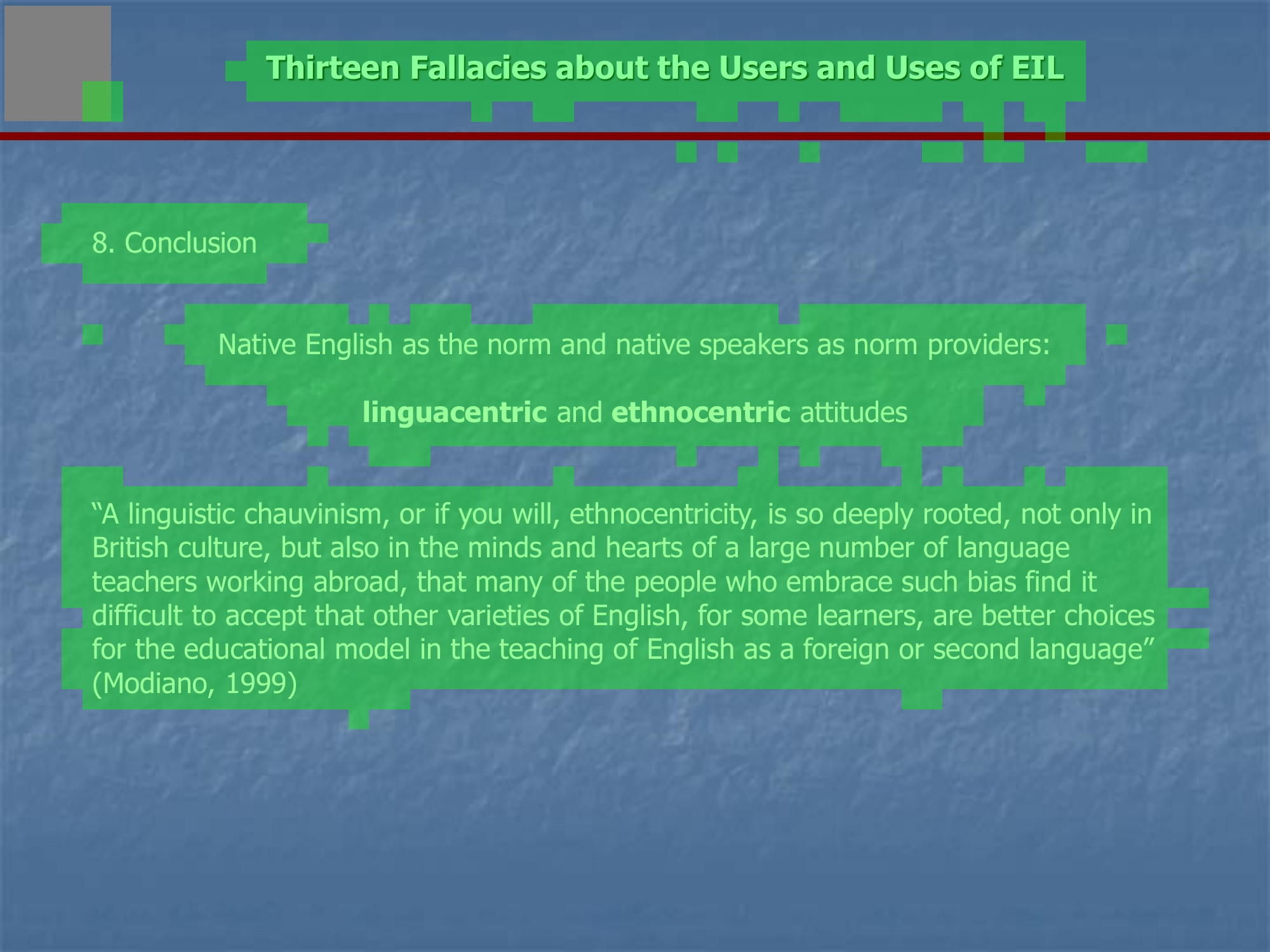}
        \\
        \multicolumn{2}{c}{\small Ours} \\
    \end{tabular}
    \caption{Visualization comparison of visual token pruning methods. Heatmaps: yellow/purple indicate high/low-score tokens. Masks: green regions are preserved tokens. Additional visualization can be found in the appendix.}
    \label{fig:visualization}
\end{figure}
\begin{table*}[htbp]
    \caption{Ablation studies on the effects of model architectures. We report Logics-Parsing-V2 results on OmniDocBench.}
    \centering
    \begin{tabular}{cccccc}
    \hline
& Compression & OmniDocBench \\
Method          & rate (\%)  & Score & Prefill Lat. & Prefill FLOPs & KV Cache \\ \hline
\\
Logics-Parsing-V2 & 0 & 92.819 & 257.137 & 47.638 & 617.532 \\
\\
& 10 & 92.069 & 226.624 & 41.965 & 551.904 \\
& 20 & 92.086 & 202.883 & 37.091 & 492.988 \\
Logics-Parsing-V2 & 30 & 91.595 & 178.034 & 32.647 & 436.941 \\
+ LayoutLite & 40 & 91.092 & 154.809 & 28.409 & 381.146 \\
& 50 & 90.609 & 132.802 & 24.302 & 324.572 \\
& 60 & 85.751 & 110.181 & 20.215 & 265.482 \\
& 70 & 59.424 & 84.154  & 15.855 & 198.815 \\
    \hline
    \end{tabular}
\end{table*}

Table 1 reports the performance of LayoutLite under different compression ratios on OmniDocBench. LayoutLite removes a substantial amount of redundant visual tokens while maintaining OCR performance. Below 20\% compression, the overall score stays virtually unchanged from the baseline (92.75) and even improves slightly, which we attribute to the suppression of repetitive generation observed in a few samples. Even at 50\% compression, LayoutLite scores 91.43, only a 1.32-point drop while removing half of the visual tokens.

The individual metrics follow the same trend. Text recognition stays highly stable, with edit distance rising only from 0.044 to 0.056 at 50\% compression, and formula recognition is robust, with CDM decreasing only from 94.32 to 93.05. Table recognition is more sensitive, yet TEDS remains as high as 86.84 after discarding half the tokens. A noticeable drop appears only beyond 55\% compression, suggesting most redundant information has already been removed by that point.

To understand why LayoutLite remains accurate under aggressive compression, we visualize the token selection of different methods in Fig.~\ref{fig:visualization}, all under a 50\% average compression ratio. FastV produces coarse importance maps and struggles to separate redundant backgrounds from informative content, while PixelPrune operates on image patches rather than visual tokens and, despite preserving text in simple cases, is less effective on documents with complex backgrounds. In contrast, LayoutLite generates more structured and accurate importance maps, consistently assigning low scores to blank margins and redundant regions while preserving text-rich areas. It therefore removes substantially more redundant tokens without discarding OCR-critical information, explaining its strong performance across compression ratios.

\subsection{Efficiency analysis}

LayoutLite significantly reduces the number of visual tokens processed by the LLM, lowering inference latency and memory consumption. Since it is inserted between the vision encoder and the language model, it mainly shortens the visual prefix during the prefill stage, so we focus on prefill-stage metrics rather than end-to-end decoding latency. As shown in Table 2, the efficiency gains grow with the compression ratio, while the overhead introduced by LayoutLite itself is negligible. Prefill latency, FLOPs, and KV cache memory all scale approximately linearly with the number of retained tokens: every additional 5\% token reduction yields about a 4--5\% decrease in each, showing that LayoutLite effectively translates token compression into practical inference and memory savings.

\subsection{Ablation studies}

\subsubsection{Effects of model architectures}
To demonstrate the scalability of our method, we further evaluate LayoutLite on Logics-Parsing-V2, another representative end-to-end OCR model. As shown in Table 3 and Fig. 1, LayoutLite again preserves strong OCR performance under practical compression ratios while substantially reducing inference latency, FLOPs, and KV cache consumption, with negligible overhead from LayoutLite itself. This confirms the effectiveness and generalization of LayoutLite across different OCR architectures.

\begin{table}[htbp]
    \caption{Ablation study on the effect of layout supervision.}
    \centering
    \begin{tabular}{
    >{\centering\arraybackslash}m{2cm}
    >{\centering\arraybackslash}m{1.6cm}
    >{\centering\arraybackslash}m{1cm}
    >{\centering\arraybackslash}m{2.3cm}
    }
    \hline
Method & compression ratio (\%) & GRPO & GRPO + Layout supervision \\ \hline
\\
FireRed-OCR & 0 & \multicolumn{2}{l}{\qquad\quad 92.753} \\
\\
& 10 & 92.149 & \textbf{92.754} \\
& 20 & 91.961 & 92.738 \\
& 30 & 90.903 & 92.388 \\
FireRed-OCR & 40 & 89.727 & 92.227 \\
+ LayoutLite & 50 & 85.862 & 91.432 \\
& 60 & 74.346 & 84.355 \\
& 70 & 48.795 & 49.386 \\
    \hline
    \end{tabular}
\end{table}

\subsubsection{Effects of Layout Supervision}
To evaluate the contribution of the proposed layout supervision, we compare training with the GRPO objective alone against the combination of GRPO and layout supervision under different compression ratio. As shown in Table 4, layout supervision consistently improves OCR performance across almost the entire compression range.

At practical compression ratio, incorporating layout supervision substantially narrows the performance gap with the uncompressed baseline. For example, at compression ratio of 20\%, 30\%, and 50\%, the OmniDocBench score increases from 91.961 to 92.738 (+0.777), from 90.903 to 92.388 (+1.485), and from 85.862 to 91.432 (+5.570), respectively. 

Overall, these results demonstrate that layout supervision is a key component of LayoutLite, enabling stable and high-accuracy visual token compression over a wide range of compression ratio.

\begin{table}[htbp]
    \caption{Ablation study on the effect of cluster-based threshold.}
    \centering
    \begin{tabular}{@{}lccc@{}}
    \hline
    \multicolumn{1}{c}{Method} &
    \makecell[c]{Compression\\rate (\%)} &
    \makecell[c]{Global\\threshold} &
    \makecell[c]{Cluster-based\\threshold} \\
    \hline
\\
FireRed-OCR & 0 & \multicolumn{2}{c}{92.753} \\
\\
& 5 & 92.750 & 92.809 \\
& 10 & 92.709 & 92.754 \\
& 15 & 92.648 & \textbf{92.830} \\
FireRed-OCR & 20 & 92.577 & 92.738 \\
+ LayoutLite & 25 & 92.427 & 92.603 \\
& 30 & 92.277 & 92.388 \\
& 35 & 92.398 & 92.341 \\
& 40 & 92.269 & 92.227 \\
    \hline
    \end{tabular}
\end{table}

\subsubsection{Effects of cluster-based threshold}

To investigate the effectiveness of the proposed cluster-based thresholding strategy, we compare it with a global thresholding strategy, where a single score threshold is determined over all visual tokens and applied uniformly to every image. As shown in Table~5, the proposed cluster-based thresholding strategy consistently achieves comparable or better performance across the practical compression ratios. Compared with the global thresholding strategy, it provides a higher performance upper bound and maintains more stable OCR accuracy over a wide range of compression ratios.

\section{Conclusion}

In this paper, we proposed LayoutLite, a lightweight and plug-and-play visual token compression module for efficient VLM-based document OCR. Instead of relying on explicit layout detection, LayoutLite performs implicit layout analysis at the token level: it scores the informativeness of each visual token and removes redundant ones while preserving OCR-critical information. Trained with a reinforcement learning objective and auxiliary layout supervision, LayoutLite requires only a small amount of unlabeled data and leaves the VLM model entirely frozen. Extensive experiments on multiple OCR architectures demonstrate that LayoutLite substantially reduces visual token length and inference cost with negligible performance degradation under practical compression ratios, providing an effective and practical solution for accelerating end-to-end document OCR.

\bibliography{aaai2027}

\newpage
\twocolumn

\section{Appendix}

\subsection{Training Details}

All experiments are conducted on a single A100-80GB GPU. The hyperparameters corresponding to the best-performing checkpoint used throughout our experiments are listed in Table~\ref{tab:training_details}. The hyperparameters $a$, $m$, and $\lambda$ correspond to the reward function described in the main paper.

\begin{table}[htbp]
    \caption{Training arguments for LayoutLite}
    \label{tab:training_details}
    \centering
    \begin{tabular}{cccc}
    \hline
    Key & Value & Key & Value \\
    \hline
    optimizer & Adam & dtype & bfloat16 \\
    batch size & 5 & training data size & 600 \\
    learning rate & 1e-3 & epoch & 1.0 \\
    betas & (0.9, 0.999) & $a$ & 0.5 \\
    epsilon & 1e-8 & $m$ & 1.5 \\
    weight decay & 0 & $\lambda$ & 5.0 \\
    \hline
    \end{tabular}
\end{table}

\subsection{FastV Implementation Details}

The official implementation of FastV was developed for LLaVA-style architectures and does not support Qwen3-VL. We reimplemented FastV by modifying the source code of Qwen3-VL following the pruning strategy described in the original paper.

Specifically, we apply FastV during the prefill stage and set $K=2$, such that visual token pruning is performed after the second transformer layer, following the default configuration suggested in the original FastV implementation. We vary $R$ from 0 to 0.6 to evaluate model performance under different visual token compression ratios, where $R$ denotes the proportion of visual tokens removed. The resulting performance under different compression ratios is reported in Table~\ref{tab:fastv}.

\begin{table}[htbp]
    \caption{Main results of FastV on OmniDocBench.}
    \label{tab:fastv}
    \centering
    \begin{tabular}{ccc}
    \hline
    & Compression rate & \\
    Method & (\%) & OmniDocBench Score \\ \hline
    \\
    FireRed-OCR & 0 & 92.753 \\
    \\
    & 10 & 89.356 \\
    & 20 & 86.909 \\
    FireRed-OCR & 30 & 82.352 \\
    + FastV & 40 & 74.792 \\
    & 50 & 63.435 \\
    & 60 & 51.654 \\
    \hline
    \end{tabular}
\end{table}

\subsection{PixelPrune Implementation Details}

PixelPrune provides an official implementation interface for Qwen3-VL. Therefore, we directly integrate PixelPrune through its provided \texttt{apply\_pixelprune} function before loading the OCR model. PixelPrune achieves an average visual token compression rate of 43.92\% and obtains an OmniDocBench score of 86.88.

\subsection{Model Architectures Details}

We evaluate LayoutLite on FireRed-OCR and Logics-Parsing-V2. Both models adopt the Qwen3-VL architecture with a 24-layer vision encoder, while differing in the configuration of the language model. Following the DeepStack design, LayoutLite aggregates multi-level visual features from the 6th, 12th, 18th, and 24th vision layers. The detailed architecture configurations are summarized in Table~\ref{tab:model_architecture}.

\subsection{FLOPs Analysis Details}

Following PixelPrune, the total theoretical FLOPs are computed as

\begin{equation}
\begin{aligned}
\mathrm{FLOPs}_{\mathrm{total}}
=&\ L_v \left[
N\left(8D_v^2 + 4D_vD_{f,v}\right)
+4N^2D_v
\right] \\
&+L_m\frac{N}{M^2}
\left(
2D_{\mathrm{in}}^2
+2D_{\mathrm{in}}D_l
\right) \\
&+L_l\left(
N_lC_l
+4N_l^2D_l
\right),
\end{aligned}
\end{equation}

where $N$ is the number of visual tokens before pruning, $L_v$ is the number of vision transformer layers, $D_v$ and $D_{f,v}$ denote the hidden size and FFN dimension of the vision encoder, respectively, $M=2$ is the spatial merge factor, and $D_{\mathrm{in}}=M^2D_v$ is the input dimension of the Patch Merger. $L_m=1+|\mathcal{I}_{\mathrm{DS}}|$ is the number of merger modules, where $\mathcal{I}_{\mathrm{DS}}$ denotes the selected DeepStack visual layers. $L_l$ is the number of LLM layers, $N_l=N/M^2+T$ is the LLM sequence length with $T$ text tokens, $D_l$ is the LLM hidden size, and $C_l$ denotes the per-token linear computation cost of the LLM. All formulas follow the FLOPs estimation in PixelPrune.

\begin{table}[t]
    \caption{Details of the model architectures used in our experiments.}
    \label{tab:model_architecture}
    \centering
    \resizebox{\columnwidth}{!}{
    \begin{tabular}{lcc}
        \toprule
        {Architecture} 
        & {FireRed-OCR} 
        & {Logics-Parsing-V2} \\
        \midrule

        Num visual tokens
        & Dynamic
        & Dynamic \\

        Num visual layers
        & 24
        & 24 \\

        Visual feature size
        & 1024
        & 1024 \\

        Num visual attention heads
        & 16
        & 16 \\

        Visual intermediate size
        & 4096
        & 4096 \\

        Visual patch size
        & 16
        & 16 \\

        Spatial merge size
        & 2
        & 2 \\

        Num LLM layers
        & 28
        & 36 \\

        Num LLM attention heads
        & 16
        & 32 \\

        Num LLM key-value heads
        & 8
        & 8 \\

        LLM hidden size
        & 2048
        & 2560 \\

        LLM intermediate size
        & 6144
        & 9728 \\

        Selected visual layers
        & [6, 12, 18, 24]
        & [6, 12, 18, 24] \\

        LayoutLite input shape
        & $4 \times N \times 1024$
        & $4 \times N \times 1024$ \\

        LayoutLite hidden size
        & 2048
        & 2048 \\

        LayoutLite convolution kernel
        & 4
        & 4 \\

        LayoutLite output size
        & 1
        & 1 \\

        \bottomrule
    \end{tabular}
    }
\end{table}

\subsection{Visualization}

\begin{figure*}[t]
\centering

\begin{subfigure}[t]{0.16\linewidth}
    \centering
    \includegraphics[width=\linewidth]{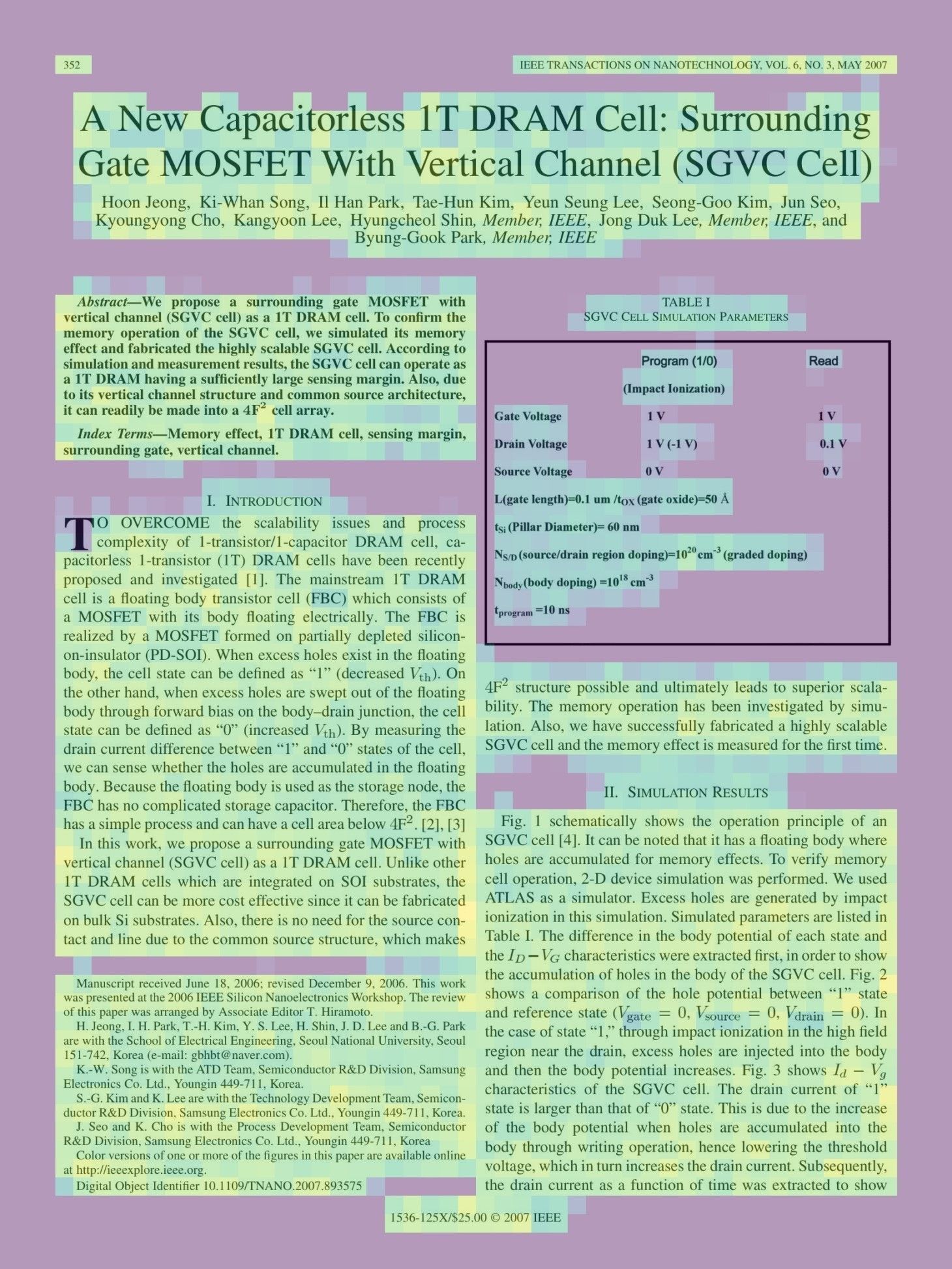}
\end{subfigure}
\hfill
\begin{subfigure}[t]{0.16\linewidth}
    \centering
    \includegraphics[width=\linewidth]{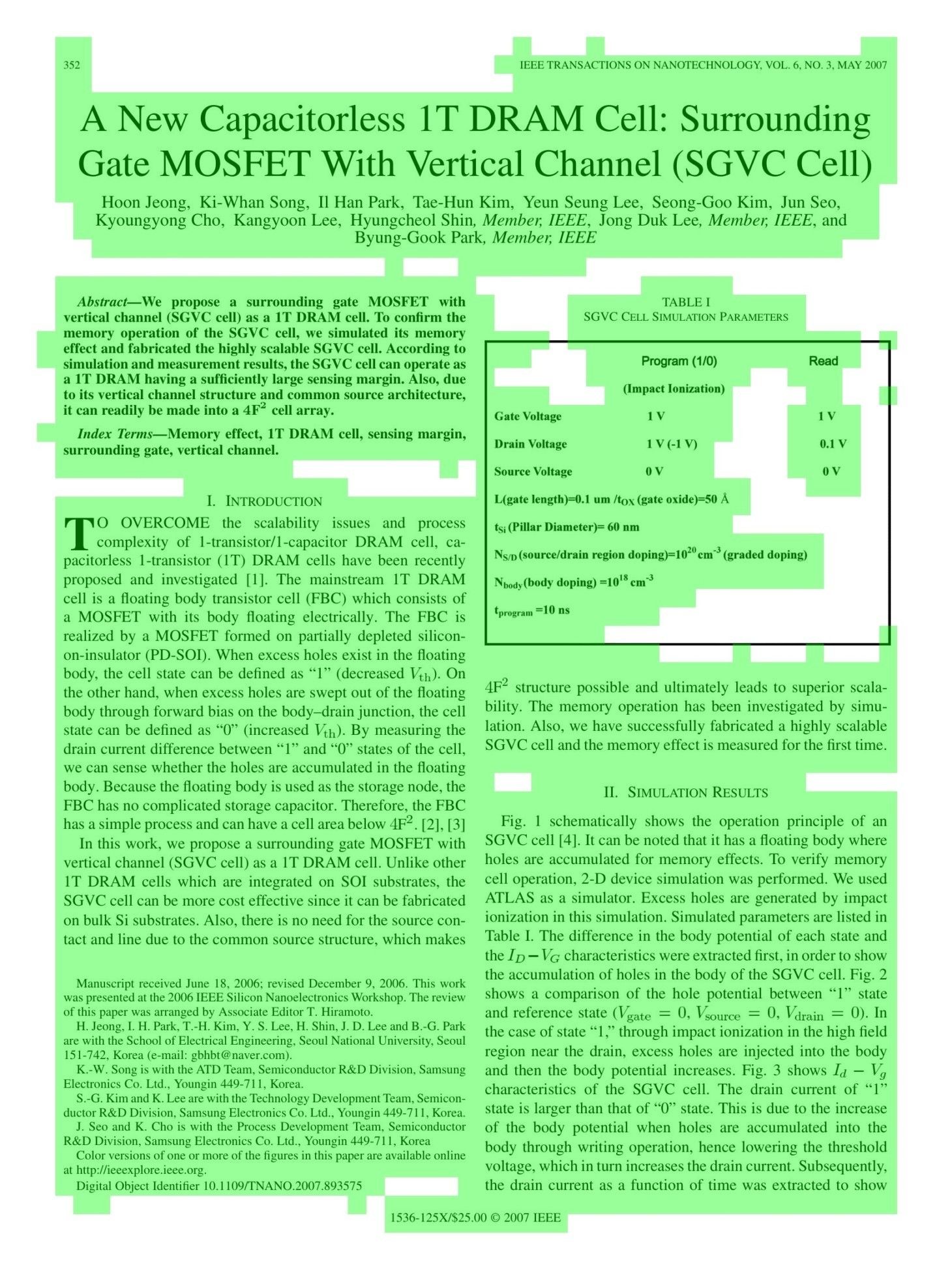}
\end{subfigure}
\hfill
\begin{subfigure}[t]{0.16\linewidth}
    \centering
    \includegraphics[width=\linewidth]{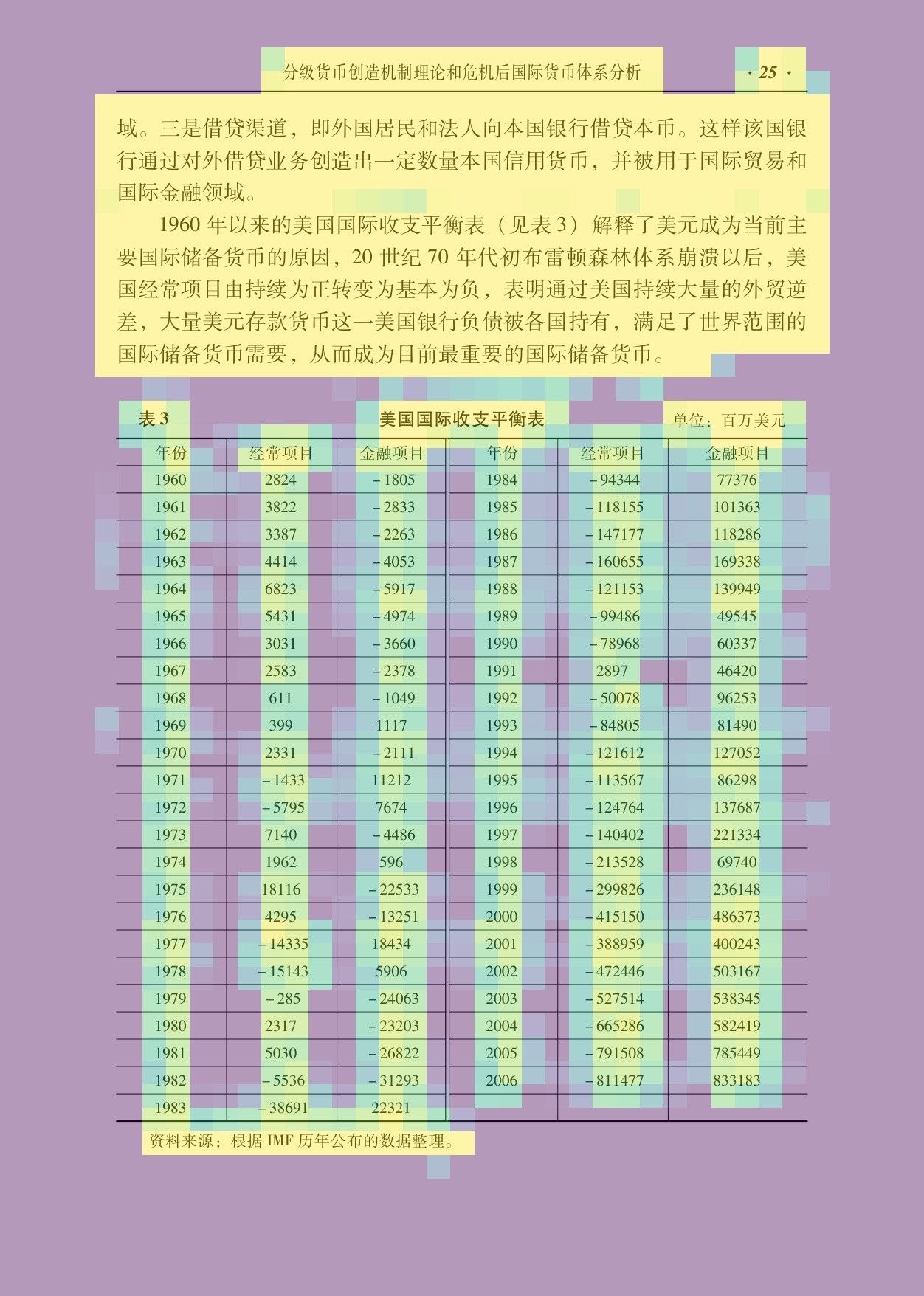}
\end{subfigure}
\hfill
\begin{subfigure}[t]{0.16\linewidth}
    \centering
    \includegraphics[width=\linewidth]{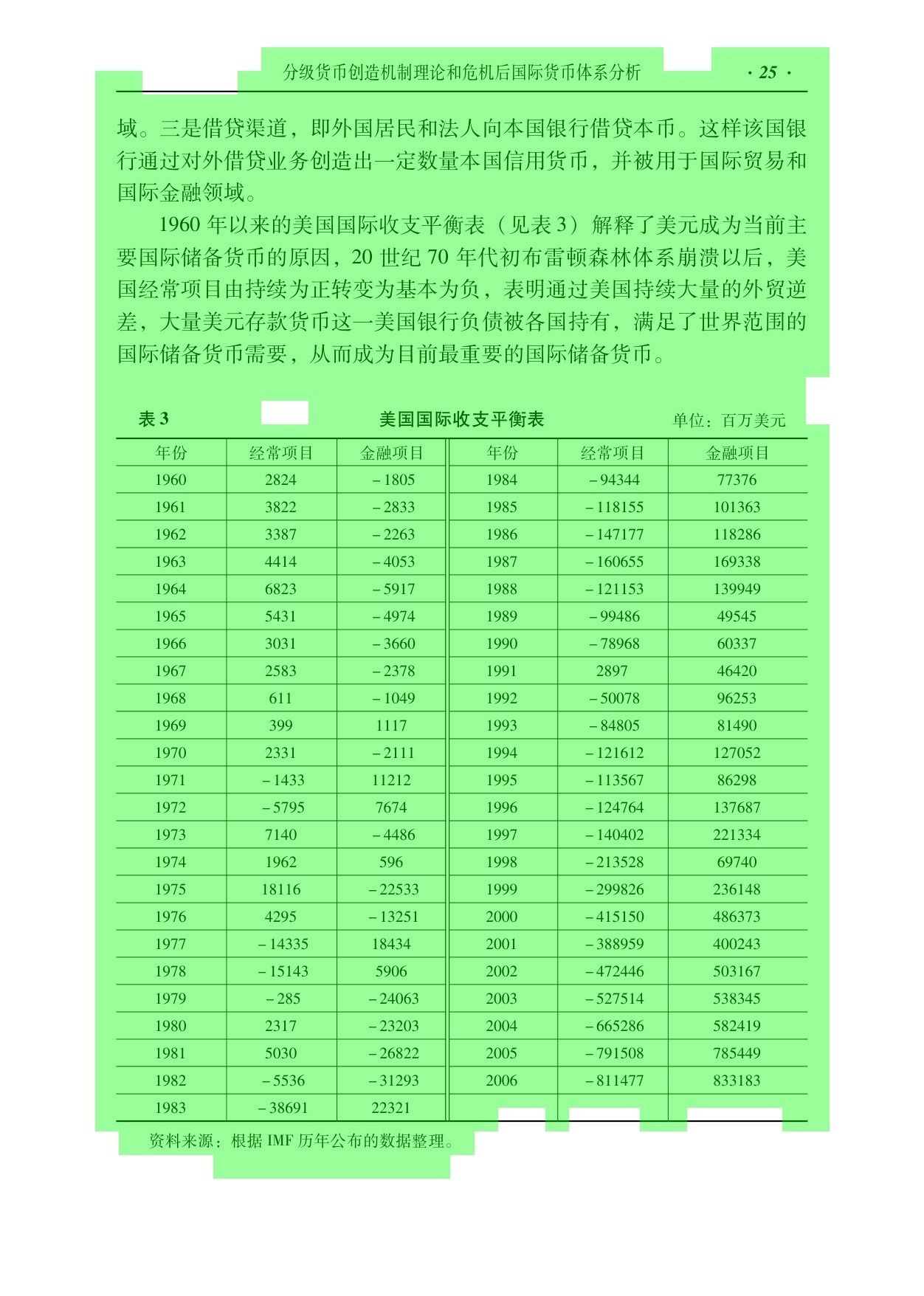}
\end{subfigure}
\hfill
\begin{subfigure}[t]{0.16\linewidth}
    \centering
    \includegraphics[width=\linewidth]{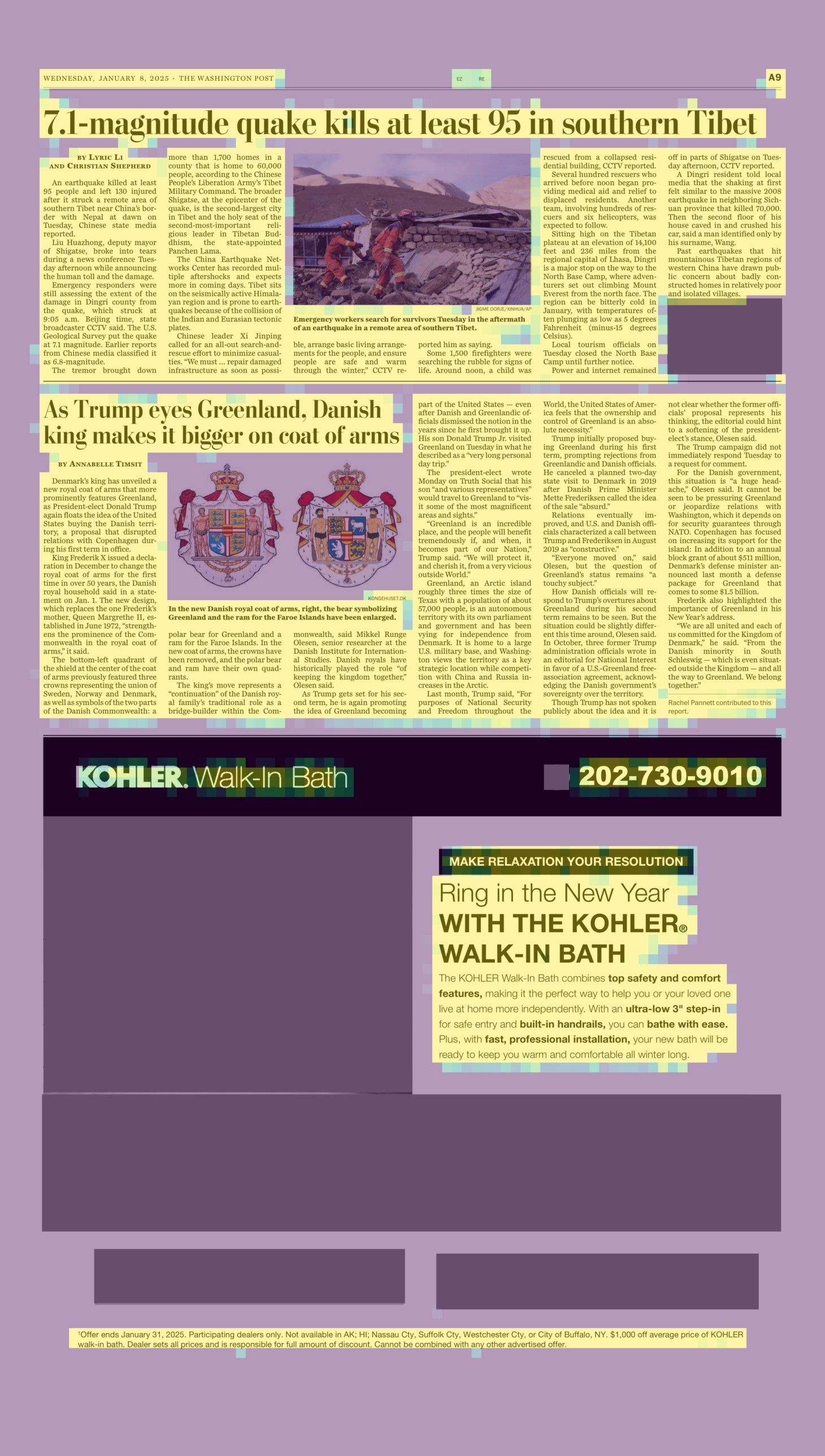}
\end{subfigure}
\hfill
\begin{subfigure}[t]{0.16\linewidth}
    \centering
    \includegraphics[width=\linewidth]{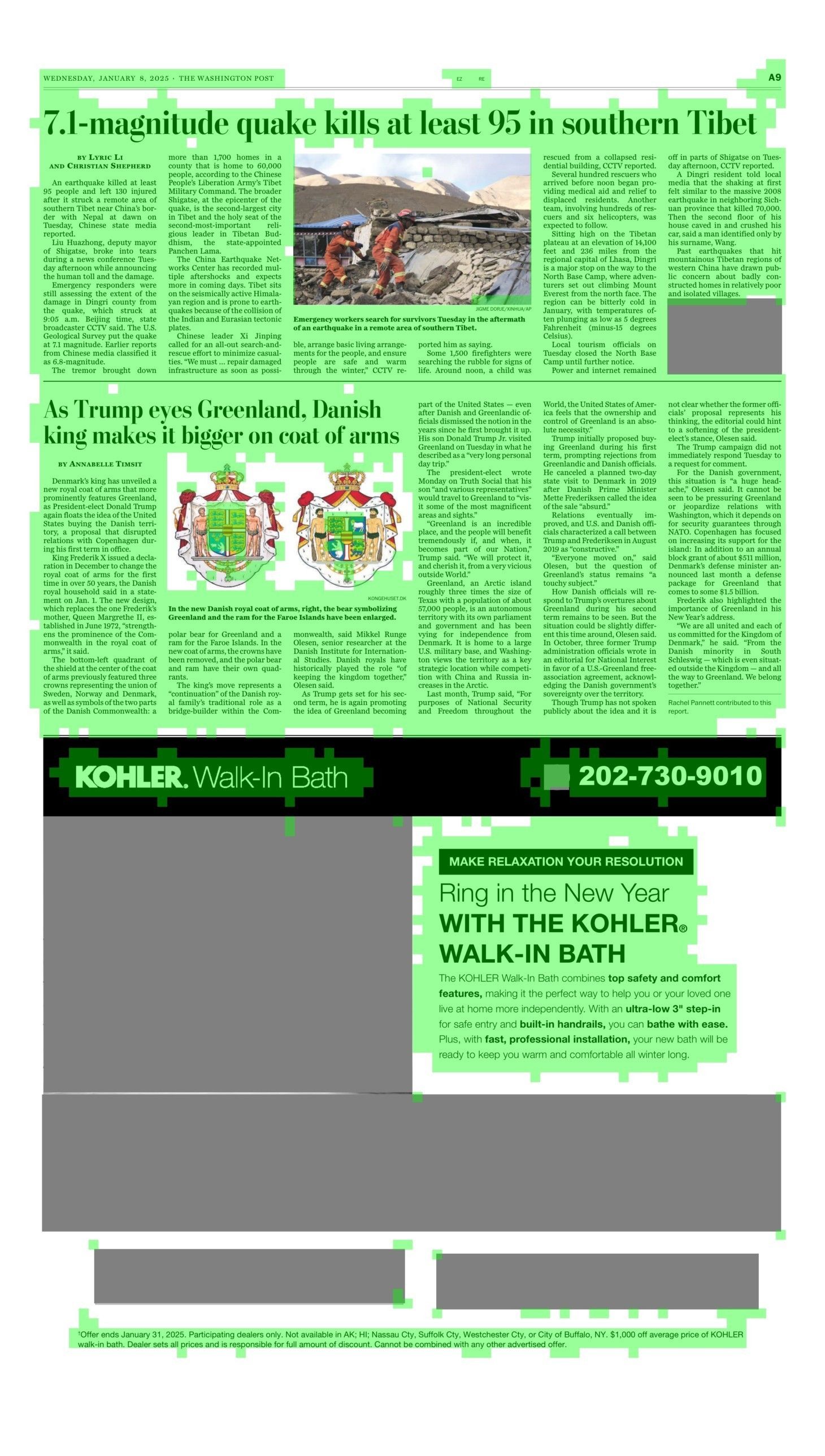}
\end{subfigure}

\vspace{0.5em}

\begin{subfigure}[t]{0.16\linewidth}
    \centering
    \includegraphics[width=\linewidth]{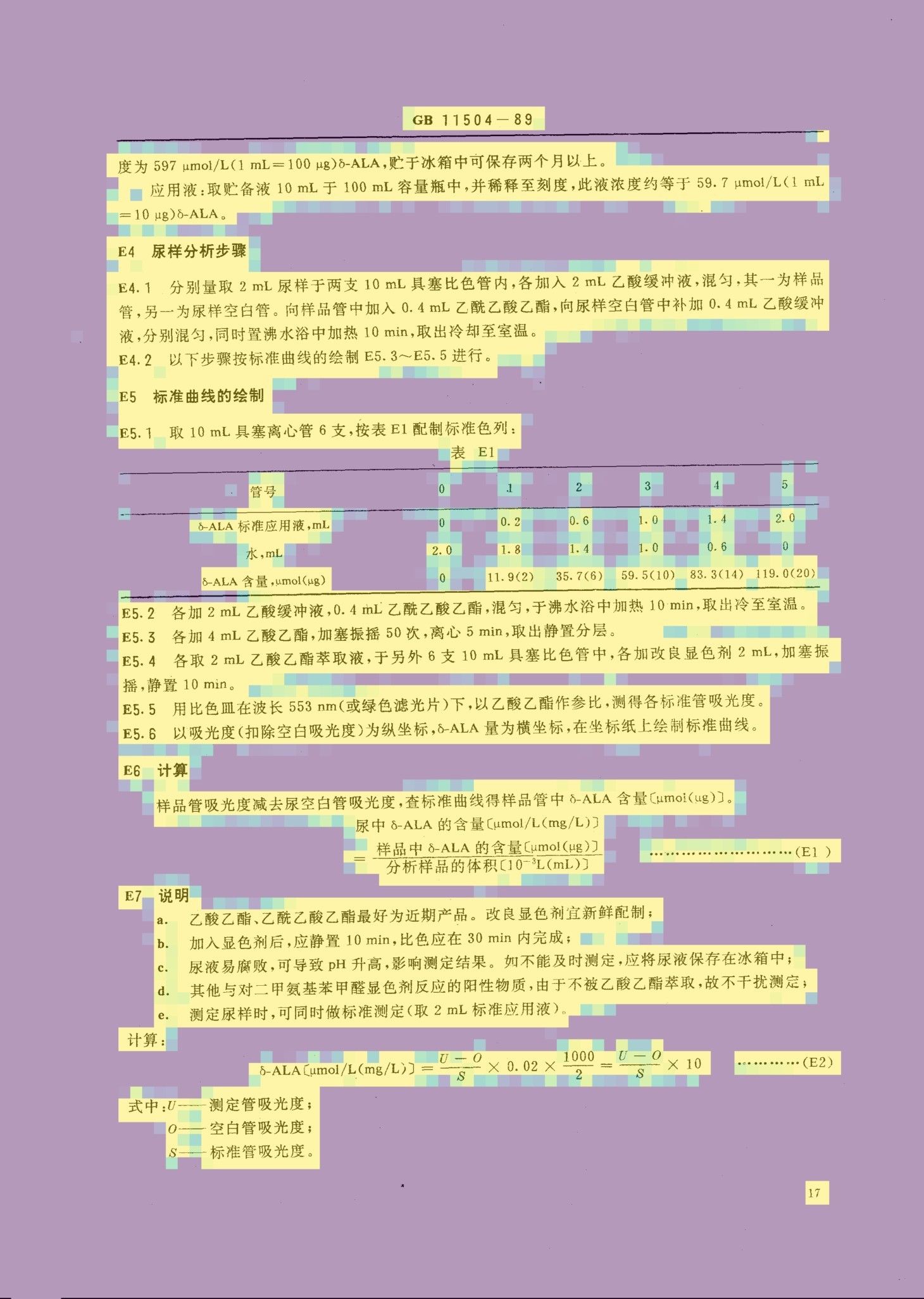}
\end{subfigure}
\hfill
\begin{subfigure}[t]{0.16\linewidth}
    \centering
    \includegraphics[width=\linewidth]{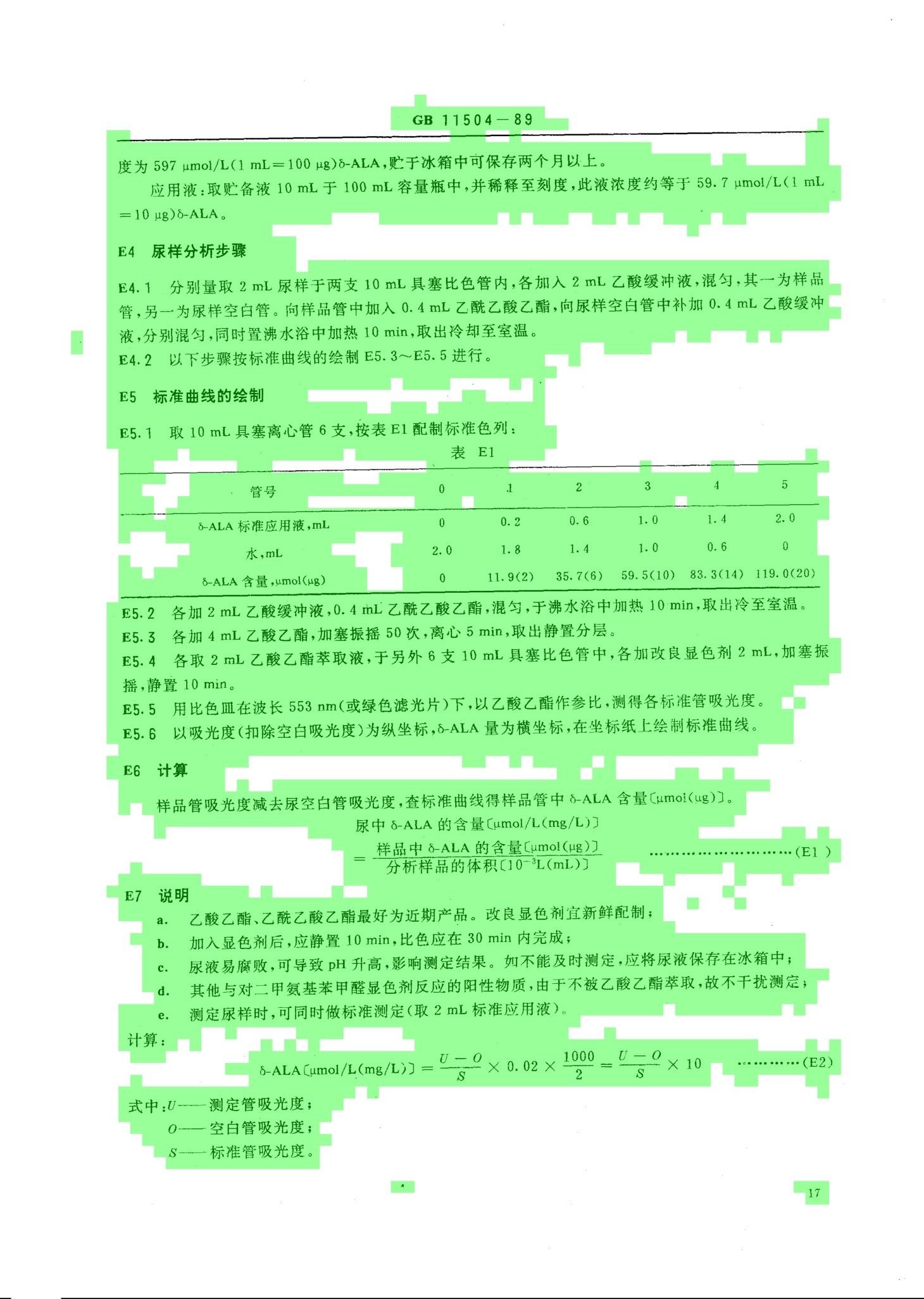}
\end{subfigure}
\hfill
\begin{subfigure}[t]{0.16\linewidth}
    \centering
    \includegraphics[width=\linewidth]{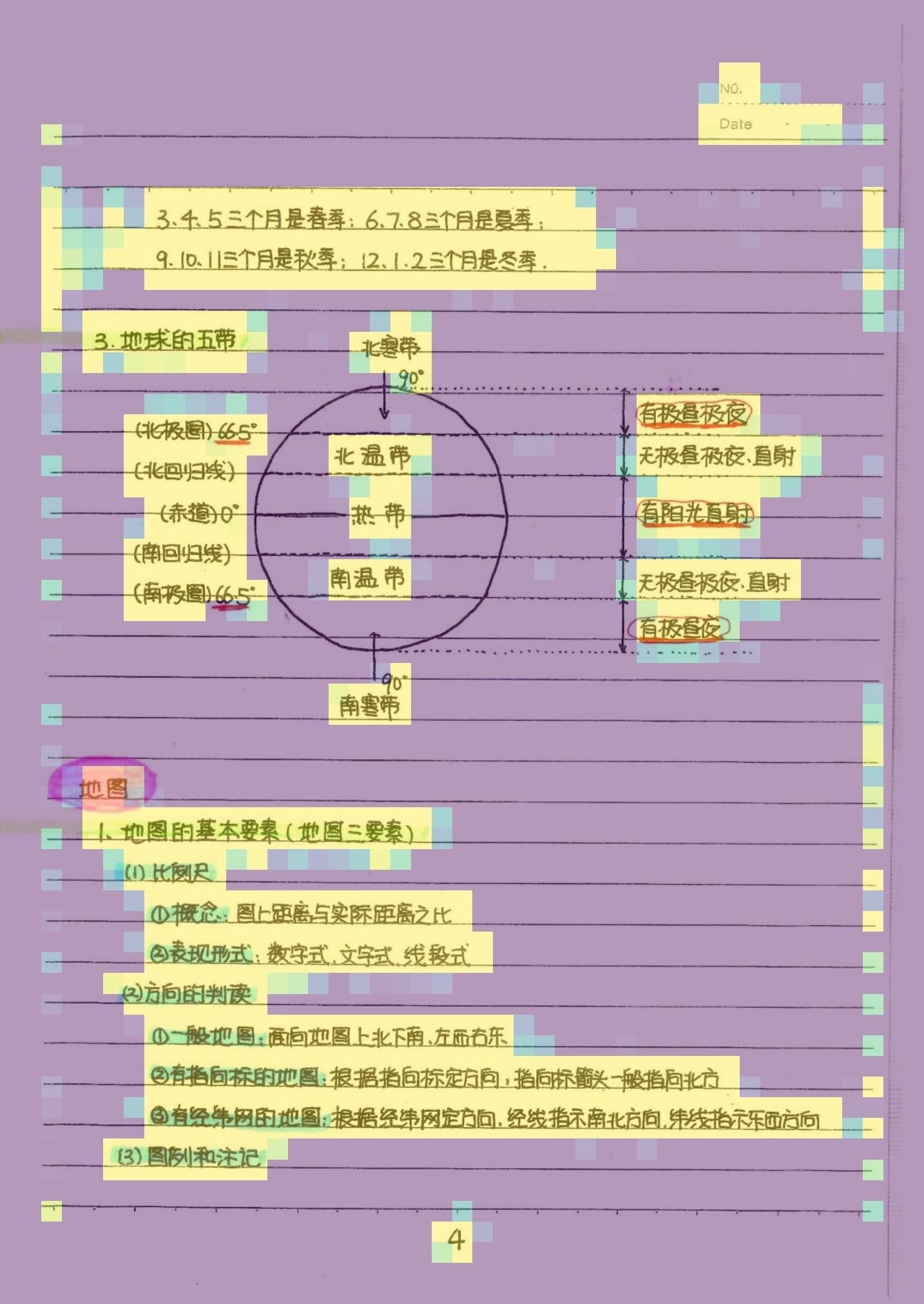}
\end{subfigure}
\hfill
\begin{subfigure}[t]{0.16\linewidth}
    \centering
    \includegraphics[width=\linewidth]{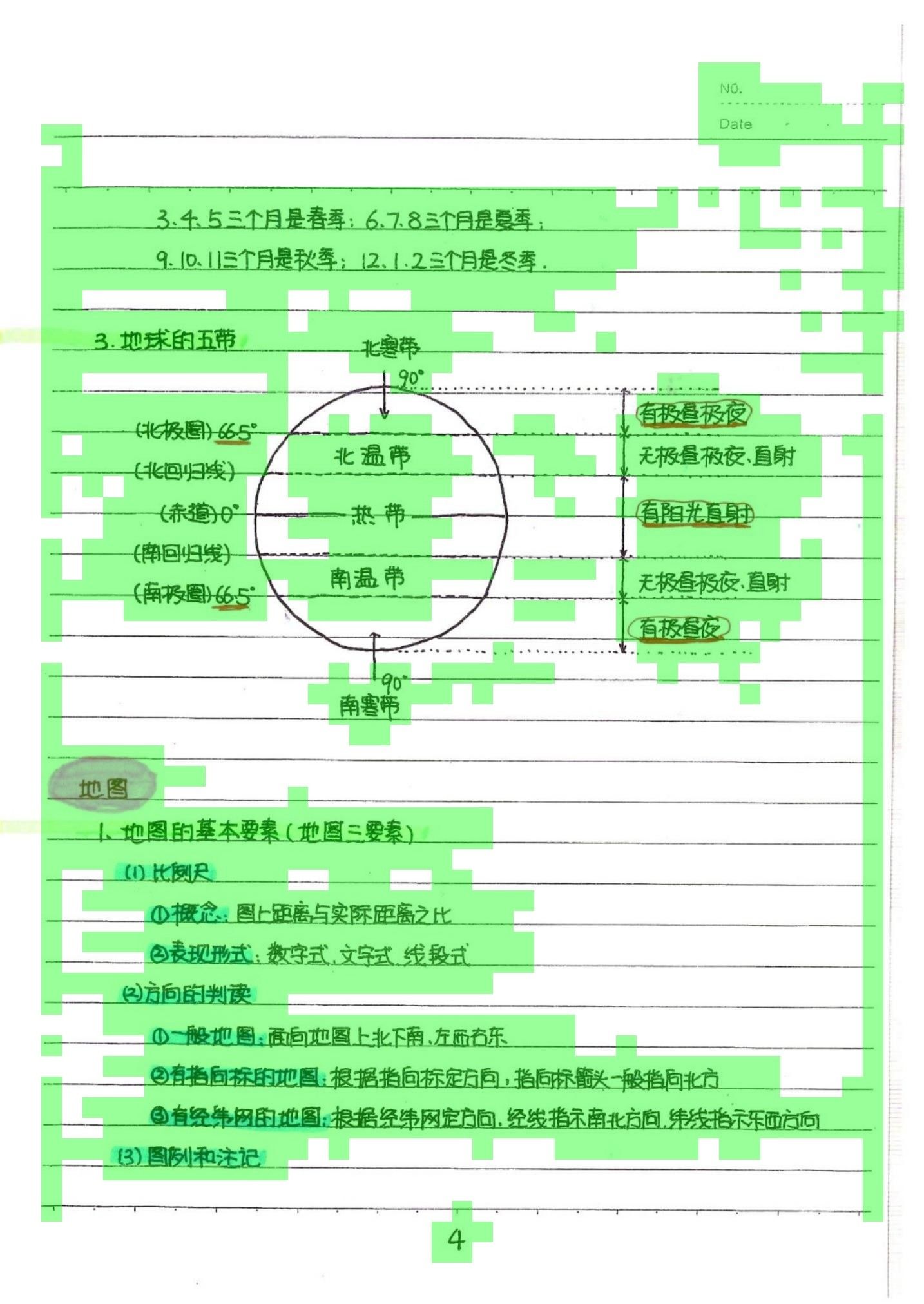}
\end{subfigure}
\hfill
\begin{subfigure}[t]{0.16\linewidth}
    \centering
    \includegraphics[width=\linewidth]{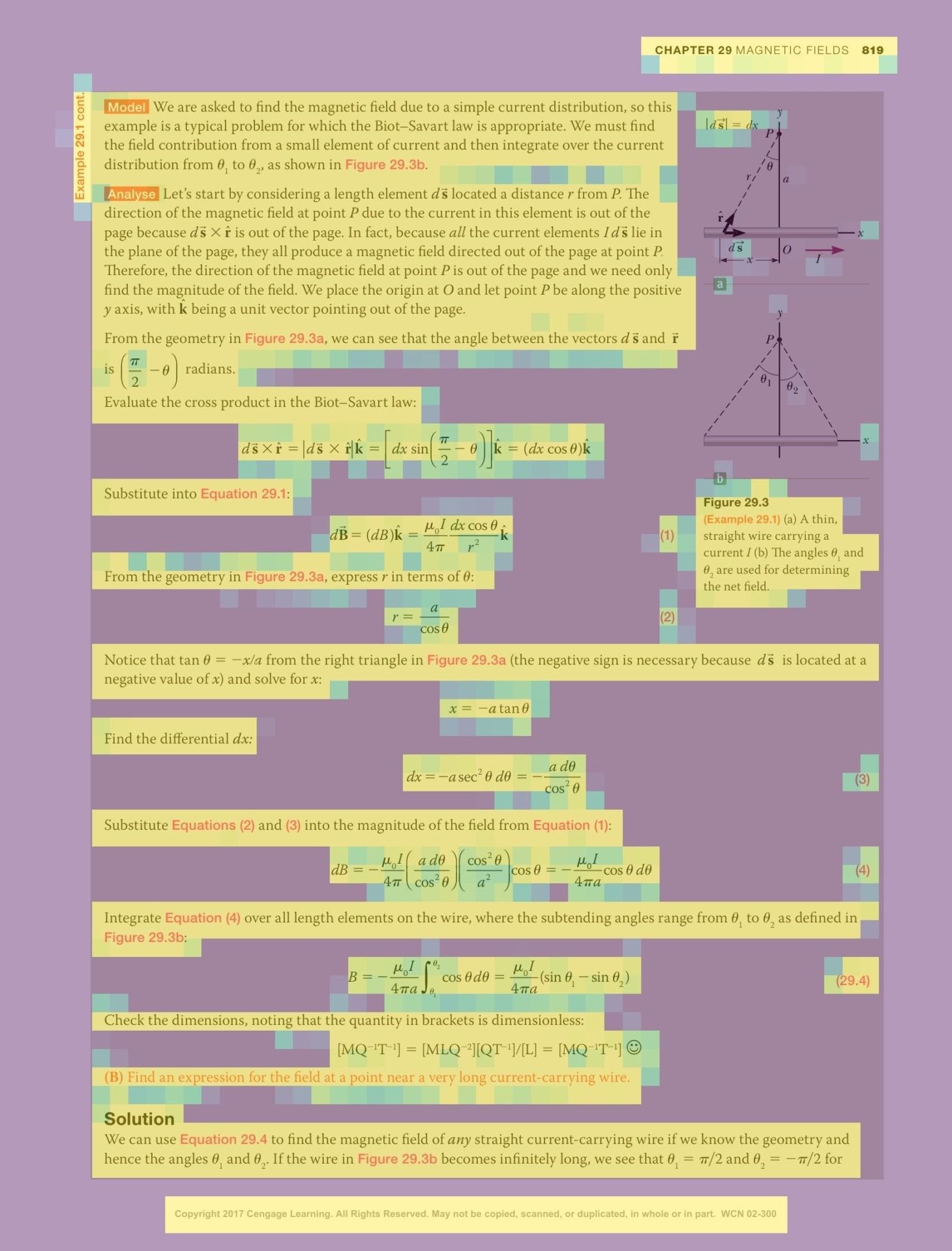}
\end{subfigure}
\hfill
\begin{subfigure}[t]{0.16\linewidth}
    \centering
    \includegraphics[width=\linewidth]{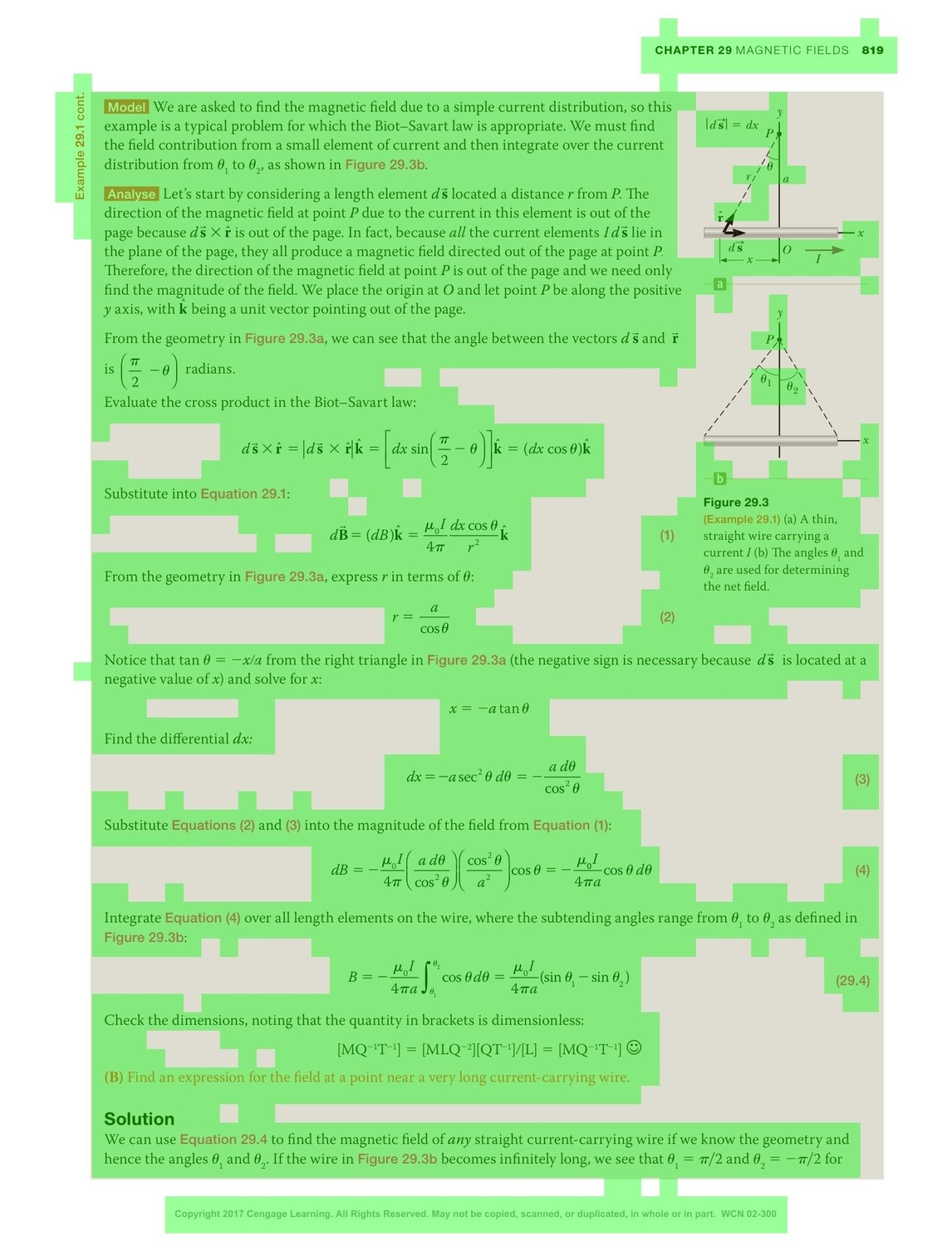}
\end{subfigure}

\vspace{0.5em}

\begin{subfigure}[t]{0.16\linewidth}
    \centering
    \includegraphics[width=\linewidth]{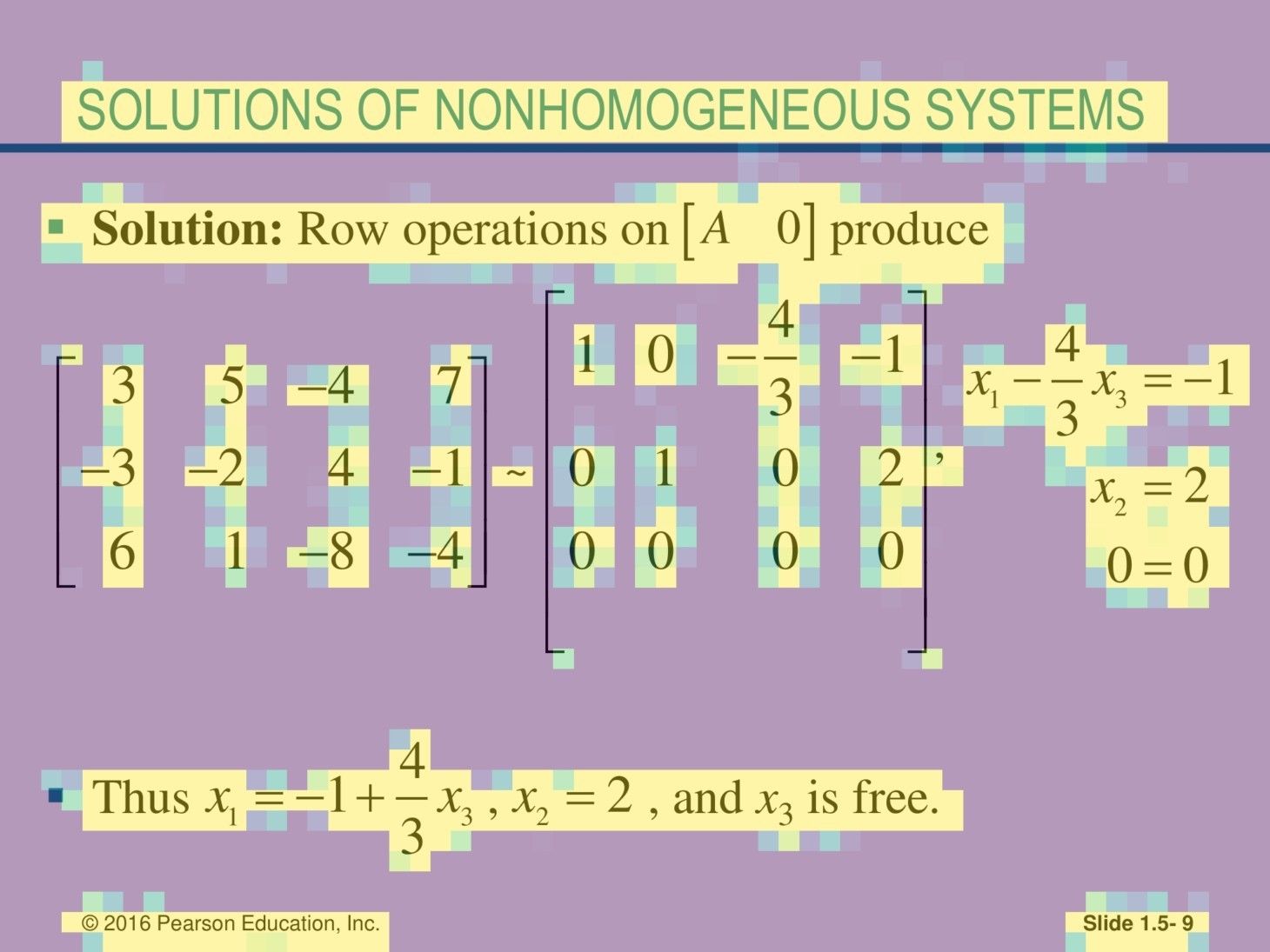}
\end{subfigure}
\hfill
\begin{subfigure}[t]{0.16\linewidth}
    \centering
    \includegraphics[width=\linewidth]{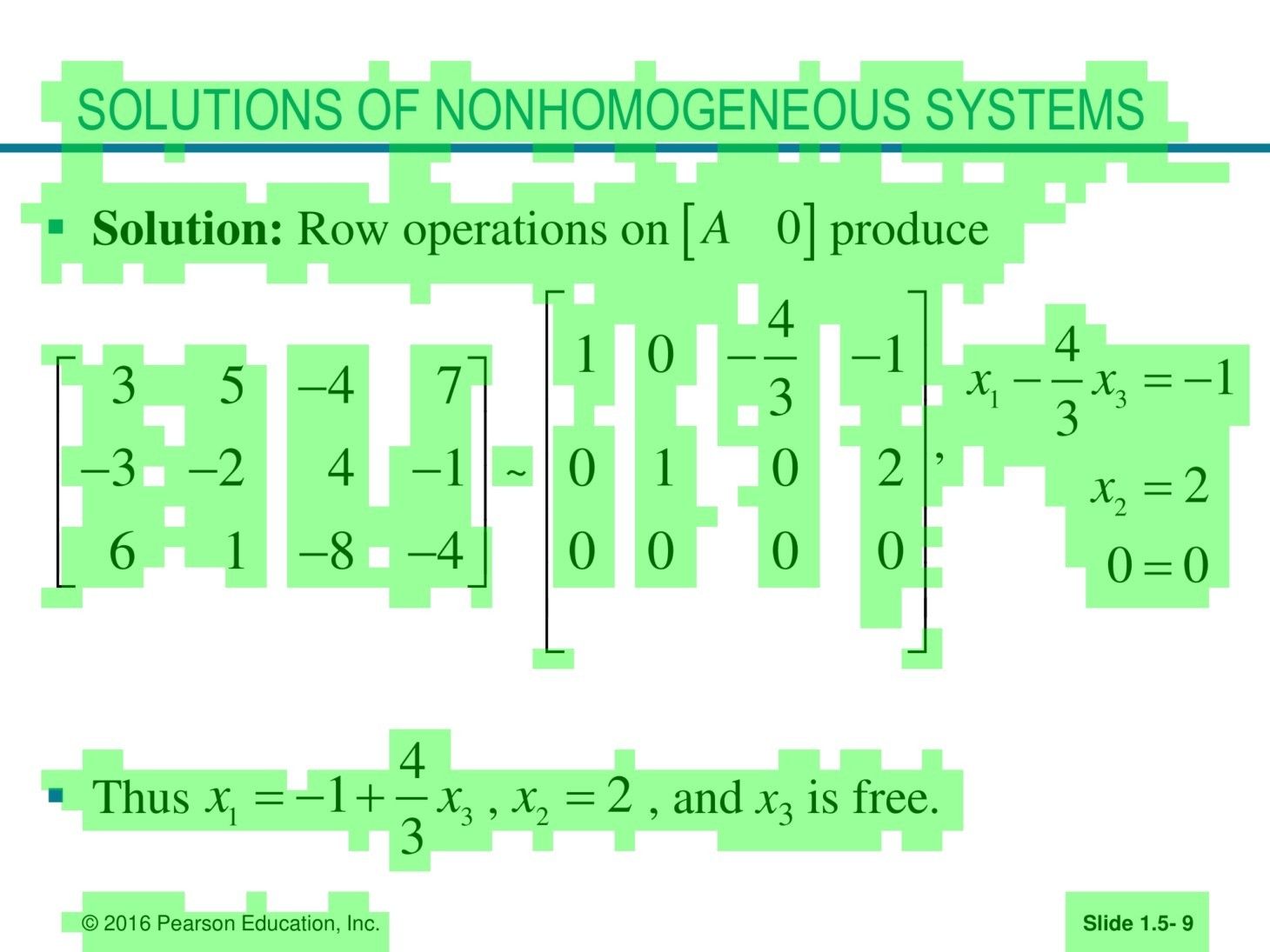}
\end{subfigure}
\hfill
\begin{subfigure}[t]{0.16\linewidth}
    \centering
    \includegraphics[width=\linewidth]{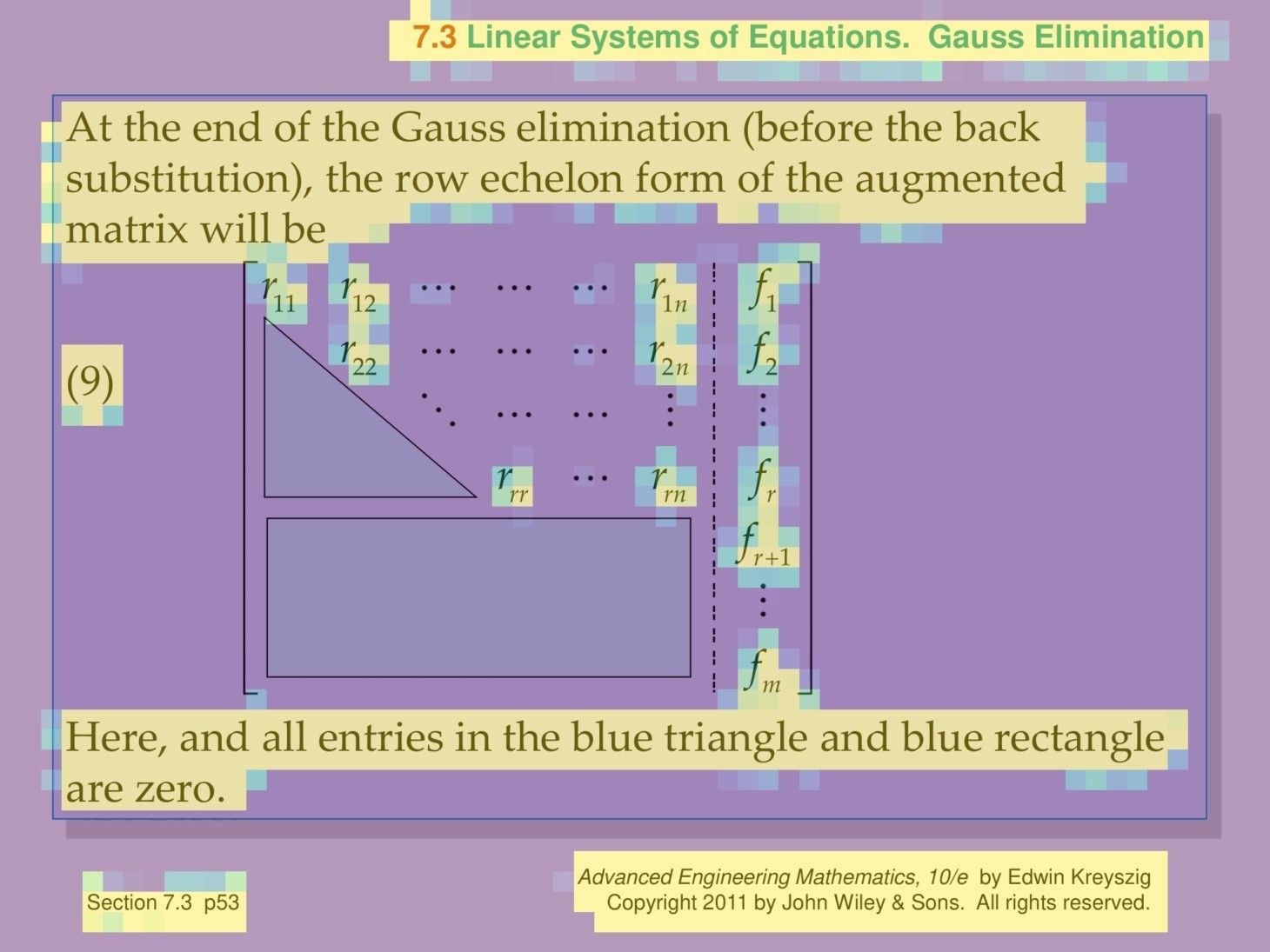}
\end{subfigure}
\hfill
\begin{subfigure}[t]{0.16\linewidth}
    \centering
    \includegraphics[width=\linewidth]{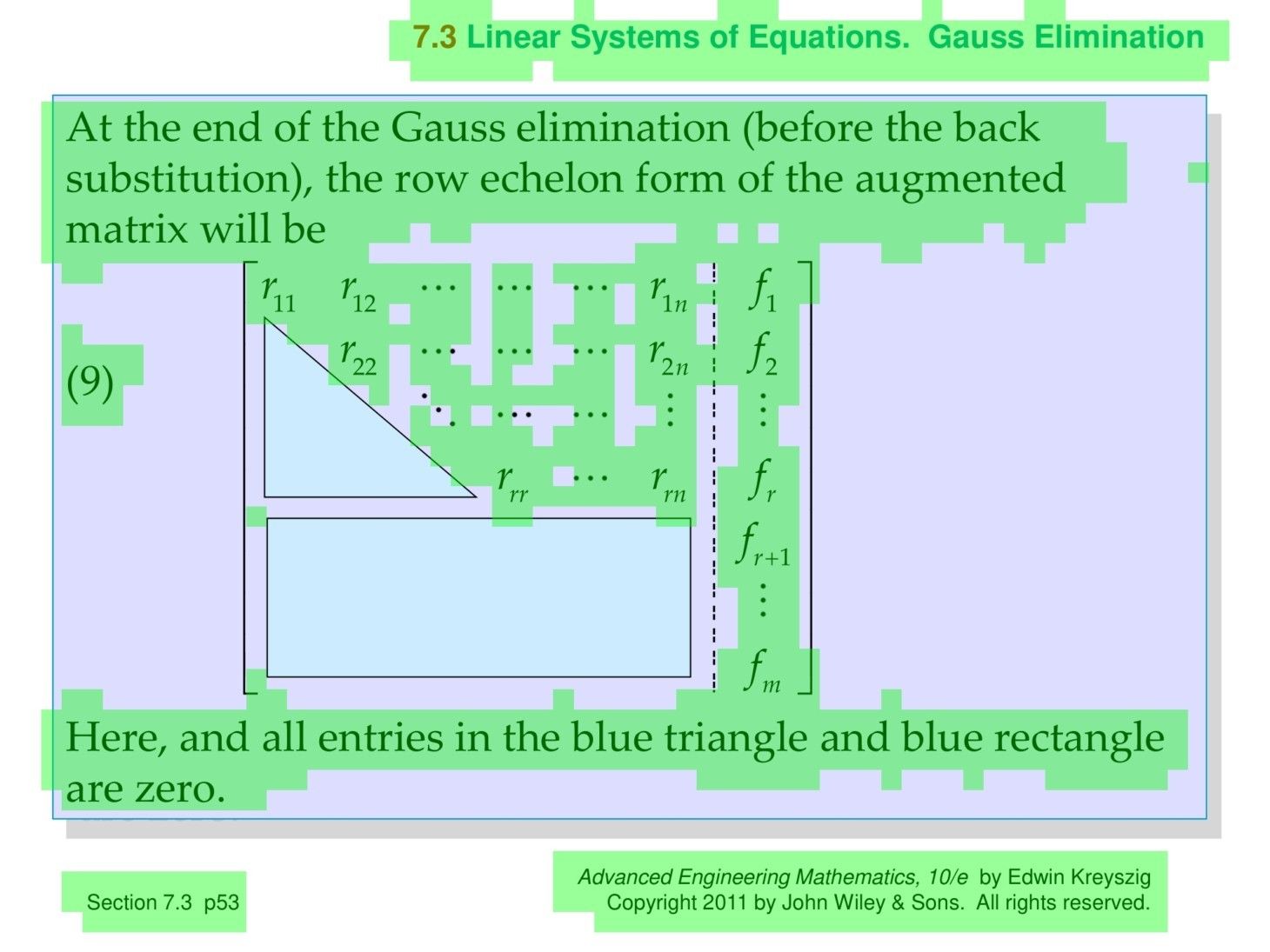}
\end{subfigure}
\hfill
\begin{subfigure}[t]{0.16\linewidth}
    \centering
    \includegraphics[width=\linewidth]{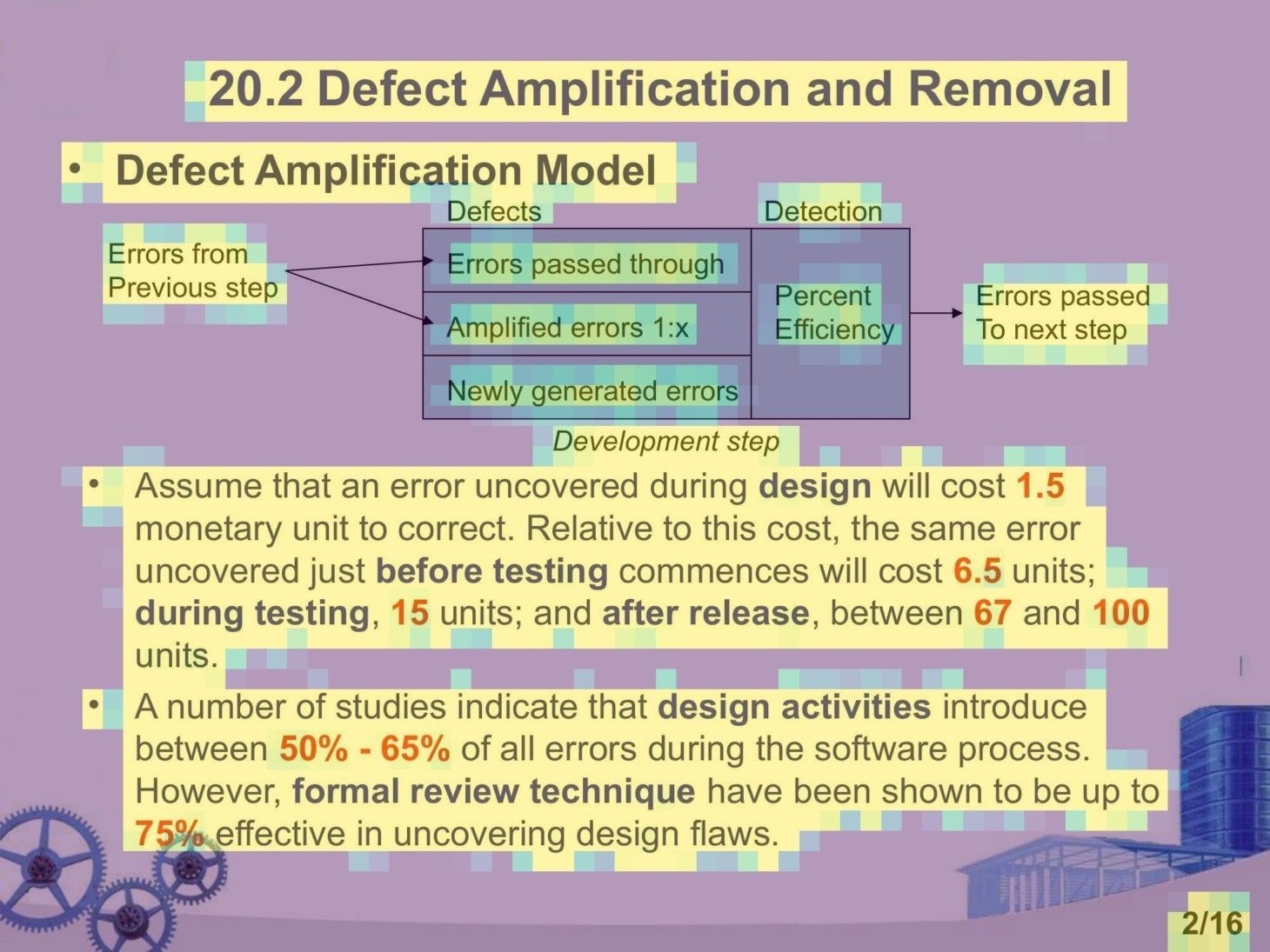}
\end{subfigure}
\hfill
\begin{subfigure}[t]{0.16\linewidth}
    \centering
    \includegraphics[width=\linewidth]{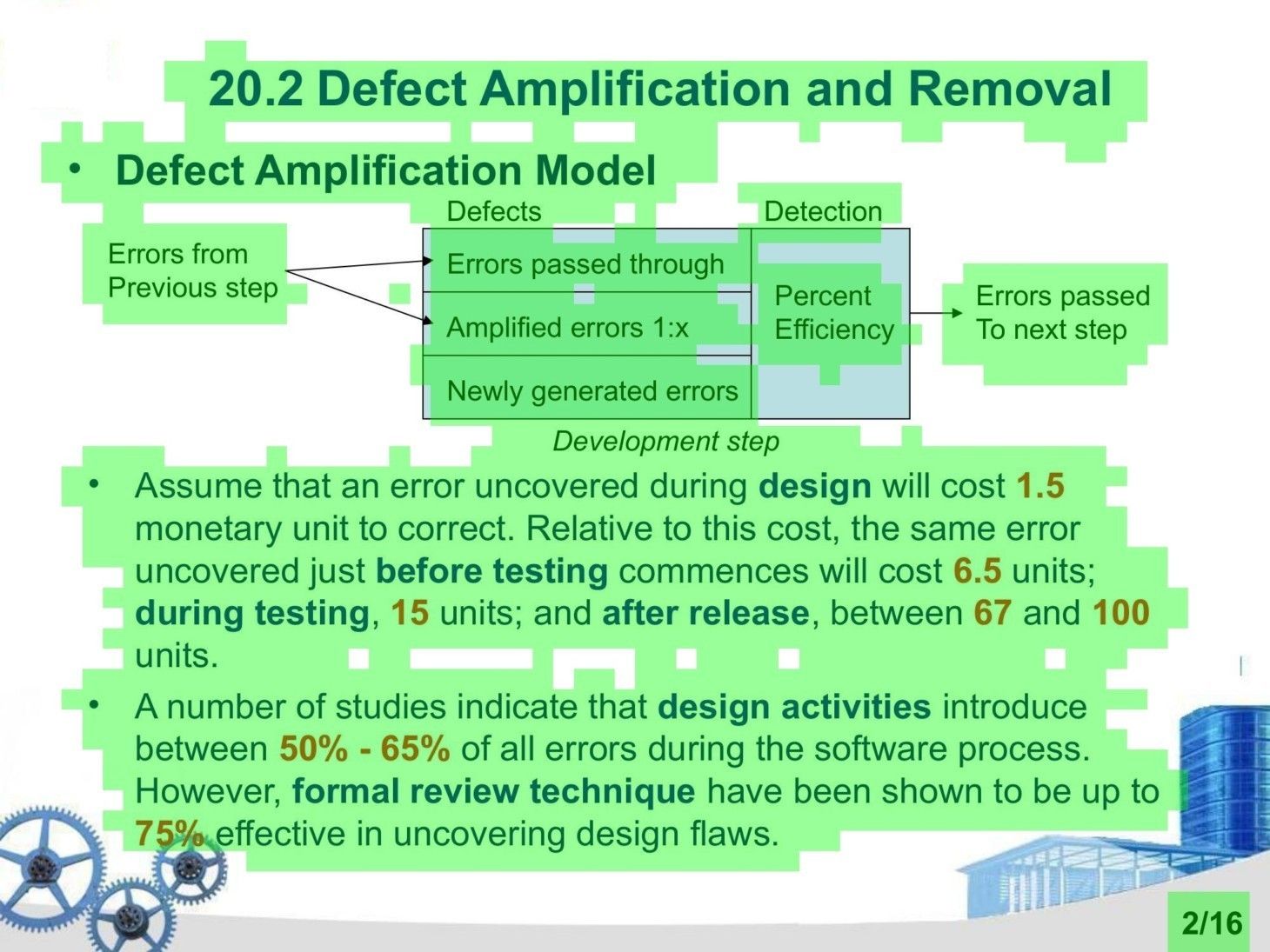}
\end{subfigure}

\vspace{0.5em}

\begin{subfigure}[t]{0.16\linewidth}
    \centering
    \includegraphics[width=\linewidth]{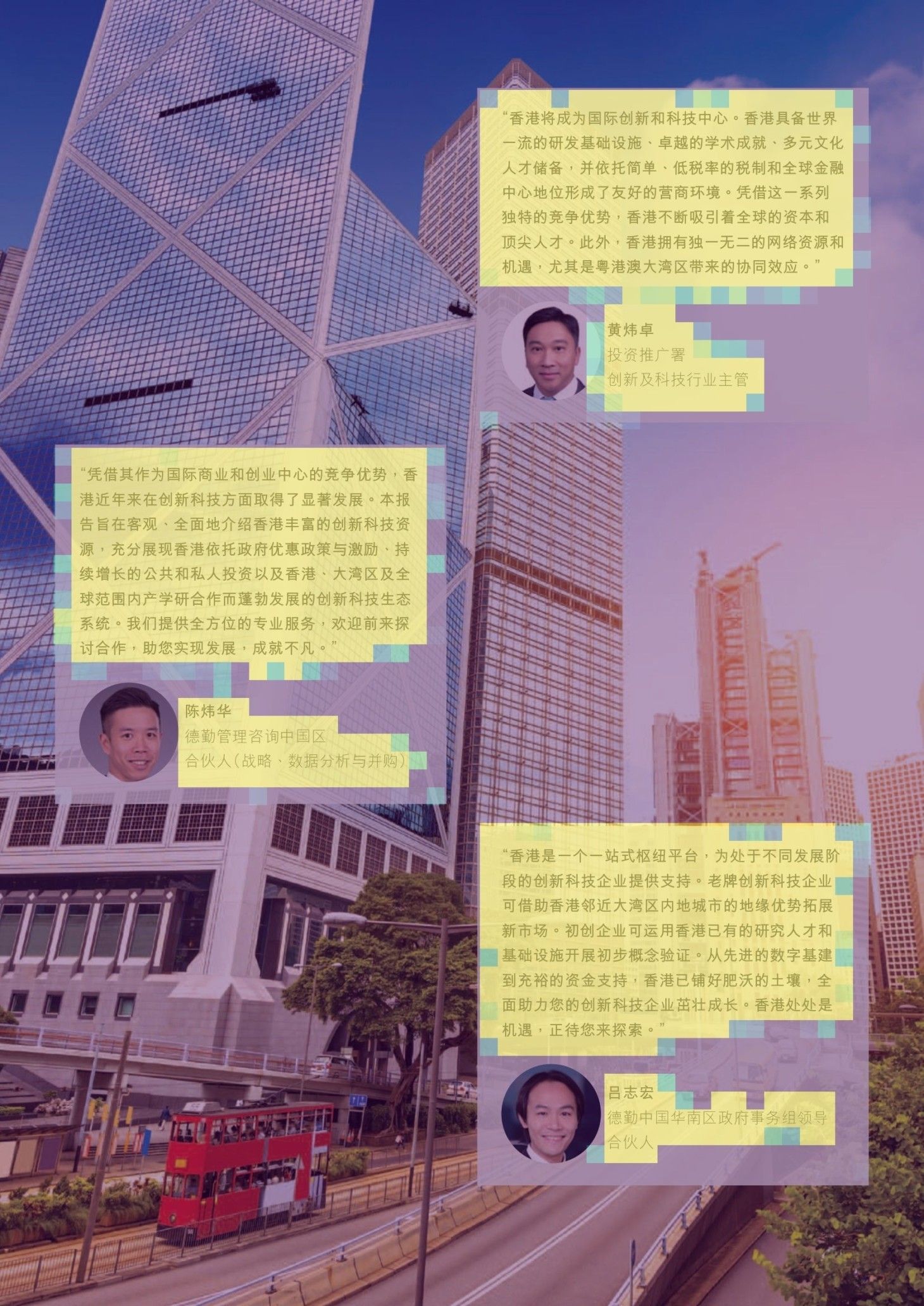}
\end{subfigure}
\hfill
\begin{subfigure}[t]{0.16\linewidth}
    \centering
    \includegraphics[width=\linewidth]{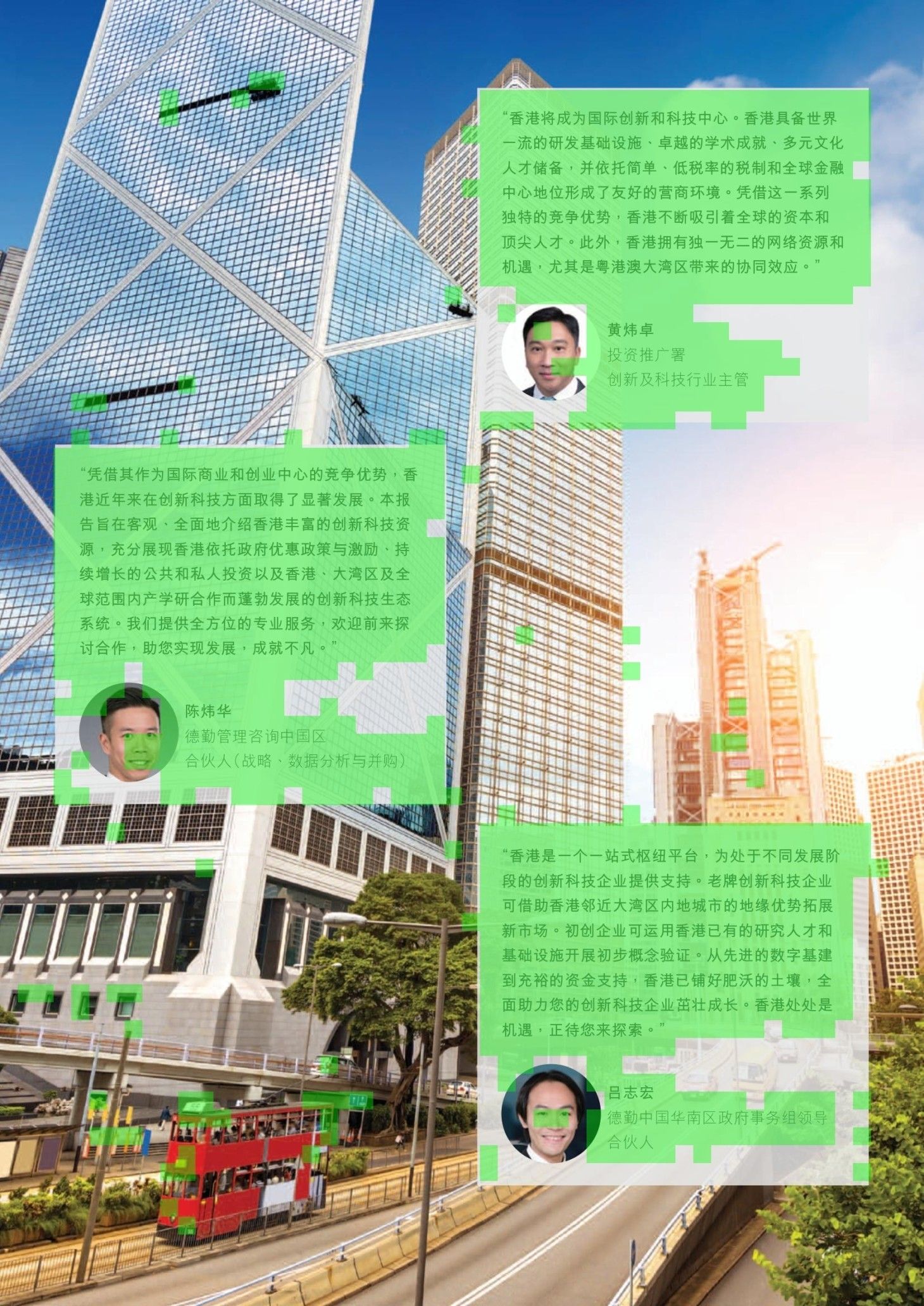}
\end{subfigure}
\hfill
\begin{subfigure}[t]{0.16\linewidth}
    \centering
    \includegraphics[width=\linewidth]{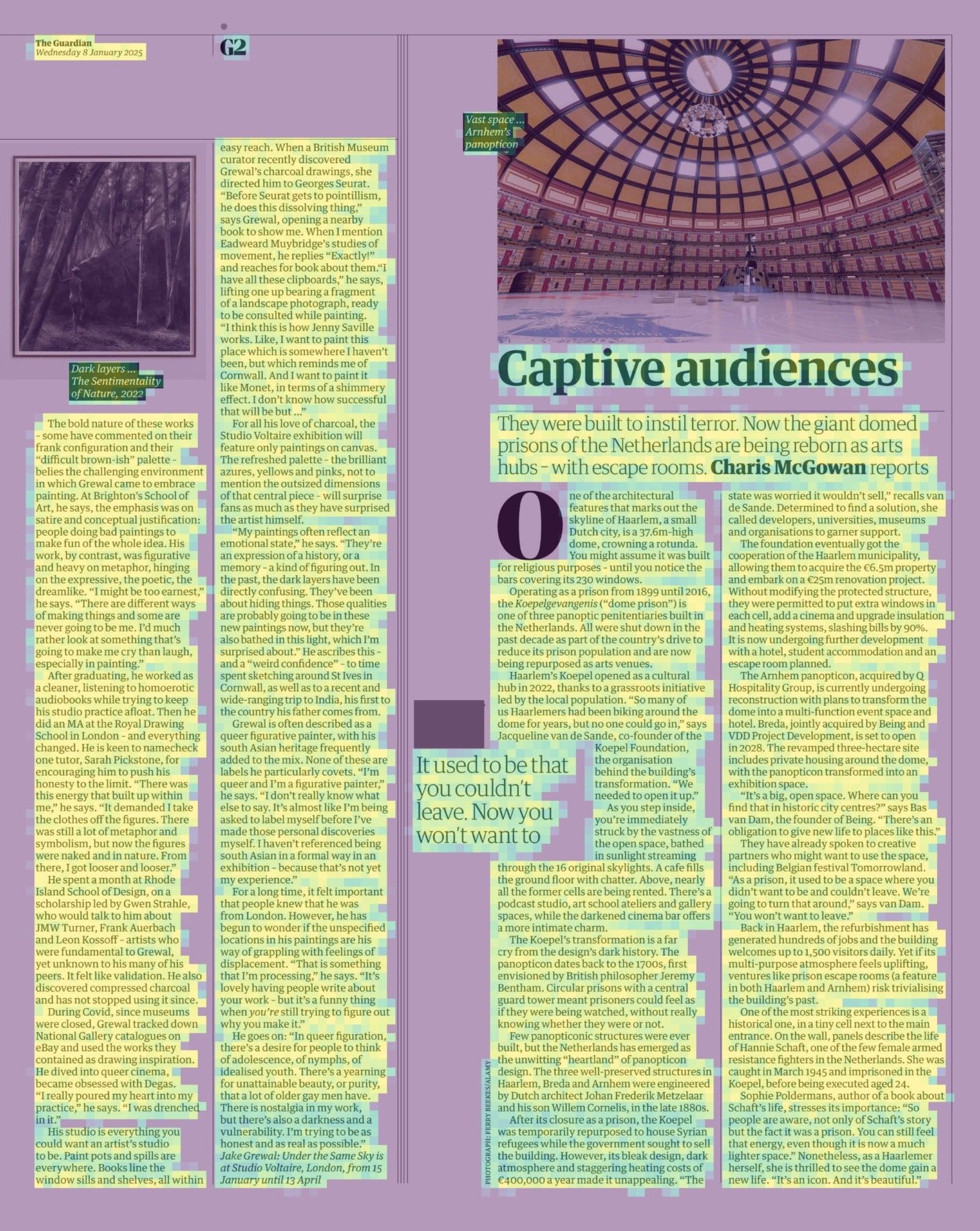}
\end{subfigure}
\hfill
\begin{subfigure}[t]{0.16\linewidth}
    \centering
    \includegraphics[width=\linewidth]{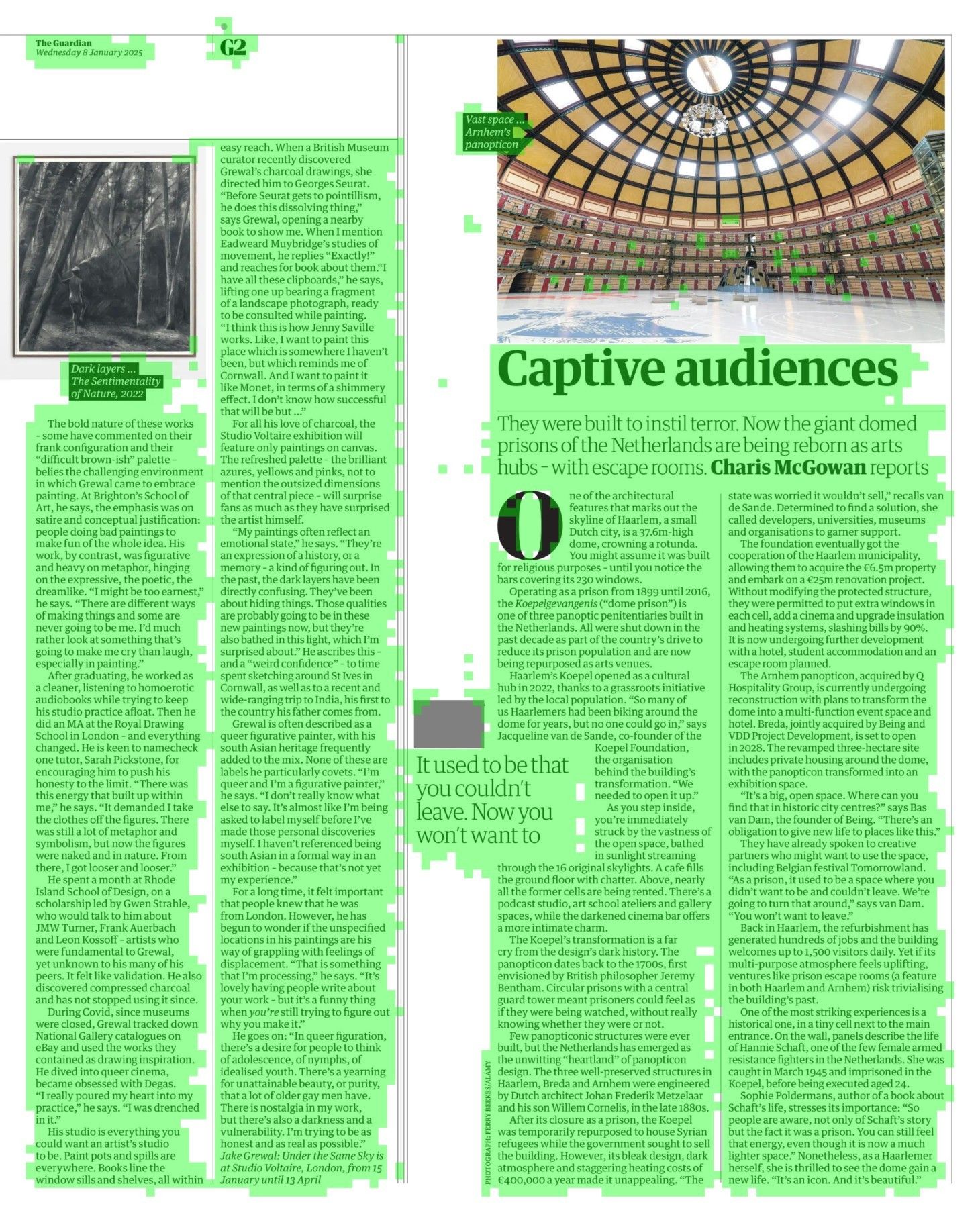}
\end{subfigure}
\hfill
\begin{subfigure}[t]{0.16\linewidth}
    \centering
    \includegraphics[width=\linewidth]{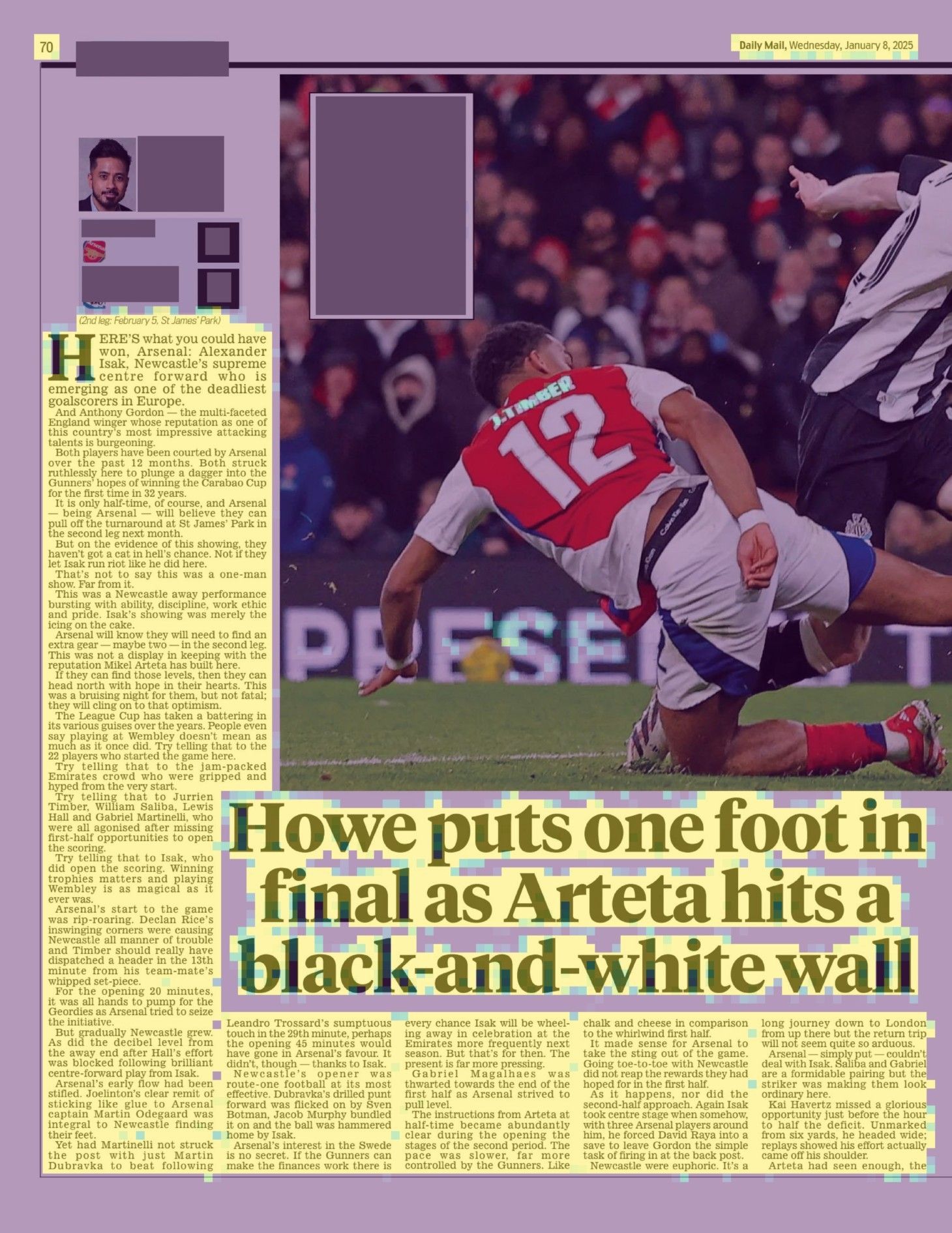}
\end{subfigure}
\hfill
\begin{subfigure}[t]{0.16\linewidth}
    \centering
    \includegraphics[width=\linewidth]{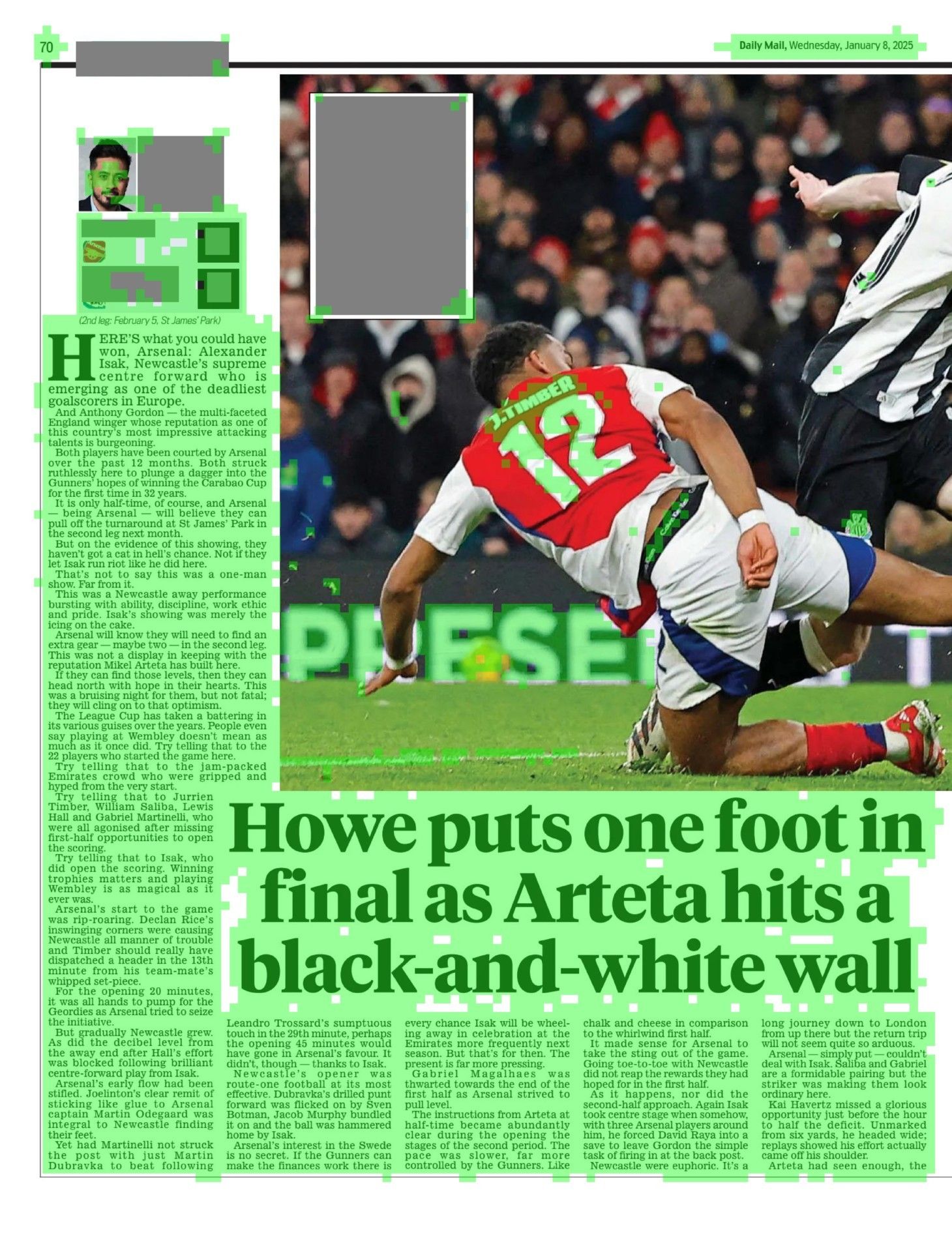}
\end{subfigure}

\begin{subfigure}[t]{0.16\linewidth}
    \centering
    \includegraphics[width=\linewidth]{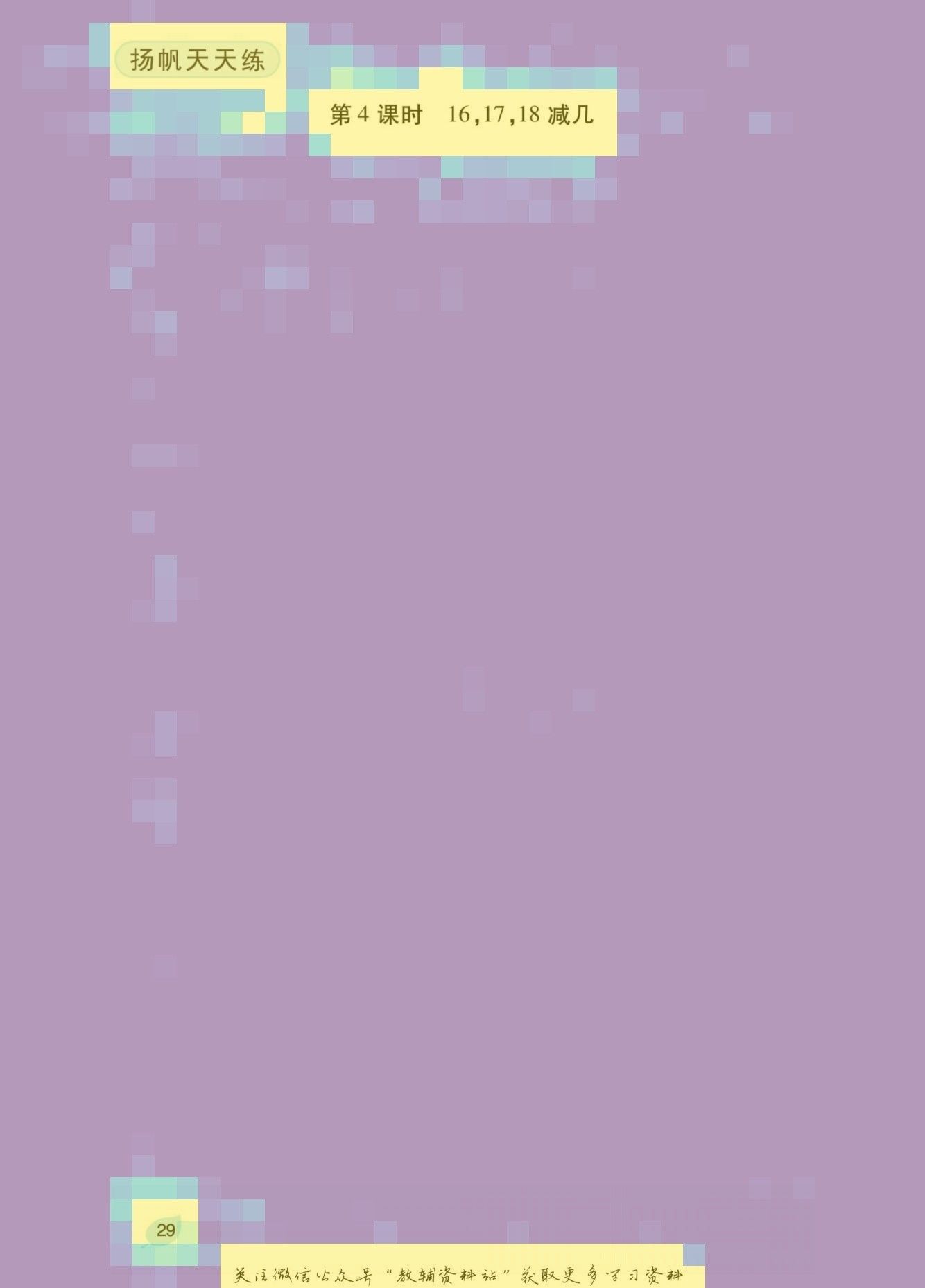}
\end{subfigure}
\hfill
\begin{subfigure}[t]{0.16\linewidth}
    \centering
    \includegraphics[width=\linewidth]{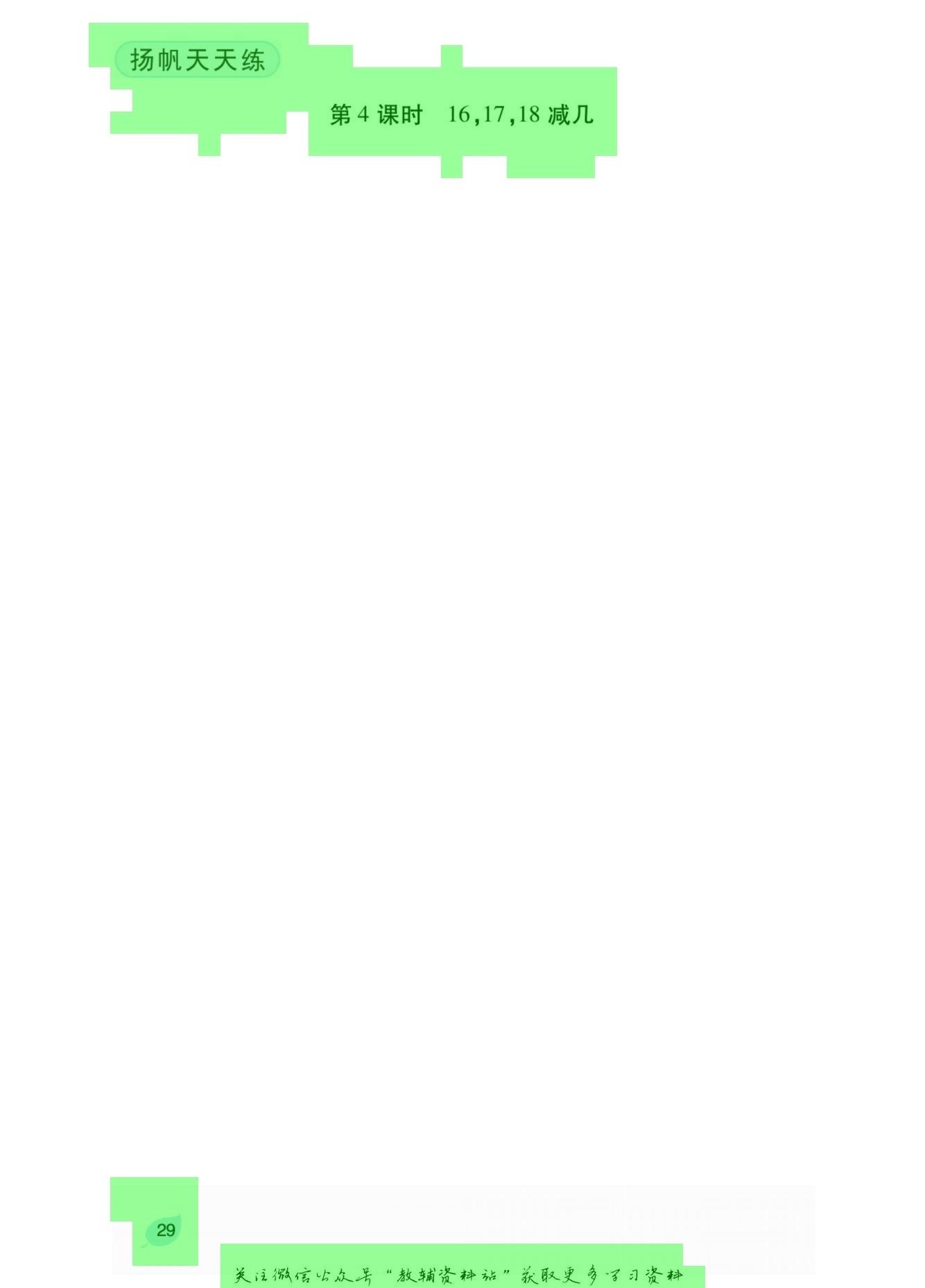}
\end{subfigure}
\hfill
\begin{subfigure}[t]{0.16\linewidth}
    \centering
    \includegraphics[width=\linewidth]{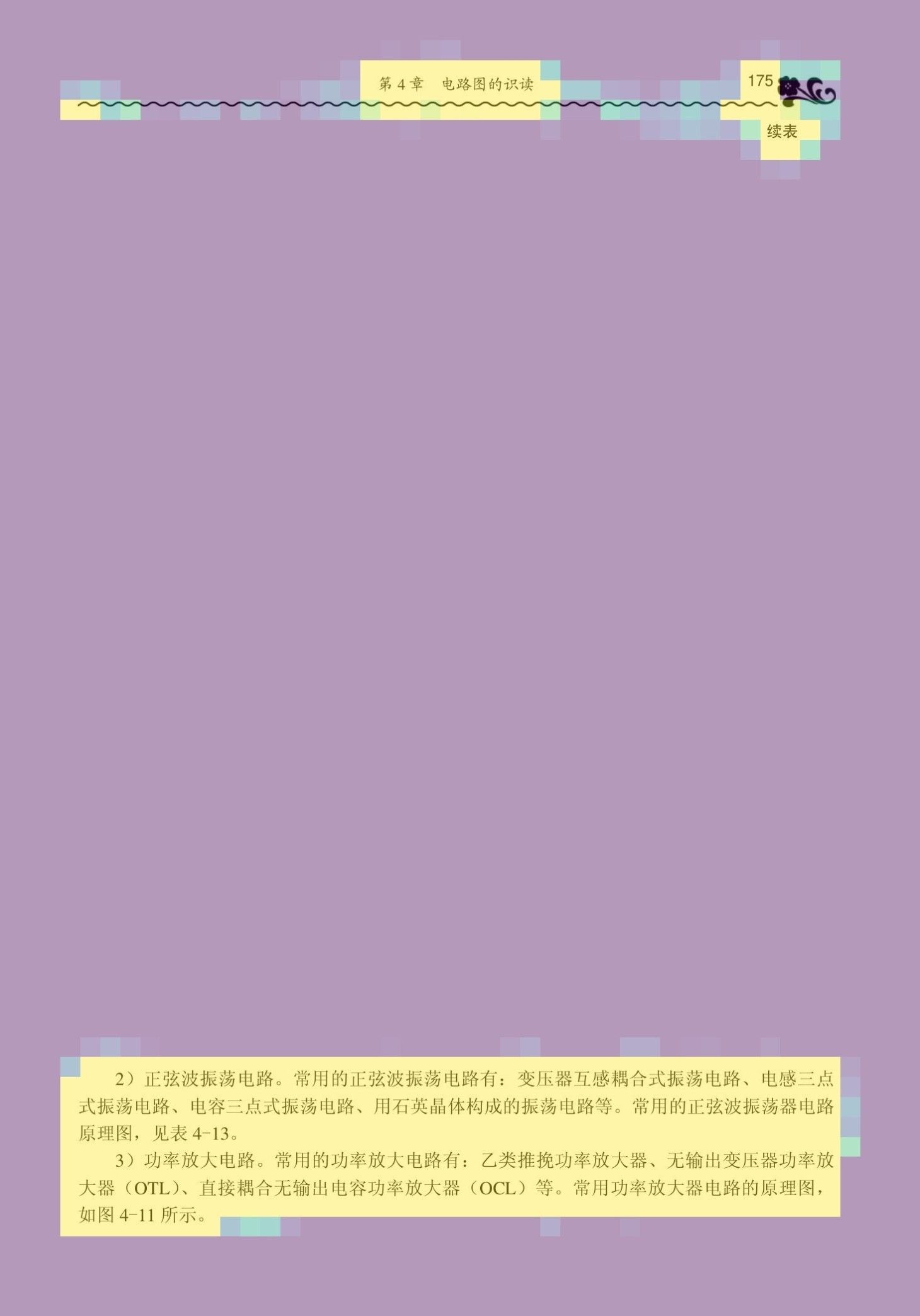}
\end{subfigure}
\hfill
\begin{subfigure}[t]{0.16\linewidth}
    \centering
    \includegraphics[width=\linewidth]{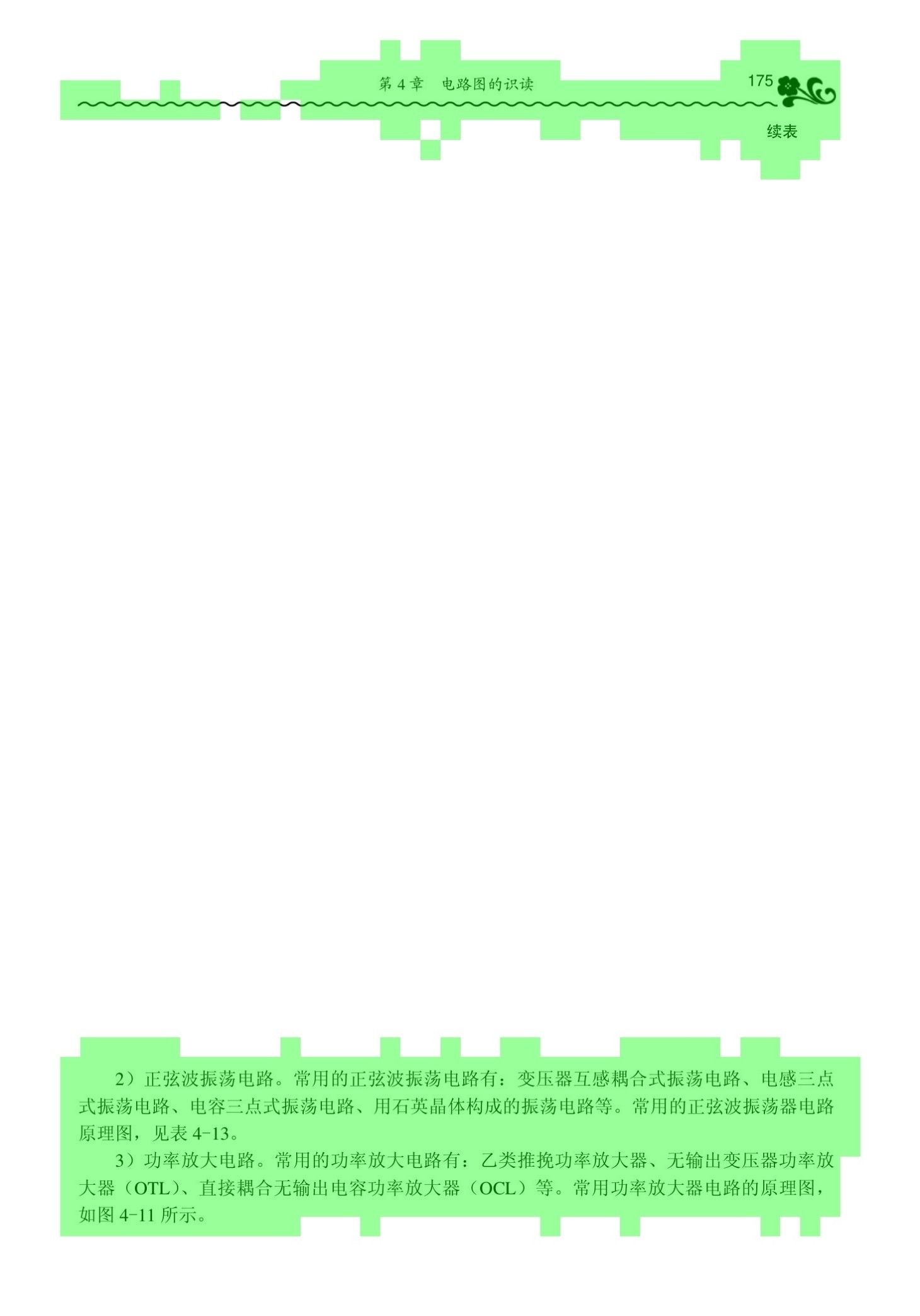}
\end{subfigure}
\hfill
\begin{subfigure}[t]{0.16\linewidth}
    \centering
    \includegraphics[width=\linewidth]{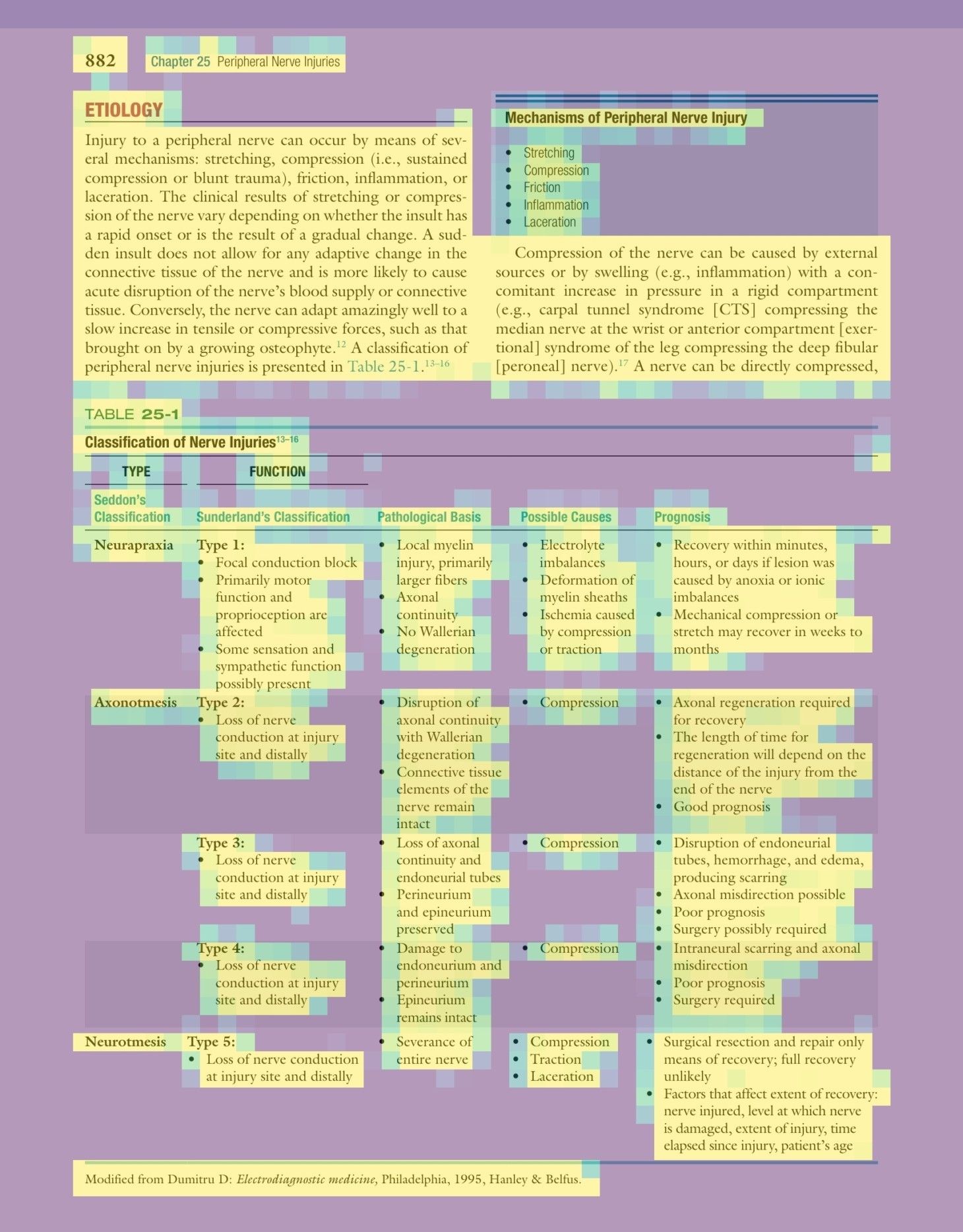}
\end{subfigure}
\hfill
\begin{subfigure}[t]{0.16\linewidth}
    \centering
    \includegraphics[width=\linewidth]{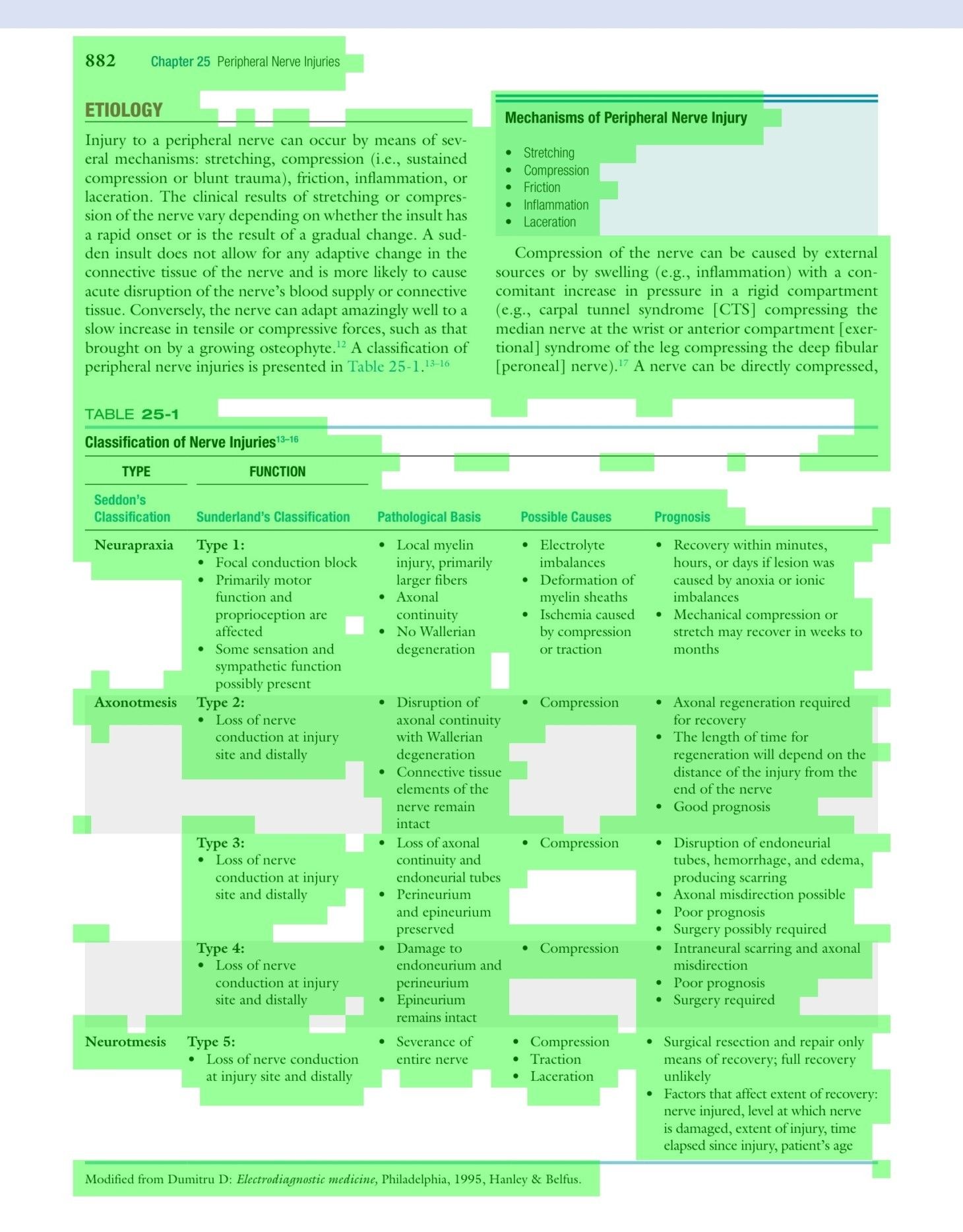}
\end{subfigure}

\vspace{0.5em}

\begin{subfigure}[t]{0.16\linewidth}
    \centering
    \includegraphics[width=\linewidth]{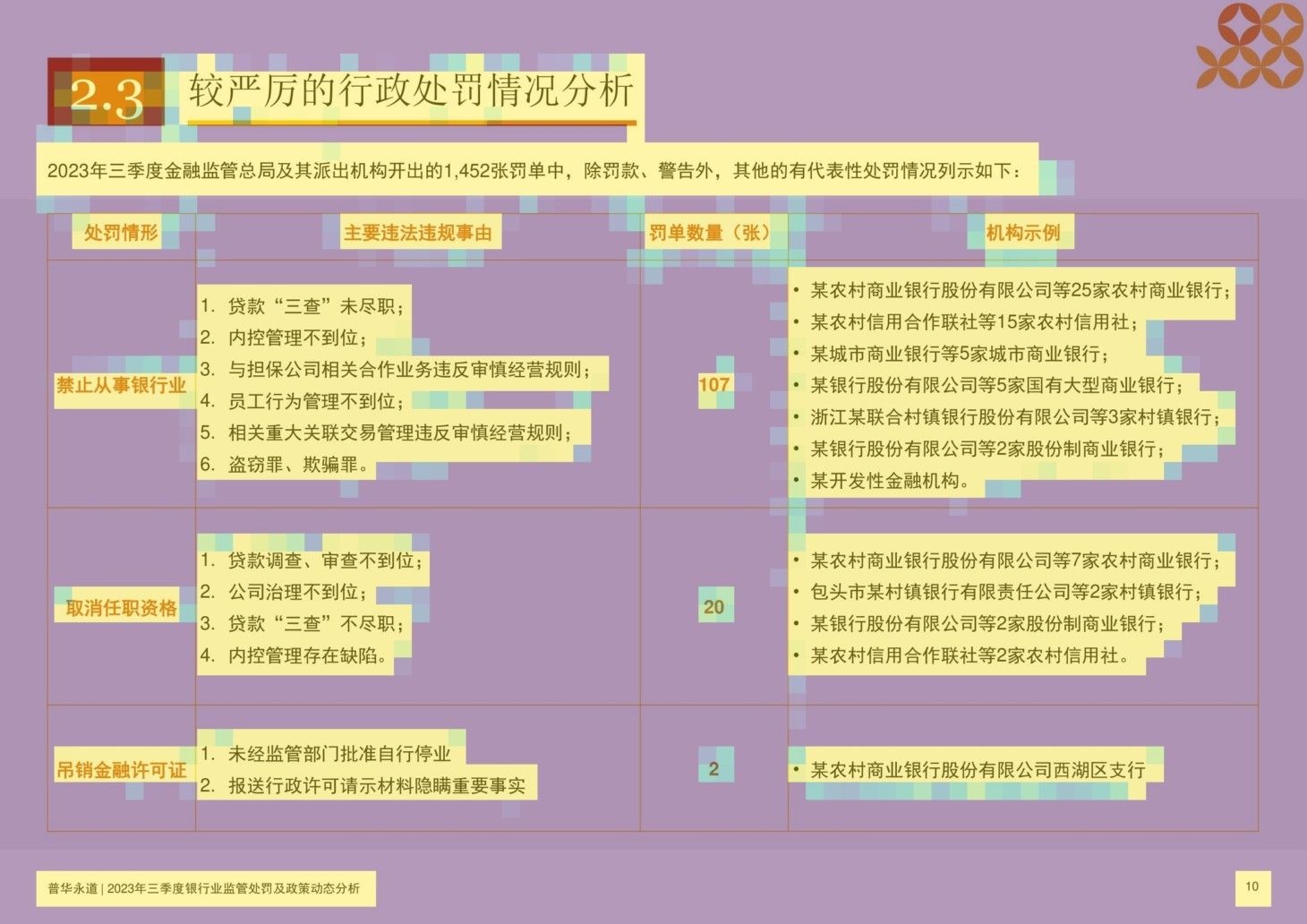}
\end{subfigure}
\hfill
\begin{subfigure}[t]{0.16\linewidth}
    \centering
    \includegraphics[width=\linewidth]{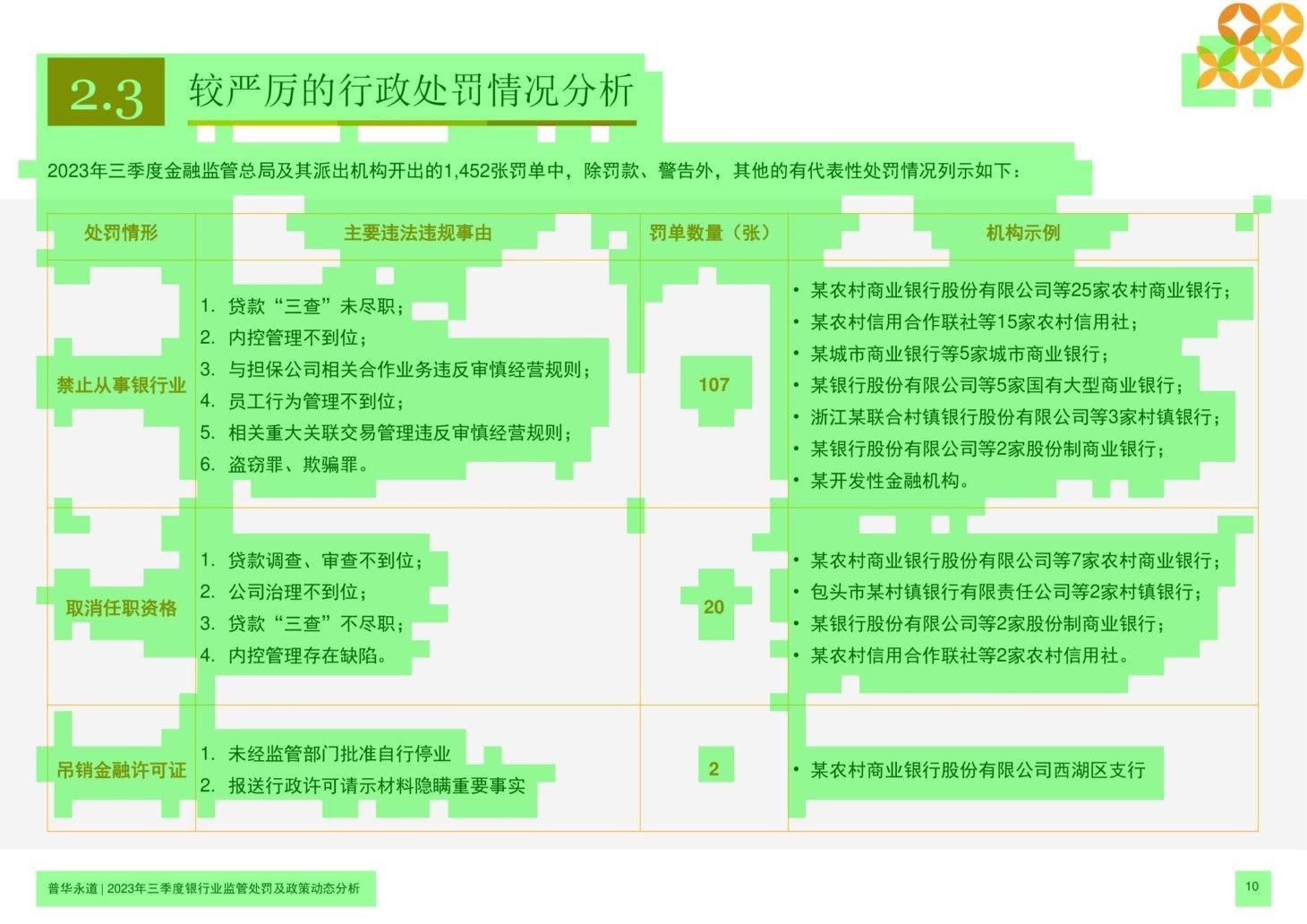}
\end{subfigure}
\hfill
\begin{subfigure}[t]{0.16\linewidth}
    \centering
    \includegraphics[width=\linewidth]{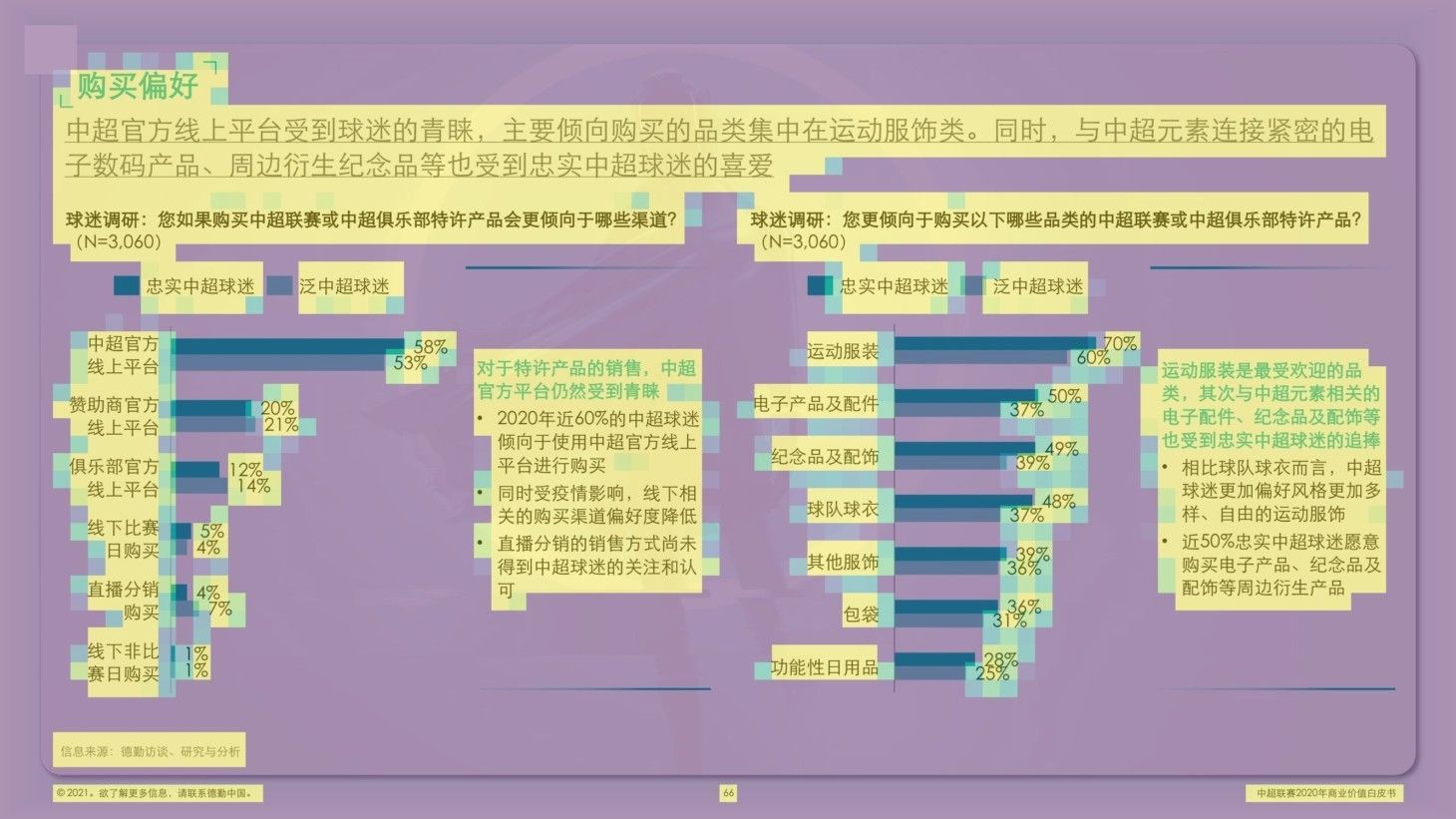}
\end{subfigure}
\hfill
\begin{subfigure}[t]{0.16\linewidth}
    \centering
    \includegraphics[width=\linewidth]{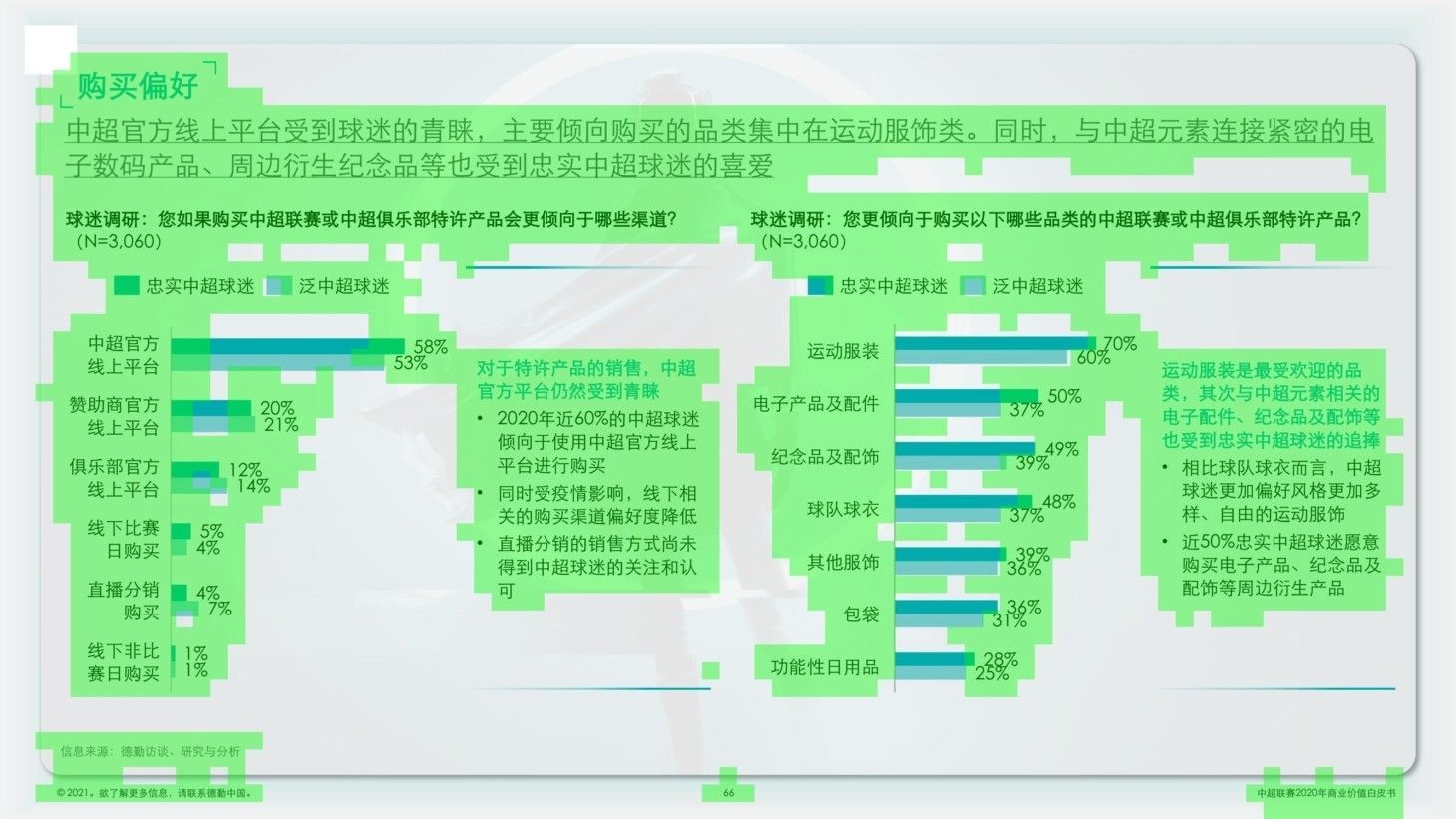}
\end{subfigure}
\hfill
\begin{subfigure}[t]{0.16\linewidth}
    \centering
    \includegraphics[width=\linewidth]{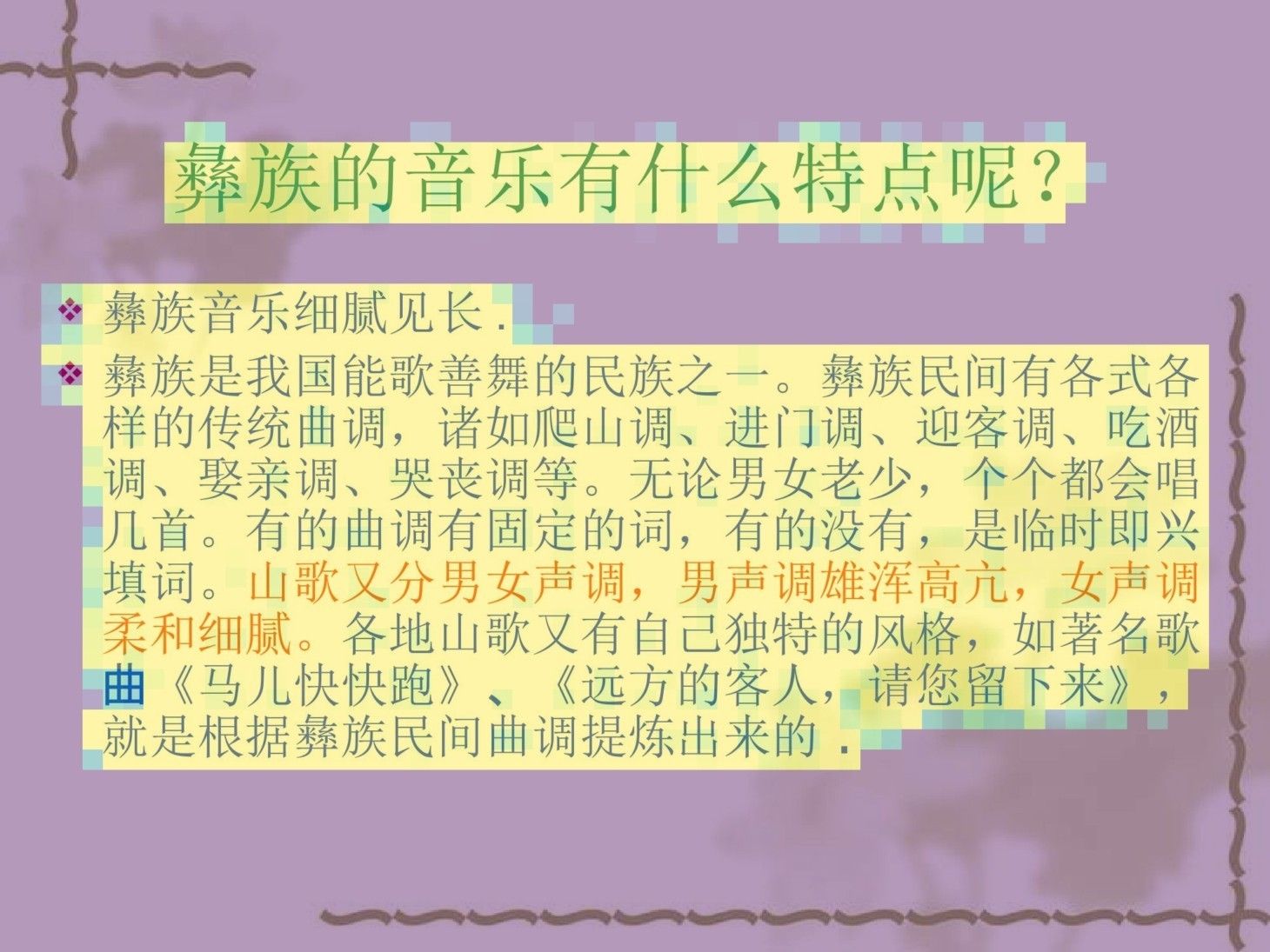}
\end{subfigure}
\hfill
\begin{subfigure}[t]{0.16\linewidth}
    \centering
    \includegraphics[width=\linewidth]{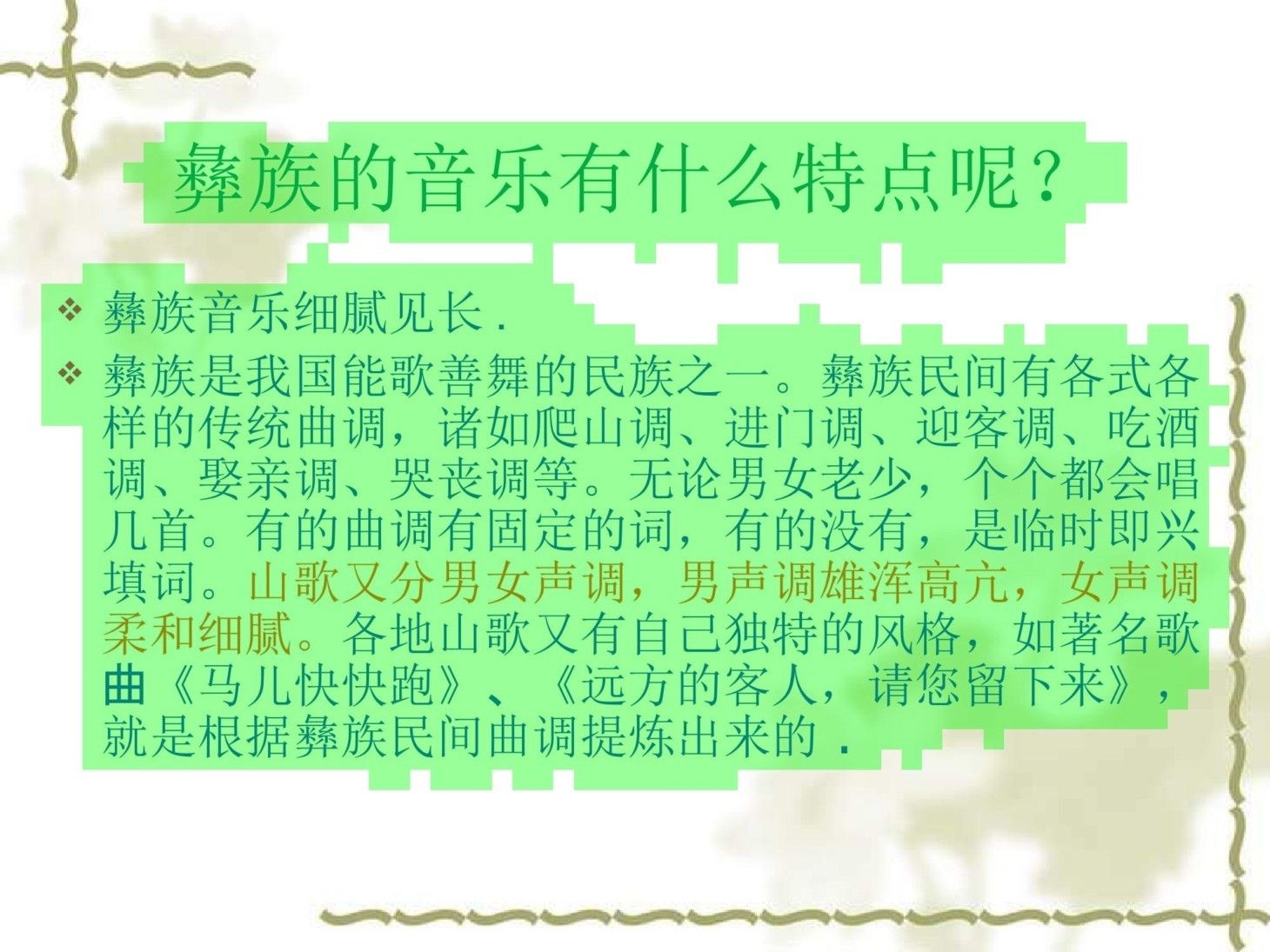}
\end{subfigure}
\caption{Visualization results. Token masks are generated to achieve an average compression rate at 50\%.}
\label{fig:vis}
\end{figure*}

Additional visualization results are presented in Figure~\ref{fig:vis}. For each example, we visualize both the token importance heatmap predicted by the scoring model and the corresponding token pruning mask under an average visual token compression ratio of 50\%.

\end{document}